\def\year{2021}\relax
\documentclass[letterpaper]{article} % DO NOT CHANGE THIS
\usepackage{aaai21}  % DO NOT CHANGE THIS
\usepackage{times}  % DO NOT CHANGE THIS
\usepackage{helvet} % DO NOT CHANGE THIS
\usepackage{courier}  % DO NOT CHANGE THIS
\usepackage[hyphens]{url}  % DO NOT CHANGE THIS
\usepackage{graphicx} % DO NOT CHANGE THIS
\usepackage{graphicx}  % DO NOT CHANGE THIS
\usepackage{natbib}  % DO NOT CHANGE THIS AND DO NOT ADD ANY OPTIONS TO IT
\usepackage{caption} % DO NOT CHANGE THIS AND DO NOT ADD ANY OPTIONS TO IT
\usepackage{style}
\title{Learning to Resolve Conflicts for Multi-Agent Path Finding with Conflict-Based Search}
\author{Taoan Huang, Bistra Dilkina, Sven Koenig\\}
\affiliations{University of Southern California\\\{taoanhua, dilkina, skoenig\}@usc.edu}
\begin{document}

\maketitle

\begin{abstract}
Conflict-Based Search (CBS) is a state-of-the-art algorithm for multi-agent path finding. At the high level, CBS repeatedly detects conflicts and resolves one of them by splitting the current problem into two subproblems. Previous work chooses the conflict to resolve by categorizing the conflict into three classes and always picking a conflict from the highest-priority class. In this work, we propose an oracle for conflict selection that results in smaller search tree sizes than the one used in previous work. However, the computation of the oracle is slow. Thus, we propose a machine-learning framework for conflict selection that observes the decisions made by the oracle and learns a conflict-selection strategy represented by a linear ranking function that imitates the oracle's decisions accurately and quickly. Experiments on benchmark maps indicate that our method significantly improves the success rates, the search tree sizes and runtimes over the current state-of-the-art CBS solver.
\end{abstract}

\section{1\,\,Introduction}
Multi-Agent Path Finding (MAPF) is the problem of finding a set of conflict-free paths for a given number of agents on a given graph that minimizes the sum of costs or the makespan. Although MAPF is NP-hard to solve optimally \cite{yu2013planning}, significant research effort has been devoted to MAPF to support its application in distribution centers~\cite{ma2017lifelong,honig2019persistent}, traffic management~\cite{dresner2008multiagent}, airplane taxiing \cite{morris2015self,balakrishnan2007framework} and computer games \cite{ma2017feasibility}. 

Conflict-Based Search (CBS) \cite{sharon2015conflict} is one of the leading algorithms for solving MAPF optimally, and a number of enhancements to CBS have been developed \cite{boyarski2015icbs,li2019improved,felner2018adding,barer2014suboptimal}. The key idea behind CBS is to use a bi-level search that resolves conflicts by adding constraints at the high level and replans paths for agents respecting these constraints at the low level. The high level of CBS is a best-first search on a binary search tree called {\it constraint tree} (CT). To expand a CT node (that consists of a set of paths and a set of constraints on these paths), CBS has to choose a conflict in the current set of paths to resolve and add constraints that prevent this conflict in the child nodes. Picking good conflicts is important, and a good strategy for conflict selection could have a big impact on the efficiency of CBS by reducing both the size of CT and its runtime. \cite{boyarski2015icbs} proposes to prioritize conflicts by categorizing them into three classes and always picking one from the top class. Such a strategy has been proven to be efficient \cite{boyarski2015icbs} and is commonly used for conflict selection in recent research \cite{li2019improved,felner2018adding,li2019multi}. In this paper, we propose a new conflict-selection oracle that results in smaller CT sizes than the one used in previous work but is much more computationally expensive since it has to compute 1-step lookahead heuristics for each conflict.

To overcome the high computational cost of the oracle, we leverage insights from studies on variable selection for branching in Mixed Integer Linear Programming (MILP) solving and propose to use machine learning (ML) techniques for designing conflict-selection strategies that imitate the oracle's decisions to speed up CBS. Variable selection for branching in MILP is analogous to conflict selection in CBS. As part of the branch-and-bound algorithm for MILP \cite{wolsey1999integer}, non-leaf nodes in the CT must be expanded into two child nodes by selecting one of the unassigned variables and splitting its domain by adding new constraints, while CBS chooses and splits on conflicts. Recent studies \cite{khalil2016learning,khalil2017learning,he2014learning} have shown that data-driven ML approaches for MILP solving are competitive with and can even outperform state-of-the-art commercial solvers. 

We borrow such ML tools from MILP solving \cite{khalil2016learning} and propose a data-driven framework for designing conflict-selection strategies for CBS. In the first phase of our framework, we observe and record decisions made by the oracle on a set of instances and collect data on features that characterize the conflicts at each CT node. In the second phase, we learn a ranking function for conflicts in a supervised fashion that imitates the oracle but is faster to evaluate. In the last phase, we use the learned ranking function to replace the oracle and select conflicts in CBS to solve unseen instances. Compared to previous work on conflict selection in CBS, \textit{our approach is able to discover more efficient rules for conflict selection that significantly 
improve the success rate and reduce the CT size and the runtime of the search}. Our method is flexible since we are able to customize the conflict-selection strategies easily for different environments and do not need to hard-code different rules for different scenarios. Different from recent work on ML-guided MILP solving, we utilize problem-specific features which contain essential information about the conflicts, while previous work only takes MILP-level features (e.g., counts and statistics of variables) into account \cite{khalil2016learning,khalil2017learning}. Another advantage of our offline learning method over training an  instance-specific model on-the-fly is that our learned ranking function is able to generalize to instances and graphs unseen during training.

\section{2\,\,MAPF}
  Given an undirected unweighted underlying graph $G=(V,E)$, the \textit{Multi-Agent Path Finding (MAPF) problem} is to find a set of conflict-free paths for a set of agents $\{a_1,\ldots,a_k\}$. Each agent $a_i$ is assigned a start vertex $s_i\in V$ and a goal vertex $t_i\in V$. Time is discretized into time steps, and, at each time step, every agent can either move to an adjacent vertex or wait at the same vertex in the graph. The cost of an agent is the number of  time steps until it reaches its goal vertex and no longer moves. We consider two types of conflicts: i) a vertex conflict $\langle a_i,a_j,v,t\rangle$ occurs when agents $a_i$ and $a_j$ are at the same vertex $v$ at time step $t$; and ii) an edge conflict  $\langle a_i,a_j,u,v,t\rangle$ occurs when agents $a_i$ and $a_j$ traverse the same edge $(u,v)\in E$ in opposite directions between time steps $t$ and  $t+1$.
Our objective is to find a set of conflict-free paths that move all agents from their start vertices to their goal vertices with the optimal cost, that is, the minimum sum of all agents' costs.

\section{3\,\,Background and Related Work}
In this section, we first provide a brief introduction to CBS and its variants. Then, we summarize other related work using ML in MAPF and MILP solving.

\subsection{Conflict-Based Search (CBS)}
CBS is a bi-level tree search algorithm. It records the following information for each CT node $N$:
\begin{enumerate}
    \item $N_{\Constraint}$: There are two types of constraints: i) a vertex constraint $\langle a_i,v,t\rangle$, corresponding to a vertex conflict, prohibits agent $a_i$ from being at vertex $v$ at time step $t$; and ii) an edge constraint $\langle a_i,u,v,t\rangle$, corresponding to an edge conflict, prohibits agent $a_i$ from moving from vertex $u$ to vertex $v$ between time steps $t$ and  $t+1$.
    \item $N_{\Solution}$: A solution of $N$ consists of a set of individually cost-minimal paths for all agents respecting the constraints in $N_{\Constraint}$. An individually cost-minimal path for an agent is the cost-minimal path between its start and goal vertices assuming it is the only agent in the graph.
    \item $N_{\Cost}$: the cost of $N$ defined as the sum of costs of the paths in $N_{\Solution}$.
    \item $N_{\Conflicts}$: the set of conflicts  in $N_{\Solution}$.
\end{enumerate}
On the high level, CBS starts with a tree node whose set of constraints is empty and expands the CT in a best-first manner by always expanding a tree node with the lowest $N_{\Cost}$.
After choosing a tree node to expand, CBS identifies the set of conflicts $N_{\Conflicts}$ in $N_{\Solution}$. If there are none, CBS terminates and returns $N_{\Solution}$. Otherwise, CBS randomly (by default) picks one of the conflicts to resolve and adds two child nodes of $N$ to the CT by imposing, depending on the type of conflict, an edge or vertex constraint for one of two conflicting agents to $N_{\Constraint}$ of one of the child node under and for the other conflicting agent to $N_{\Constraint}$ of the other child node.
On the low level, it replans the paths in $N_{\Solution}$ to accommodate the newly-added constraints, if necessary. CBS guarantees completeness by exploring both ways of resolving each conflict and optimality by performing best-first searches on both of its high and low levels.
\begin{table*}[t]
\centering
\small{
\begin{tabular}{|c|r|r|r|r||r|r|r|r|}
\hline
 & \multicolumn{4}{c||}{The Random Map} & \multicolumn{4}{c|}{The Game Map} \\
\hline

 & {Runtime} &  CT Size & Oracle Time & Search Time & Runtime & CT Size & Oracle Time & Search Time \\
\hline
CBSH2+$O_0$ & { 9.95s} & 2,362 nodes & {\bf0.00s} & 9.95s & { 2.3min} & 952 nodes& {\bf0.0min} & 2.3min \\
\hline
CBSH2+$O_1$ & 24.89s & 746 nodes & 21.34s & 3.55s & 19.8min & {\bf 565 nodes}  & 19.0min & {\bf 0.8min} \\
\hline
CBSH2+$O_2$ & 12.13s &{\bf 632 nodes}  & 9.52s & \bf{2.61s} & 27.4min & 2,252 nodes & 23.4min & 4.0min\\
\hline 
Our Solver &{ \bf 6.19s}  & 998 nodes & 0.88s & 5.31s & {\bf1.6min} & 754 nodes & 0.2min& 1.4min\\
\hline
\end{tabular}}

\caption{Performance of CBSH2 with different oracles and our solver. Oracle time is the total time that the oracle takes per instance. Search time is the runtime minus the oracle time. All entries are averages taken over the instances that are solved by all solvers.\label{oracleperform}}
\end{table*}
\subsection{Variants of CBS}
CBS chooses conflicts randomly, but this conflict-selection strategy can be improved. Improved CBS (ICBS) \cite{boyarski2015icbs} categorizes conflicts into three types to prioritize them. A conflict is cardinal iff, when CBS uses the conflict to split CT node $N$, the costs of both resulting child nodes are strictly larger than $N_{\Cost}$. A conflict is semi-cardinal iff the cost of one of the child nodes is strictly larger than $N_{\Cost}$ and the cost of the other child node is the same as $N_{\Cost}$. A conflict is non-cardinal otherwise. By first resolving cardinal conflicts, then semi-cardinal conflicts and finally non-cardinal conflicts, CBS is able to improve its efficiency since it increases the lower bound on the optimal cost more quickly by generating child nodes with larger costs. 
ICBS uses Multi-Valued Decision Diagrams (MDD) to classify conflicts. An MDD for agent $a_i$ is a directed acyclic graph consisting of all cost-minimal paths from $s_i$ to $t_i$ of a given cost that respect the current constraints $N_{\Constraint}$. The nodes at depth $t$ of the MDD are exactly the nodes that agent $a_i$ could be at when following one of its cost-minimal paths. A vertex (edge) conflict $\langle a_i,a_j,v,t\rangle$  ($\langle a_i,a_j,u,v,t\rangle$) is cardinal iff vertex $v$ (edge $(u,v)$) is the only vertex at depth $t$ (the only edge from depth $t$ to depth $t+1$) in the MDDs of both agents. \citet{li2019disjoint} proposes to add disjoint constraints to two child nodes when expanding a CT node in CBS and prioritize conflicts based on the number of singletons in or the widths of the MDDs of both agents.

Another line of research focuses on speeding up CBS by calculating a tighter lower bound on the optimal cost to guide the high-level search. When expanding a tree node $N$, CBSH \cite{felner2018adding} uses the CG heuristic, which builds a conflict graph (CG) whose vertices represent agents and whose edges represent cardinal conflicts in $N_{\Solution}$. Then, the lower bound on the optimal cost within the subtree rooted at $N$ is guaranteed to increase at least by the size of the minimum vertex cover of this CG. We refer to this increment as the $h$-value of the CT node.
Based on CBSH, CBSH2 \cite{li2019improved} uses the DG and WDG heuristics that generalize CG and compute $h$-values for CT nodes using (weighted) pairwise dependency graphs that take into account semi-cardinal and non-cardinal conflicts besides cardinal ones. CBSH2 with the WDG heuristic is the current state-of-the-art CBS solver for MAPF \cite{li2019improved}.

To the best of our knowledge, other than prioritizing conflicts using MDDs, conflict prioritization has not yet been explored. \citet{barer2014suboptimal} proposes a number of heuristics to prioritize CT nodes for the high-level search, including those using the number of conflicts, the number of conflicting agents and the number of conflicting pairs of agents. However, this work uses conflict-related metrics to select CT nodes, while we learn to select conflicts.  

\subsection{Other Related Work}
ML techniques are not often applied to MAPF. \cite{sartoretti2019primal} proposes a reinforcement-learning framework for learning decentralized policies for agents offline to avoid the cost of planning online. Our work is different from their work since we focus on search algorithms and use ML to find efficient and flexible conflict-selection strategies to speed up them. Furthermore, our ML model is simple and easy to implement, without the need to train and fine-tune a deep neural network. 

Using ML to speed up search has been explored in the context of MILP solving. \cite{khalil2016learning} uses ML to design strategies for branching that mimic strong branching. Our overall framework is similar to \cite{khalil2016learning} but different in several aspects. Instead of collecting training data and learning a model online, we collect training data and learn a model offline. We leverage insights from existing heuristics for computing $h$-values to design problem-specific labels and features for learning. Finally, once our model is learned, it performs well on unseen instances while \cite{khalil2016learning} learns instance-specific models. Their line of work also includes learning when to run primal heuristics to find incumbents in a tree search \cite{khalil2017learning} and  learning how to order nodes adaptively for branch-and-bound algorithms \cite{he2014learning}.

\section{4\,\,Oracles for Conflict Selection}
Given a MAPF instance, at a particular CT node $N$ with the set of conflicts $N_{\Conflicts}$, an oracle for conflict selection is a ranking function that takes $N_{\Conflicts}$ as input, calculates a real-valued score per conflict and outputs the ranks determined by the scores. We say that CBS follows an oracle for conflict selection iff CBS builds the CT by always resolving the conflict with the highest rank.
We define oracle $O_0$ to be the one proposed by \cite{boyarski2015icbs}, that uses MDDs to rank conflicts.
\setcounter{section}{4}
\begin{definition}
Given a CT node $N$, oracle $O_0$ ranks the conflicts in $N_{\Conflicts}$ in the order of cardinal conflicts, semi-cardinal conflicts and non-cardinal conflicts, breaking ties in favor of  conflicts at the smallest time step and remaining ties randomly.
\end{definition}

Next, we define oracles $O_1$ and  $O_2$ that both calculate 1-step lookahead scores by using, for each conflict, the two child nodes of $N$ that would result if the conflict were resolved at $N$.
\begin{definition}
Given  a CT node $N$, oracle $O_1$ computes the score $v_c=\min\{g_c^l+h_c^l,g_c^r+h_c^r\}$ for each conflict $c\in N_{\Conflicts}$, where $g_c^l$ and $g_c^r$ would be the costs of the two child nodes of $N$ and $h_c^l$ and $h_c^r$ would be the $h$-values given by the WDG heuristic of the two child nodes of $N$ if conflict $c$ were resolved at $N$. Then, it outputs the ranks determined by the descending order of the scores (i.e., the highest rank for the highest score).
\end{definition}
Oracle $O_1$ chooses the conflict that results in the tightest lower bound on the optimal cost in the child nodes. 
We use the WDG heuristic to compute the $h$-values since it is the state of the art.
The intuition behind using this oracle is that the sum of the cost and the $h$-value of a node is a lower bound on the cost of any solution found in the subtree rooted in the node, and, thus, we want CBS to increase the lower bound as much as possible to find a solution quickly.

\begin{definition}
Given  a CT node $N$, oracle $O_2$ computes the score $v_c=\min\{m_c^l, m_c^r\}$ for each conflict $c\in N_{\Conflicts}$, where $m_c^l$ and $m_c^r$ would be the number of conflicts in the two child nodes of $N$ if conflict $c$ were resolved at $N$. Then, it outputs the ranks determined by the increasing order of the scores (i.e., the highest rank for the lowest score).
\end{definition}
Oracle $O_2$ chooses the conflict that results in the least number of conflicts in the child nodes.

We use CBSH2 with the WDG heuristic as our search algorithm and run it with oracles $O_0,O_1$ and $O_2$ on (1) a random map, a  $20\times 20$ four-neighbor grid map with $25\%$ randomly generated blocked cells, and (2) the game map { ``lak503d''} \cite{sturtevant2012benchmarks}, a $192\times 192$ four-neighbor grid map   with $51\%$ blocked cells. The figures of the maps are shown in Table \ref{thehugetable}. The experiments are conducted on 2.4 GHz Intel Core i7 CPUs with 16 GB RAM. We set the runtime limit to 20 minutes for the random map and 1 hour for the game map. We set the number of agents to $k=18$ for the random map and $k=100$ for the game map and run the solvers on 50 instances for each map. Following \citet{stern2019multi}, 
the start and goal vertices are randomly paired among all vertices in each map's largest connected component for each instance throughout the paper. %with randomly pairwise different start vertices and goal vertices for each map. We ensure that there exists a path between any pair of start and goal vertices.
In Table \ref{oracleperform}, we present the performance of the three oracles as well as our solver. All entries are averages taken over the instances that are solved by all solvers. We consider the CT size since a small CT size implies a small runtime and first look at the performance of CBSH2 with the three oracles. Oracle $O_2$ is the best for the random map, followed closely by oracle $O_1$. Oracle $O_1$ is the best for the game map. Overall, oracle $O_1$ is the best. Therefore, 
%While oracle $O_1$ is computationally expensive, it reduces the CT sizes by $68\%$ on the small map and $41\%$ on the large map in comparison to  oracle $O_0$. Although oracle $O_2$ reduces the CT sizes by $73\%$ on the small map, even more than oracle $O_1$, it is almost 4 times worse than oracle $O_1$ on the large map. 
in the rest of the paper, we mainly focus on learning a ranking function to imitate oracle $O_1$. Table \ref{oracleperform} shows that by learning to imitate oracle $O_1$, our solver achieves the best performance in term of the runtime, even though it induces a larger CT than CBSH2+$O_1$. We introduce our machine learning methodology in Section 5 and show experimental results in Section 6. We use the solver, \MLS, introduced in Section 6 to generate results of our solver in Table \ref{oracleperform}. %since it is the most promising one for both small and large maps.
\begin{table*}[t]
%\small
\begin{tabular}{|p{0.9\textwidth}|c|}
\hline
\multicolumn{1}{|c|}{Feature Descriptions} & Count \\ \hline
Types of the conflict: binary indicators for edge conflicts, vertex conflicts, cardinal conflicts, semi-cardinal conflicts and non-cardinal conflicts. & 5 \\ \hline
Number of conflicts involving agent $a_i$ ($a_j$) that have been selected and resolved so far during the search: their min., max. and sum. & 3 \\ \hline
Number of conflicts that have been selected and resolved so far during the search at vertex $u$ ($v$): their min., max. and sum. & 3 \\ \hline
Number of conflicts that agent $a_i$ ($a_j$) is involved in: their min., max. and sum. & 3 \\ \hline
Time step $t$ of the conflict. & 1 \\ \hline
Ratio of $t$ and the makespan of the solution. & 1 \\ \hline
Cost of the path of agent $a_i$ ($a_j$): their min., max., sum, absolute difference and ratio. & 5 \\ \hline
Difference of the costs of the path of agent $a_i$ ($a_j$) and its individually cost-minimal path: their min. and max.  & 2\\ \hline
Ratio of the cost of the path of agent $a_i$ ($a_j$) and the cost of its individually cost-minimal path: their min. and max. & 2\\ \hline
%Difference and ratio of the costs of the paths of agents $a_i$ and $a_j$.  & 2 \\ \hline
Difference of the cost of the path of agent $a_i$ ($a_j$) and $t$: their min. and max. & 2\\ \hline
Ratio of the cost of the path of agent $a_i$ ($a_j$) and $t$: their min. and max. & 2\\ \hline
%The cost of the individually cost-minimal path of agent $a_i$ (agent $a_j$). &2\\ \hline
Ratio of the cost of the path of agent $a_i$ ($a_j$) and $N_{\Cost}$: their min. and max. & 2 \\ \hline
Binary indicator whether none (at least one) of agents $a_i$ and $a_j$ has reached its goal by time step $t$. & 2 \\ \hline
%Differences and ratios of the costs and $t$ for both agents. & 4 \\ \hline
%For a conflict $c'=\langle a'_i,a'_j,u',t'\rangle$ ($\langle a'_i,a'_j,u',v',t'\rangle$) in $N_{\Conflicts}$, let $d_{1}^{c'}$ be the
Number of conflicts $c'\in N_{\Conflicts}$ such that $\min\{d_{q,q'}:q\in V^T_c, q'\in V^T_{c'}\}=w$ ($0\leq w\leq 5$). & 6 \\ \hline
%Number of conflicts at any of the vertices  $k\, (0\leq k\leq 5)$  steps away from $(u,t)$ (and $(v,t)$) in the time-expanded graph. & 6 \\ \hline
Number of agents $a$ such that there exists $q'\in V_a$ and $q\in V_c^T$ such that $d_{q,q'}=w$ ($0\leq w\leq 5$). & 6 \\ \hline
%Number of  agents at any of the vertices $k\,(0\leq k\leq 5)$  steps away from $(u,t)$ (and $(v,t)$) in the time-expanded graph. & 6 \\ \hline

Number of conflicts $c'\in N_{\Conflicts}$ such that $\min\{d_{q,q'}:q\in V_c,q'\in V_{c'}\}=w$ ($0\leq w\leq 5$). & 6 \\ \hline
%Number of conflicts at any of the vertices $k\,(0\leq k\leq 5)$ steps away from $u$ (and $v$) in graph $G$. & 6 \\ \hline
Width of level $w\, (|w-t|\leq 2)$ of the MDD for agent $a_i (a_j)$: their min. and max. \cite{li2019disjoint}. & 10 \\ \hline
 Weight of the edge between agents $a_i$ and $a_j$ in the weighted dependency graph \cite{li2019improved}. & 1 \\ \hline
Number of vertices $q'$ in graph $G$ such that $\min\{d_{q',q}:q\in V_c\}=w$ ($1\leq w\leq 5$). & 5 \\ \hline
\end{tabular}
\caption{Features of a conflict $c=\langle a_i,a_j,u,t\rangle$ ($\langle a_i,a_j,u,v,t\rangle$) of a CT node $N$. Given the underlying graph $G=(V,E)$, let $V_T=\{(v,t):v\in V, t\in \mathbb{Z}_{\geq 0}\}$, $E_T=\{((u,t),(v,t+1)):t\in \mathbb{Z}_{\geq 0}\land (u=v\lor (u,v)\in E)\}$ and define the time-expanded graph as an unweighted graph $G_T=(V_T,E_T)$. Let $d_{u,v}$ be the cost of the cost-minimal path between vertices $u$ and $v$ in $G$ and $d_{(u',t'),(u,t)}$ be the minimum distance between vertices $(u',t')$ and $(u,t)$ in $G_T$. For a conflict $c'=\langle a'_i,a'_j,u',t'\rangle$ ($\langle a'_i,a'_j,u',v',t'\rangle$) in $N_{\Conflicts}$, define $V_{c'}=\{u'\}$ ($V_{c'}=\{u',v'\}$) and $V^T_{c'}=\{((u',t'))\}$ ($V^T_{c'}=\{(u',t'),(v',t')\}$). For an agent $a$, define $V_a=\{(u,t):\text{agent $a$ is at vertex $u$ at time step $t$ following its path}\}.$ The counts are the numbers of features contributed by the corresponding entries, which add up to $p=67$.
%An agent is at vertex $(u,t)$ in the time-expanded graph iff it is at node $u$ at time step $t$ following its current path in the underlying graph. A conflict $\langle a_i,a_j,v,t\rangle$ ($\langle a_i,a_j,u,v,t\rangle$) is defined to be at vertex $(u,t)$ (vertex $(u,t)$ and $(v,t)$) in the time-expanded graph.   
\label{featureTable}}
\end{table*}
\section{5\,\,Machine Learning Methodology}
We now introduce our framework for learning which conflict to resolve in CBS. 
The key idea is that, by observing and recording the features and ranks of conflicts determined by the scores given by the oracle, we learn a ranking function that ranks the conflicts as similarly as possible to the oracle without actually probing the oracle. Our framework consists of three phases:
\begin{enumerate}
    \item Data collection. We obtain two set of instances, a training dataset $\calI_{\Train}$ and a test dataset $\calI_{\Test}$.  For each instance $I\in\calI_{\Train}\cup \calI_{\Test}$, we obtain a dataset $D_I$ by running the oracle.
    \item Model learning. The training dataset is fed into a machine learning algorithm to learn a ranking function that maximizes the prediction accuracy.
    \item ML-guided search. We replace the oracle with the learned ranking function to rank conflicts in the CBSH2 solver. We run the new solver on randomly generated instances on the same graphs seen during training or unseen graphs.  %drawn from that distribution with the same and different numbers of agents and randomly generated start and goal vertices for them.
\end{enumerate}

\subsection{Data Collection}
The first task in our pipeline is to construct a training dataset from which we can learn a model that imitates the oracle's output. We first fix the graph underlying the instances that we want to solve and the number of agents. The number of agents is only fixed during the data collection and model learning phases. We obtain two sets of instances, $\calI_{\Train}$ for training and $\calI_{\Test}$ for testing. A dataset $D_{I}$ is obtained for each instance $I\in \mathcal{I}_{\Train}$ $(\calI_{\Test})$, and the final training (test) dataset is obtained by concatenating these datasets. To obtain dataset $D_I$, oracle $O_1$ is run for each CT node $N$ to produce the ranking for $N_{\Conflicts}$.  The data consists of: (i) a set of CT nodes $\calN$; (ii) a set of conflicts $N_{\Conflicts}$ for a given $N\in\calN$; (iii) binary labels $y_{N}\in \{0,1\}^{|N_{\Conflicts}|}$ for all $N\in\calN$ transformed from the oracle's ranking of the conflicts; and (iv) a feature map $\Phi_{N}: N_{\Conflicts}\rightarrow [0,1]^p$ for all $N\in\calN$ that describes conflict $c\in N_{\Conflicts}$ at each CT node with $p$ features. The test dataset is used to evaluate the prediction accuracy of the learned model.

\begin{table*}[!htb]

\begin{center}
\begin{tabular}{|l|l|r|r|r|r|r|r|}
\hline
 \multicolumn{2}{|l|}{}  & Game & Random & Maze & Room & Warehouse & City \\ \hline
  \multicolumn{2}{|l|}{Number of agents in instances for data collection} & 100 & 18 & 30 & 22 & 30 & 180\\\hline
\multirow{2}{*}{\begin{tabular}[l]{@{}l@{}}Training on \\ the same map\end{tabular}} & Swapped pairs (\%) & 4.40 & 10.89 & 4.5 & 12.58 & 5.78 & 2.89 \\ \cline{2-8} 
% & 0-1 Error (\%) & 48.74 & 37.28 & 15.94 & 34.33 & 15.08 & 25.23 \\ 
& Top pick accuracy (\%) & 60.16 & 69.03 & 87.69 & 67.56 & 84.93 & 83.05\\
\hline
\multirow{2}{*}{\begin{tabular}[l]{@{}l@{}}Training on \\ the other maps\end{tabular}} & Swapped pairs (\%) & 7.45 & 19.64 & 21.98 & 15.24 & 6.08 & 7.66 \\ \cline{2-8} 
% & 0-1 Error (\%) & 55.49 & 56.55 & 52.05 & 35.67 & 13.23 & 25.23 \\ 
& Top pick accuracy (\%) & 53.13 & 50.44 & 49.90 & 66.80 & 86.85 & 78.57\\

\hline
\end{tabular}
\end{center}

\caption{Numbers of agents in instances for data collection, test losses and accuracies. The swapped pairs (\%) are the fractions of swapped pairs averaged over all test CT nodes and the top pick accuracies are the accuracies of the ranking function picking the conflicts labled as 1 in the test dataset. \label{exppara}}
\end{table*}

\begin{table*}[!tbp]
\scriptsize
\centering
\begin{tabular}{|c|c|rrr|rrr|rrr|rrr|}
\hline
\multirow{2}{*}{Map} & \multirow{2}{*}{$k$} & \multicolumn{3}{c|}{Success Rate (\%)} & \multicolumn{3}{c|}{Runtime (min)} & \multicolumn{3}{c|}{CT Size (nodes)} & \multicolumn{3}{c|}{PAR10 Score (min)} \\ \cline{3-14} 
 &  & CBSH2 & ML-S &ML-O & CBSH2 &  ML-S &ML-O & CBSH2 &  ML-S &ML-O & CBSH2 &  ML-S &ML-O\\ \hline
\multirow{6}{*}{
\begin{minipage}{.07\textwidth}
\begin{center}
Warehouse
\\
\includegraphics[width=1.3cm]{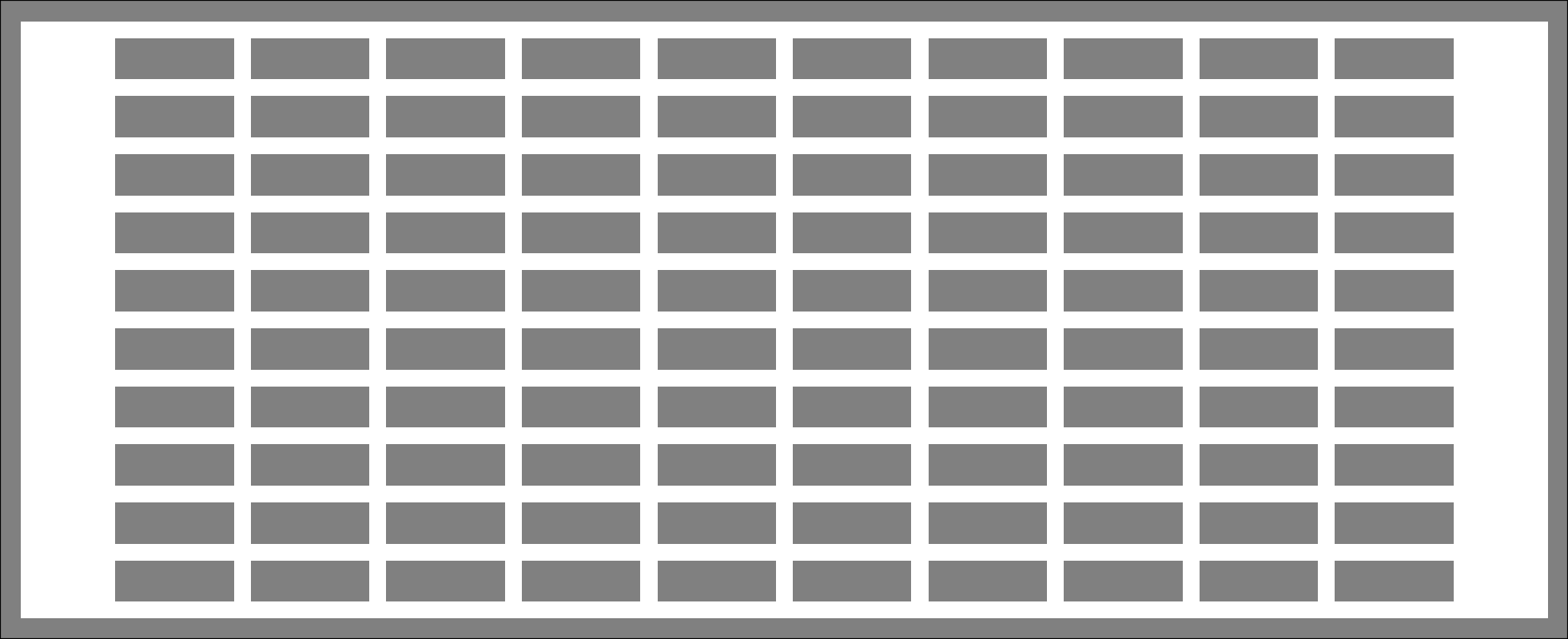}\end{center}
\end{minipage}}
&30 & 93 & {\bf96 (93)} & {\bf96 (93)} & 0.20 & {\bf0.06} & 0.07 & 1,154 & {\bf294} & 378 &  7.18 & {\bf4.14} & 4.25\\ \cline{2-14} 
% & 33 & 92 & {\bf96} (91) & 95 (91) & 0.37 & 0.21 & {\bf0.14} & 2,367 & 1,053 & {\bf721} &  8.37 & {\bf4.38} & 5.27\\ \cline{2-14} 
 & 36 & 72 & 86 (71) & {\bf88} (71) & 0.54 & 0.24 & {\bf0.19} & 3110 & 980 & {\bf977} &  28.46 & 14.56 & {\bf12.81} \\ \cline{2-14} 
% & 39 & 59 & {\bf77} (57) & 75 {\bf(59)} & 0.77 & {\bf0.39} & 0.49 & 4,064 & {\bf1,395} & 2,000 &  41.52 & {\bf23.83} & 25.84 \\ \cline{2-14} 
 & 42 & 55 & 68 {\bf(55)} & {\bf70 (55)} & 1.27 & 0.65 & {\bf0.38} & 6,834 & 2,874 & {\bf1,781} &  45.70 & 32.61 & {\bf30.56} \\ \cline{2-14} 
% & 45 & 36 & {\bf55 (36)} & 53 {\bf(36)} & 2.10 & 0.99 & {\bf0.68} & 9,887 & 4,980 & {\bf3,333} &  64.76 & {\bf45.94} & 47.79 \\ \cline{2-14} 
 & 48 & 17 & {\bf32 (17)} & {\bf32 (17)} & 1.99 & 1.12 & {\bf0.56} & 9,646 & 5,357 & {\bf2,221} &  83.34 & 68.64 & {\bf68.48} \\ \cline{2-14} 
%  & 51 & 11 & 22 {\bf(11)} & {\bf23 (11)} & 1.71 & 0.93 & {\bf0.16} & 7,685 & 3,768 & {\bf429} &  89.19 & 78.31 & {\bf77.26} \\ \cline{2-14} 
 & 54  & 6 & {\bf16 (6)} & 15 {\bf(6)} & 2.82 & 1.70 & {\bf1.23} & 12,816 & 8,886 & {\bf6,427} &  94.17 & {\bf84.42} & 85.36\\ \cline{2-14} 
 & \multicolumn{4}{|r|}{Improvement over CBSH2} & 0 & 49.8\% & {\bf64.4\%} & 0 & 56.6\% & {\bf68.2\%} &0&{\bf22.9\%}&22.7\%\\ 
 \hline  \hline

 \multirow{6}{*}{\begin{minipage}{.07\textwidth}
\begin{center}
Room
\\
\includegraphics[width=1.1cm]{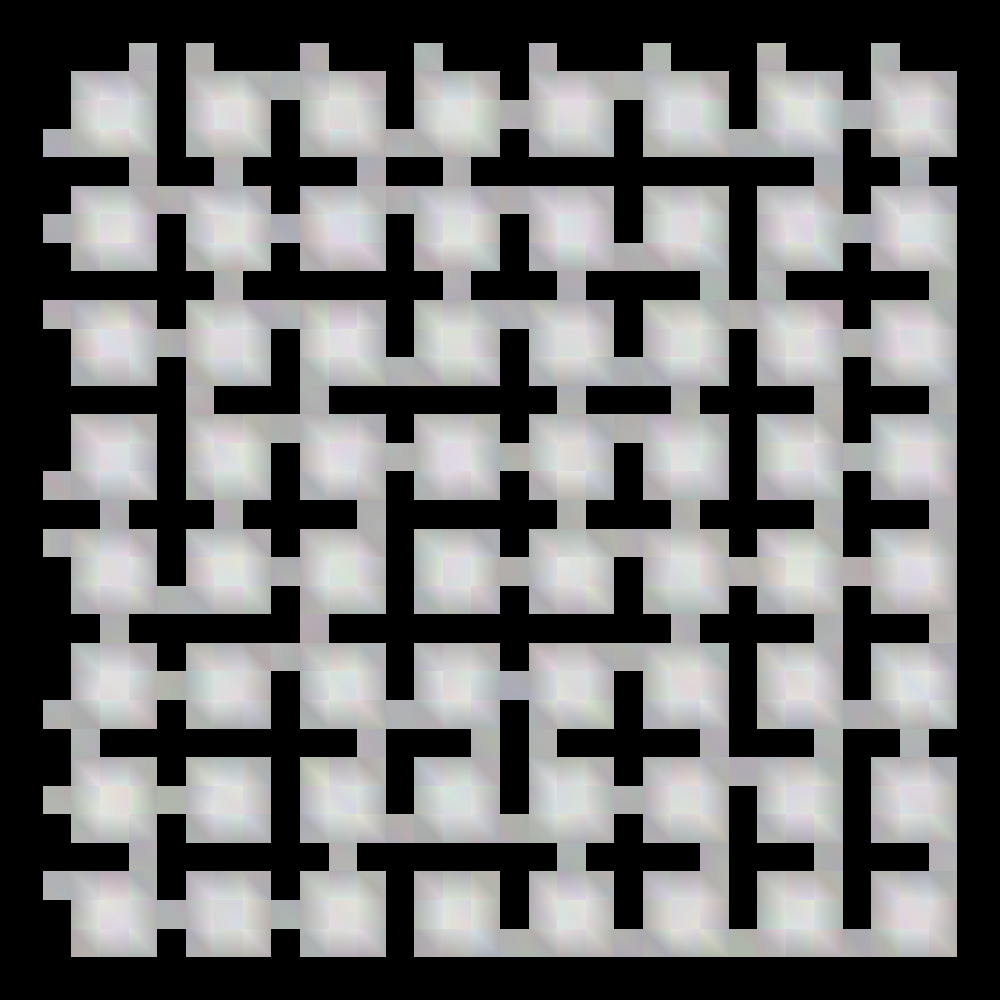}\end{center}
\end{minipage}}
%& 20 & 89 & {\bf96 (89)} & 94 {\bf(89)} & 0.22 & {\bf0.17} & 0.18 & 1,823 & {\bf1,038} & 1,101 &  11.19 & {\bf4.38} & 6.33 \\ \cline{2-14} 
 & 22 & 83 & {\bf91 (83)} & {\bf91 (83)} & 0.61 & {\bf0.49} & 0.51 & 7,851 & {\bf5,648} & 5,888 &  17.51 & {\bf9.76} & 9.83 \\ \cline{2-14} 
% & 24 & 79 & {\bf86 (79)} & 84 {\bf(79)} & 0.69 & {\bf0.53} & 0.55 & 8,392 & {\bf5,007} & 5,160 &  21.55 & {\bf14.83} & 16.68 \\ \cline{2-14} 
 & 26 & 47 & {\bf57 (47)} & 55 (46) & 1.32 & {\bf1.01} & 1.14 & 15,791 & {\bf11,087} & 12,108 &  53.68 & {\bf43.97} & 45.91 \\ \cline{2-14} 
% & 28 & 45 & {\bf52 (45)} & 50 (44) & 1.45 & {\bf0.95} & 0.96 & 15,951 & {\bf10,184} & 10294 &  55.70 & {\bf48.93} & 50.81 \\ \cline{2-14} 
 & 30 & 28 & {\bf36 (28)} & 34 {\bf(28)} & 2.08 & {\bf1.21} & 1.45 & 21,279 & {\bf10,284} & 12,117&  73.22 & {\bf65.32} & 67.28  \\ \cline{2-14} 
 & 32 & 17 & {\bf24 (17)} & {\bf24 (17)} & 1.88 & {\bf1.39} & 1.70 & 22,152 & {\bf13,943} & 16,327&  83.77 & {\bf77.02} & 77.14 \\ \cline{2-14} 
  &34 & 9 & {\bf14 (9)} & {\bf14 (9)} & 3.99 & {\bf2.70} & 3.24 & 39,447 & {\bf22,611} & 28,392 &  91.36 & {\bf86.56} & 86.63 \\ \cline{2-14} 
 & \multicolumn{4}{|r|}{Improvement over CBSH2} & 0 & {\bf26.6\%} & 21.3\% & 0 & {\bf35.2\%} & 32.0\%&0&{\bf 14.0\%}&11.6\% \\ \hline \hline
 \multirow{6}{*}{\begin{minipage}{.07\textwidth}
\begin{center}
Maze
\\\includegraphics[width=1.1cm]{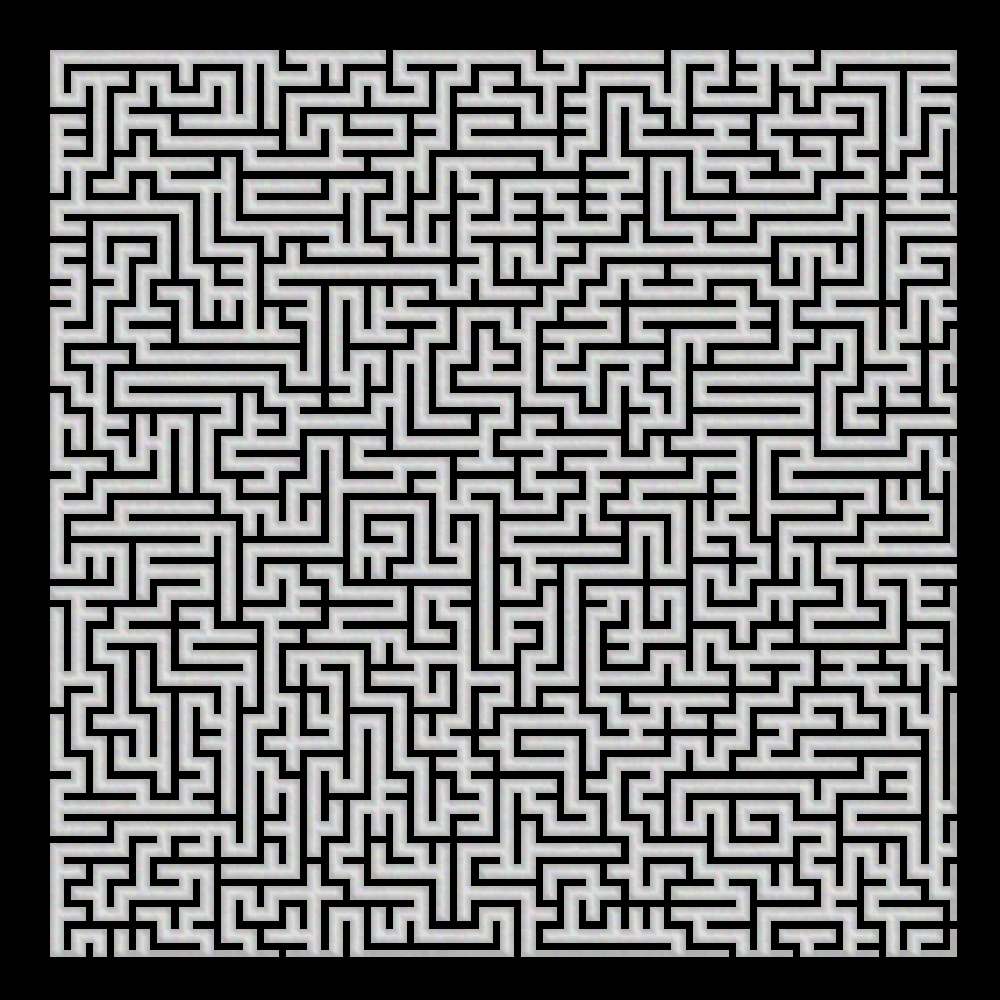}\end{center}
\end{minipage}}
&30 & 90 & {\bf91 (90)} & 90 {\bf(90)} & 0.54 & 0.47 & {\bf0.42} & 500 & 373 & {\bf289} &  10.49 & {\bf9.51} & 10.38 \\ \cline{2-14} 
 & 32 & 84 & {\bf87 (84)} & {\bf87 (84)} & 0.49 & {\bf0.39} & 0.42 & 519 & 427 & {\bf397} &  16.42 & {\bf13.59} & 13.60 \\ \cline{2-14} 
% & 34 & 80 & 82 {\bf(80)} & {\bf84 (80)} & 0.58 & {\bf0.50} & 0.52 & 908 & {\bf763} & 780 &  20.46 & 18.59 & {\bf16.73} \\ \cline{2-14} 
 & 36 & 80 & 81 {\bf(80)} & {\bf82} (79) & 0.73 & 0.65 & {\bf0.57} & 1,200 & 1,067 & {\bf910} &  20.66 & 19.68 & {\bf18.68} \\ \cline{2-14} 
 %& 38 & 64 & {\bf65 (64)} & {\bf65 (64)} & 0.68 & 0.57 & {\bf0.53} & 900 & 740 & {\bf663} &  36.44 & 35.40 & {\bf35.38} \\ \cline{2-14} 
 & 40 & 56 & 60 {\bf(56)} & {\bf62 (56)} & 0.85 & 0.80 & {\bf0.75} & 1,194 & 1,099 & {\bf1,026} &  44.47 & 40.79 & {\bf38.85} \\ \cline{2-14} 
 %& 42 & 54 & 56 {\bf(54)} & {\bf57 (54)} & 1.86 & 1.69 & {\bf1.47} & 2,223 & 1,973 & {\bf1,580} &  47.00 & 45.11 & {\bf44.03} \\ \cline{2-14} 
  & 44 & 45 & 49 {\bf(45)} & {\bf50 (45)} & 1.08 & 1.06 & {\bf0.87} & 1,389 & 1,343 & {\bf1,055}&  54.49 & 50.82 & {\bf49.75}  \\ \cline{2-14}
  %& 46 & 37 & {\bf40 (37)} & {\bf40 (37)} & 1.61 & 1.50 & {\bf1.31} & 2,021 & 1,743 & {\bf1,506} &  63.60 & 60.77 & {\bf60.71}  \\ \cline{2-14} 
 & \multicolumn{4}{|r|}{Improvement over CBSH2} & 0 & {10.3\%} & {\bf18.3\%} & 0 & {13.0\%} & {\bf24.4\%} &0&6.3\%&{\bf8.2\%}\\ \hline
  \hline
  
%  \multirow{2}{*}{Map} & \multirow{2}{*}{$k$} & \multicolumn{3}{c|}{PAR10 Score} & \multicolumn{3}{c|}{Success Rate (\%)} & \multicolumn{3}{c|}{Runtime (min)} & \multicolumn{3}{c|}{CT Size (nodes)} \\ \cline{3-14} 
% &  & CBSH2 & ML-S &ML-O & CBSH2 &  ML-S &ML-O & CBSH2 &  ML-S &ML-O & CBSH2 &  ML-S &ML-O\\ \hline
  \multirow{6}{*}{
  \begin{minipage}{.07\textwidth}
  \begin{center}
Random
\\
\includegraphics[width=1.1cm]{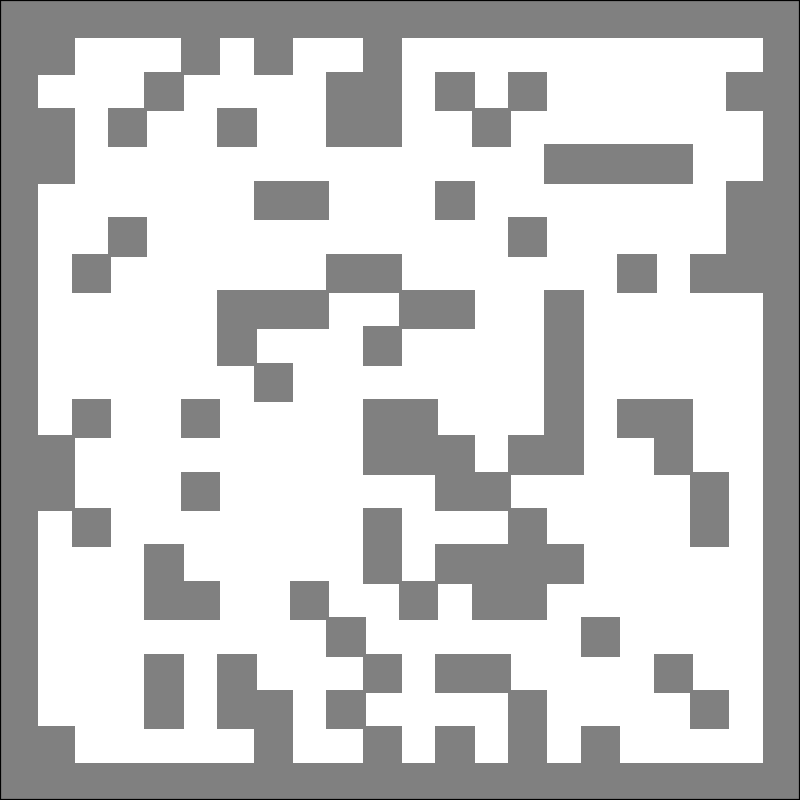}\end{center}
\end{minipage}}
%&17 & 96 & {\bf97 (96)} & {\bf97 (96)} & 0.11 & {\bf0.08} & 0.10 & 1,195 & {\bf743} & 902 &  4.10 & {\bf3.13} & 3.19 \\ \cline{2-14} 
 & 18 & 95 & {\bf95 (95)} & 94 (94) & 0.32 & {\bf0.23} & 0.31 & 5032 & {\bf3,105} & 4,148 &  5.32 & {\bf5.27} & 6.29 \\ \cline{2-14} 
% & 19 & 92 & {\bf93 (92)} & {\bf93 (92)} & 0.44 & {\bf0.32} & 0.36 & 7,208 & {\bf4,264} & 4677 &  8.41 & {\bf7.38} & 7.43\\ \cline{2-14} 
 & 20 & 88 & {\bf91 (88)} & {\bf91 (88)} & 0.43 & {\bf0.30} & 0.36 & 7,834 & {\bf3,829} & 4,595 &  12.38 & {\bf9.37} & 9.48 \\ \cline{2-14} 
% & 21 & 79 & {\bf83 (79)} & 81 {\bf(79)} & 0.59 & {\bf0.49} & 0.56 & 8,814 & {\bf5,244} & 6,927 &  21.47 & {\bf17.53} & 19.54 \\ \cline{2-14} 
% & 22 & 74 & {\bf80 (74)} & 77 {\bf(74)} & 0.90 & {\bf0.48} & 0.59 & 15,884 & {\bf6,286} & 7,577 &  26.66 & {\bf20.69} & 23.54 \\ \cline{2-14} 
 & 23  & 74 & {\bf80 (74)} & {\bf80 (74)} & 0.96 & {\bf0.56} & 0.78 & 17,952 & {\bf8,118} & 11,555 &  26.71 & {\bf20.60} & 20.81\\ \cline{2-14} 
 % & 24 & 62 & {\bf71 (62)} & 66 (60) & 1.14 & {\bf0.77} & 0.92 & 20,433 & {\bf10,413} & 12,422 &  38.87 & {\bf29.82} & 34.71 \\ \cline{2-14}
%  & 25 & 55 & {\bf62 (55)} & 59 {\bf(55)} & 1.19 & {\bf0.85} & 1.02 & 21,725 & {\bf12,137} & 14,874 &  45.66 & {\bf38.84} & 41.74 \\ \cline{2-14} 
  & 26 & 39 & {\bf48 (39)} & 45 {\bf(39)} & 1.27 & {\bf0.87} & 1.24 & 19,236 & {\bf8,053} & 13,301 &  61.50 & {\bf52.75} & 55.82 \\\cline{2-14} 
%  & 27 & 27 & {\bf38 (27)} & 37 {\bf(27)} & 1.36 & {\bf0.99} & 1.05 & 26,642 & {\bf16,597} & 17,130 &  73.41 & {\bf62.72} & 63.79 \\\cline{2-14} 
%  & 28 & 24 & {\bf31 (24)} & 29 (23) & 1.66 & {\bf0.83} & 1.10 & 26,597 & {\bf9,239} & 13,423 &  76.45 & {\bf69.41} & 71.46 \\\cline{2-14} 
  & 29 & 17 & {\bf27 (17)} & 24 {\bf(17)} & 4.04 & {\bf2.74} & 3.39 & 63,661 & {\bf35,485} & 44,179 &  83.69 & {\bf74.07} & 77.02\\ \cline{2-14} 
 & \multicolumn{4}{|r|}{Improvement over CBSH2} & 0 & {\bf33.4\%} & {17.6\%} & 0 & {\bf49.3\%} & {35.8\%} &0&{\bf15.1\%}&10.3\%\\\hline \hline
  
 \multirow{6}{*}{\begin{minipage}{.07\textwidth}
\begin{center}
City
\\
\includegraphics[width=1.1cm]{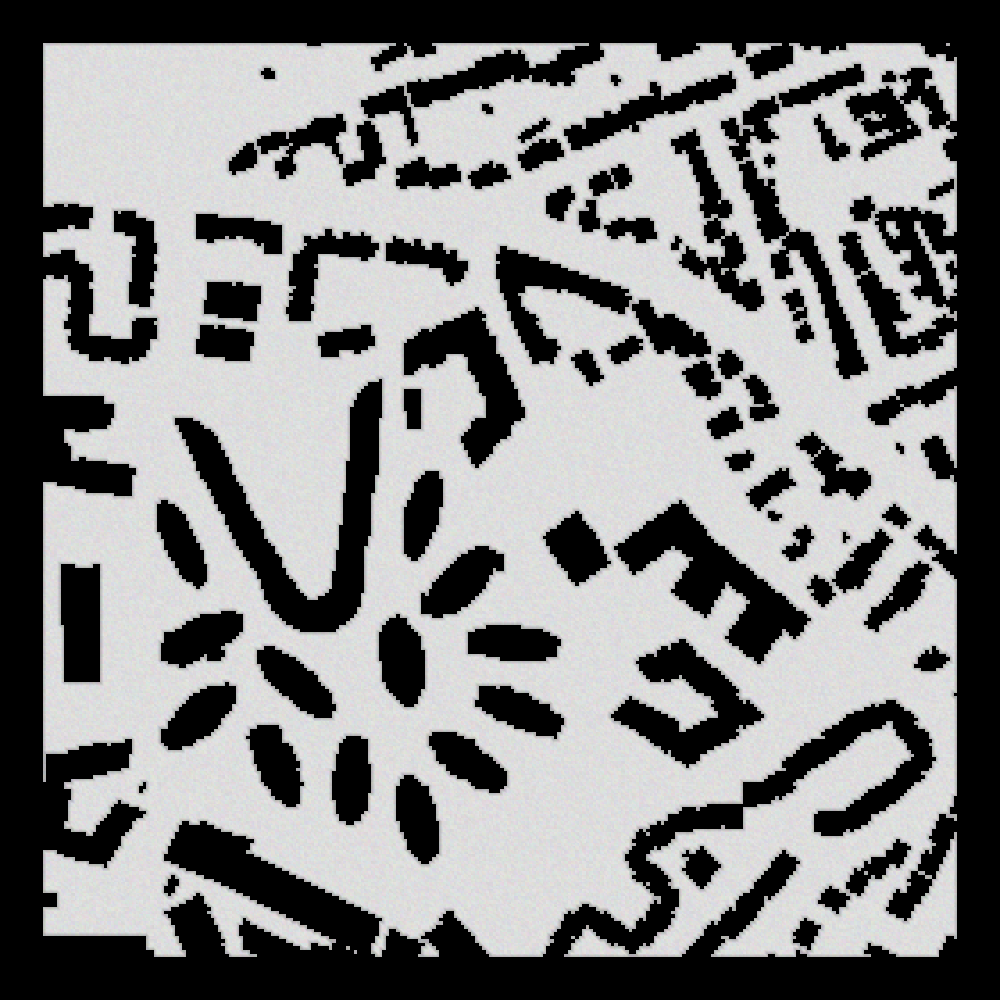}\end{center}
\end{minipage}}
%& 170 & 86 & 92 (84) & {\bf94 (86)} & 3.85 & 2.87 & {\bf2.80} & 669 & {\bf355} & 361 &  87.77 & 50.52 & {\bf38.92} \\ \cline{2-14} 
 & 180 & 78 & {\bf85} (76) & 84 (75) & 3.53 & {\bf2.43} & 2.46 & 859 & {\bf468} & 476 &  134.99 & {\bf93.04} & 99.87 \\ \cline{2-14} 
 %& 190 & 70 & 76 (69) & {\bf77} (68) & 4.42 & {\bf2.96} & 3.07 & 878 & 326 & {\bf325} &  192.05 & 155.37 & {\bf150.13} \\ \cline{2-14} 
 & 200 & 76 & 82 (75) & {\bf83} (75) & 4.78 & 5.08 & {\bf4.13} & 849 & 702 & {\bf490} &  147.96 & 113.53 & {\bf106.78}\\ \cline{2-14} 
 %& 210 & 67 & 75 (65) & {\bf76} (66) & 5.40 & {\bf3.21} & 3.27 & 956 & 339 & {\bf335}&  201.66 & 153.96 & {\bf147.72} \\ \cline{2-14} 
  %& 220 & 46 & {\bf60} (45) & 59 (45) & 8.24 & 6.91 & {\bf5.74} & 1,546 & 648 & {\bf638} &  328.21 & {\bf246.88} & 251.54 \\ \cline{2-14} 
 & 230& 57 & {\bf68} (56) & 64 (54) & 4.86 & 4.26 & {\bf4.24} & 835 & {\bf444} & 449  &  261.36 & {\bf196.99} & 220.50 \\ \cline{2-14} 
 %& 240 & 37 & {\bf55} (35) & {\bf55} (34) & 11.11 & {\bf4.73} & 6.87 & 2,368 & {\bf541} & 968 &  382.45 & {\bf276.16} & 277.02 \\ \cline{2-14} 
 %& 250& 34 & {\bf51 (34)} & 50 {\bf(34)} & 6.34 & {\bf5.44} & 6.25 & 1,288 & {\bf1,057} & 1,170 &  398.18 & {\bf299.93} & 305.62 \\ \cline{2-14}
  & 260 & 44 & {\bf54 (44)} & {\bf54} (43) & 11.69 & 9.75 & {\bf9.55} & 1,883 & {\bf1,178} & 1,219 & 341.37 & {\bf282.00} & 282.58  \\ \cline{2-14} 
 %& 270 & 26 & 35 (24) & {\bf38} (24) & 8.38 & {\bf5.81} & 6.48 & 1,293 & {\bf682} & 822 &  446.51 & 392.84 & {\bf375.70}  \\ \cline{2-14} 
 %& 280 & 23 & 29 (19) & {\bf33} (22) & 7.42 & 7.09 & {\bf4.97} & 1,099 & 811 & {\bf650} &  464.48 & 429.63 & {\bf406.60}  \\ \cline{2-14} 
 & 290  & 18 & 27 (16) & {\bf28} (17) & 11.65 & {\bf8.45} & 8.75 & 1,966 & {\bf1,372} & 1,429 &  494.10 & 441.87 & {\bf436.70}
  \\ \cline{2-14} 
 & \multicolumn{4}{|r|}{Improvement over CBSH2} & 0 & {24.0\%} & {\bf25.2\%} & 0 & {\bf47.3\%} & {46.4\%}&0&19.3\%&{\bf20.2\%} \\\hline\hline

  \multirow{6}{*}{\begin{minipage}{.07\textwidth}
\begin{center}
Game
\\
\includegraphics[width=1.1cm]{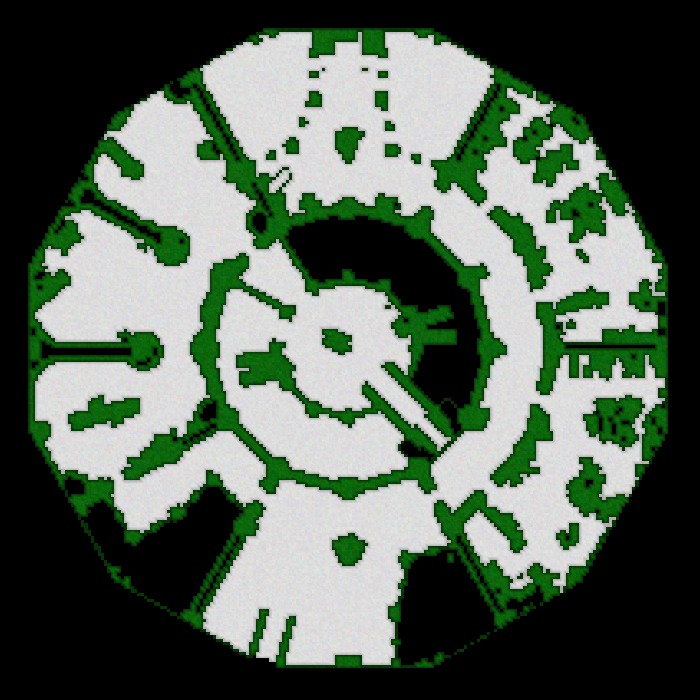}\end{center}
\end{minipage}}
%& 95 & 85 & {\bf91 (85)} & {\bf91 (85)} & 4.75 & {3.22} & {\bf3.02} & 3,006 & 1,714 & {\bf1,662} &  94.04 & 57.23 & {\bf57.13}  \\ \cline{2-14} 
 & 100 & 68 & {\bf77 (68)} & 75 {\bf(68)} & 6.76 & {\bf5.49} & 5.94 & 4,100 & {\bf3,114} & 3,341 &  196.66 & {\bf145.09} & 156.74 \\ \cline{2-14} 
 %& 105  & 64 & {\bf72 (64)} & {\bf72 (64)} & 6.59 & 5.63 & {\bf5.32} & 3,959 & 3,130 & {\bf2,896} &  220.22 & 173.47 & {\bf172.21}\\ \cline{2-14} 
  & 110 & 59 & {\bf67 (59)} & {\bf67 (59)} & 6.58 & 6.03 & {\bf5.92} & 3,978 & 3,652 & {\bf3,596} &  249.89 & 202.61 & {\bf202.39} \\ \cline{2-14} 
% & 115 & 50 & 61 {\bf(50)} & {\bf64 (50)} & 6.99 & 5.66 & {\bf5.64} & 3,864 & 2,713 & {\bf2,705} &  303.50 & 240.79 & {\bf223.41} \\ \cline{2-14} 
 & 120 & 35 & {\bf44 (35)} & {\bf44} (34) & 9.59 & {\bf8.76} & 8.80 & 5,351 & {\bf4,643} & 4,691 &  393.27 & {\bf341.63} & 341.82 \\ \cline{2-14} 
 & 125 & 34 & 41 {\bf(34)} & {\bf42 (34)} & 9.32 & 7.77 & {\bf7.58} & 5,145 & 4,153 & {\bf4,054} &  399.18 & 358.91 & {\bf353.32} \\ \cline{2-14} 
 & 130 & 19 & {\bf26 (19)} & 25 (18) & {\bf4.83} & 5.00 & 4.85 & 2,486 & 2,498 & {\bf2,338} &  487.01 & {\bf447.22} & 453.05  \\ \cline{2-14}  &\multicolumn{4}{|r|}{Improvement over CBSH2} & 0 & {16.6\%} & {\bf17.3\%} & 0 & {22.7\%} & {\bf23.3\%} &0&16.1\%&{\bf16.4\%}\\\hline 
\end{tabular}
\caption{Success rates, average runtimes and CT sizes of instances solved by all solvers and PAR10 scores for different number of agents $k$ in 6 maps. For the success rates of \MLS and \MLO, the fractions of instances solved by both our solver and CBSH2 are given in parentheses (bolded if it solves all instances that CBSH2 solves). For each map, we report the percentage of improvement of our solvers over CBSH2 
on the runtime and CT size on instances solved by all solvers and PAR10 score. \label{thehugetable}
}
\end{table*}

\subsubsection{Features}
We collect a $p$-dimensional feature  vector $\Phi_N(c)$ that describes a conflict $c\in N_{\Conflicts}$ with respect to CT node $N$.  The $p=67$ features of a conflict $\langle a_i,a_j,v,t\rangle$ ($\langle a_i,a_j,u,v,t\rangle$) in our implementation are summarized in Table \ref{featureTable}. They consist of (1) the properties of the conflict, (2) statistics of CT node $N$, the conflicting agents $a_i$ and $a_j$ and the contested vertex or edge w.r.t. the current solution, (3) the frequency of a conflict being resolved for a vertex or an agent, and (4) features of the MDD and the weighted dependency graph. For each feature, we normalize its value to the range $[0,1]$ across all conflicts in $N_{\Conflicts}$. 
All features of a given conflict $c\in N_{\Conflicts}$ can be computed in $O(|N_{\Conflicts}|+k)$ time.
%We use all 67 features in experiments and also identify the important features in Section 6. We
%Thus, the overall time complexity for feature collection at a given node $N$ is $O(|N_{\Conflicts}|(|N_{\Conflicts}|+k))$.

\subsubsection{Labels}
We aim to label each conflict in $N_{\Conflicts}$ such that conflicts with higher ranks determined by the oracle have larger labels. Instead of using the full ranking provided by oracle $O_1$, we use a binary labeling scheme similar to the one proposed by \cite{khalil2016learning}. We assign label 1 to each conflict %with $O_1$ score (the lower bound on the optimal cost in the child nodes) strictly among the top 20\% conflicts in that CT node and label 0 to the rest, with one exception.
strictly among the top 20\% of the full ranking and label 0 to the rest, with one exception. When more than 20\% of the conflicts have the same highest $O_1$ score, we assign label 1 to those conflicts and label 0 to the rest. By doing so, we ensure that at least one conflict is labeled 1 and conflicts with the same score have the same label. 
This labeling scheme relaxes the definition of ``top'' conflicts that allows the learning algorithm to focus on only high-ranking conflicts and helps avoid the irrelevant task of learning the correct ranking of conflicts with low scores.

\subsection{Model Learning}
%Following previous work on learning to branch for MILP solving \cite{khalil2016learning}, 
We learn a linear ranking function with parameters $\bw\in\mathbb{R}^p$
\[
f: \mathbb{R}^p\rightarrow \mathbb{R}: f(\Phi_N(c))=\bw^{\mathsf{T}}\Phi_N(c)
\]
that minimizes the loss function
\[
L(\bw)=\sum_{N\in\calN}l(y_N,\hat{y}_N)+\frac{C}{2}||\bw||_2^2,
\]
where $y_N$ is the ground-truth label vector, $\hat{y}_N$ is the vector of predicted scores resulting from applying $f$ to the feature vectors of every conflict in $N_{\Conflicts}$, $l(\cdot,\cdot)$ is a loss function measuring the difference between the ground truth labels and the predicted scores, and $C>0$ is a regularization parameter. The loss function $l(\cdot,\cdot)$ is based on a pairwise loss that has been used in the literature  \cite{joachims2002optimizing}. Specifically, we consider the set of pairs $\calP_N=\{(c_i,c_j): c_i,c_j\in N_{\Conflicts} \land y_N(c_i)>y_N(c_j))\}$, where $y_N(c)$ is the ground-truth label of conflict $c$ in label vector $y_N$. The loss function $l(\cdot,\cdot)$ is the fraction of swapped pairs, defined as 
\[
l(y_N,\hat{y}_N)=\frac{1}{|\calP_N|}|\{(c_i,c_j)\in \calP_N: \hat{y}_N(c_i)\leq\hat{y}_N(c_j)\}|.
\]
We use an open-source package made available by \cite{joachims2006training} that implements a Support Vector Machine (SVM) approach \cite{joachims2002optimizing} that minimizes an upper bound  on the loss, which is NP-hard to minimize. 
\subsection{ML-Guided Search}

After offline data collection and ranking function $f(\cdot)$ learning, we replace the oracle for conflict selection  in CBS with the learned function. At each CT node $N$, we first compute the feature vector $\Phi_N(c)$ for each conflict $c\in N_{\Conflicts}$ and pick the conflict with the maximum score $c^*=\arg\max_{c\in N_{\Conflicts}}f(\Phi_N(c))$. The overall time complexity for conflict selection at node $N$ is $O(|N_{\Conflicts}|(|N_{\Conflicts}|+k))$. Even though the complexity of conflict selection with oracle $O_0$ is only $O(|N_{\Conflicts}|)$, we will show in our experiments that we are able to outperform CBSH2+$O_0$ in terms of both the CT size and the runtime. 

\begin{figure}[tbp]
	\centering
	%\begin{subfigure}[htbp]{0.24\textwidth}
		%\centering
		%\includegraphics[height=3.6cm]{figures/SuccesRateSmall.png}
		%\caption{Success rates on the small map.\label{smallmapres}}
	%\end{subfigure}
		\centering
		\includegraphics[width=4cm]{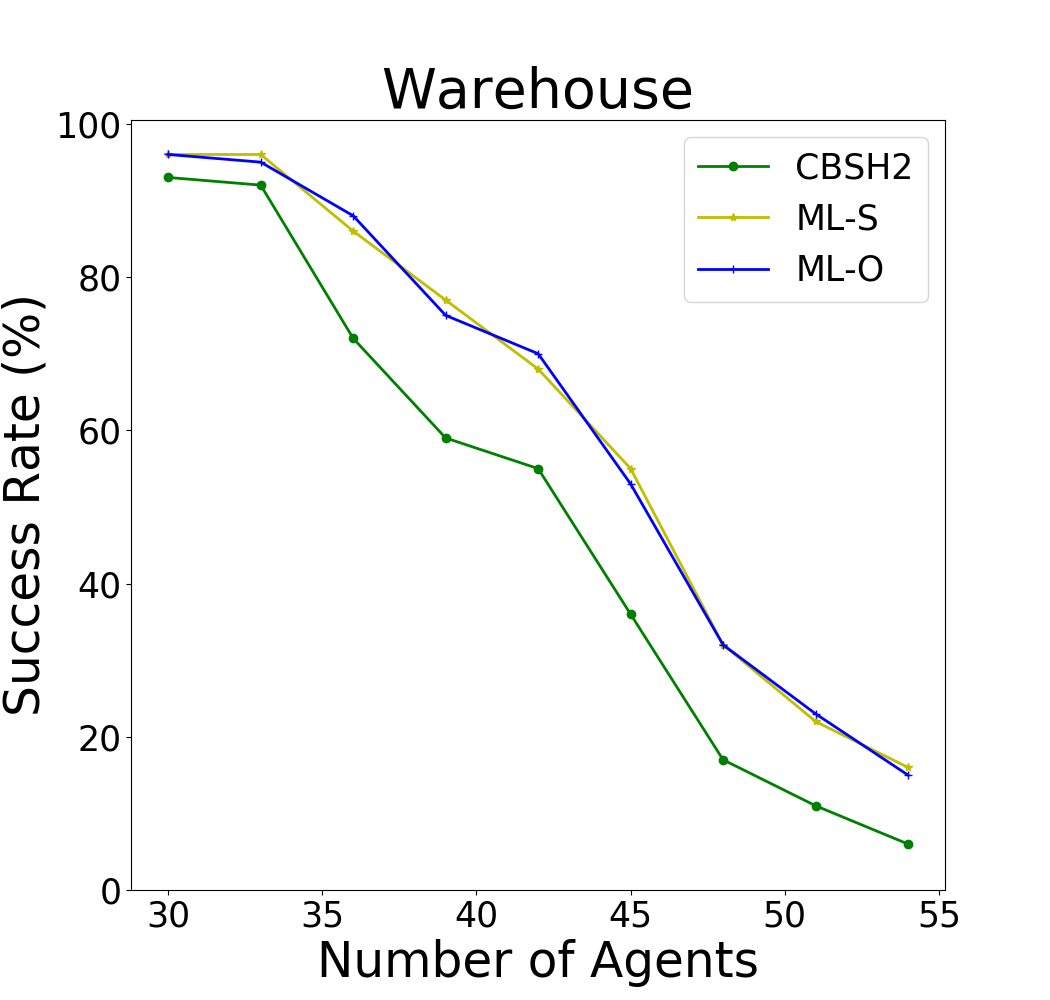}
				\includegraphics[width=4cm]{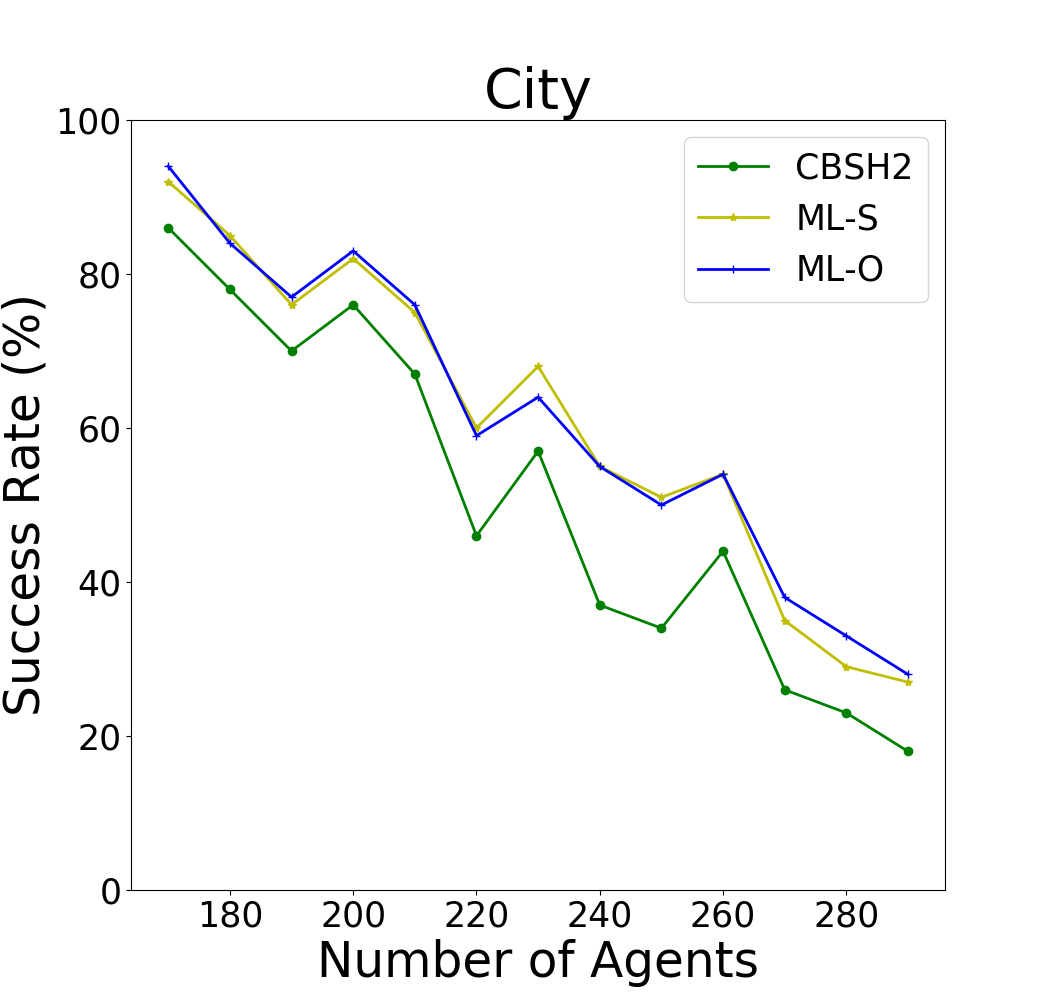}

	\caption{Success rates within the runtime limit.\label{succrates}}
\end{figure}

\section{6\,\,Experimental Results}
In this section, we demonstrate the efficiency and effectiveness of our solver, ML-guided CBS, through extensive experiments.
We use the C++ code for CBSH2 with the WDG heuristic made available by \cite{li2019improved} as our CBS version. We compare against CBSH2+$O_0$ as  baseline since $O_0$ is the most commonly used conflict-selection oracle. The reason why we choose CBSH2 with the WDG heuristic over CBS, ICBS or CBSH2 with the CG or DG heuristics is that it performs best, as demonstrated in \cite{li2019improved}. All reported results are averaged over 100 randomly generated instances. 

Our experiments provide answers to the following questions: i) If the graph underlying the instances is known in advance, can we learn a model that  performs well on unseen instances on the same graph with different numbers of agents? ii) If the graph underlying the instances is unknown, can we learn a model from other graphs that performs well on instances on that graph? %ii) Is there a golden rule for conflict selection that works well in general? iii) Can we relax and change the assumption of fixed underlying graphs to a distribution of graphs that vary in their sizes and obstacle densities?
%Experimentally, we answer Question i) in Sections \ref{sec-small} and \ref{sec-large}, Question ii) in Section \ref{sec-featImportance} and Question iii) in Section \ref{sec-mix}.

We use a set of six four-neighbor grid maps $\calM$ of different sizes and structures as the graphs underlying the instances and evaluate our algorithms on them. $\calM$ includes (1) a warehouse map \cite{li2020new}, a $79\times31$ grid map with 100 $6\times 2$ rectangle obstacles; (2) the room map ``room-32-32-4'' \cite{stern2019multi}, a $32\times 32$ grid map with 64 $3\times 3$ rooms connected by single-cell doors; (3) the maze map ``maze-128-128-2'' \cite{stern2019multi}, a $128\times 128$ grid map with two-cell-wide corridors; (4) the random map; (5) the city map ``Paris\_1\_256'' \cite{stern2019multi}, a $256\times 256$ grid map of Paris; (6) the game map. The figures of the maps are shown in Table \ref{thehugetable}. For each grid map $M\in\calM$, we collect data from randomly generated training instances $\calI^{(M)}_{\Train}$ and test instances $\calI^{(M)}_{\Test}$ on $M$ with a fixed number of agents, where $|\calI^{(M)}_{\Train}|=30$ and $|\calI^{(M)}_{\Test}|=20$. We learn two ranking functions for map $M$: a ranking function that is trained using 5,000 CT nodes i.i.d. sampled from the training dataset collected by solving instances $\calI^{(M)}_{\Train}$ on the same map and another that is trained using 5,000 CT nodes sampled from the training dataset collected by solving instances $\cup_{M'\in\calM}\calI^{(M')}_{\Train}\setminus \calI^{(M)}_{\Train}$ on the other maps, 1,000 i.i.d. CT nodes sampled for each of the five other maps. For each $M\in \calM$, we denote our solver that uses the ranking function trained on the same map as \MLS and the solver that uses the one trained on the other maps as \MLO.  We set $C=1/100$ to train an $SVM^{rank}$ \cite{joachims2002optimizing} with a linear kernel to obtain each of the ranking functions. We varied $C\in\{1/10,1/100,1/1000\}$ and found that \MLS and \MLO perform similarly. We test the learned ranking functions on the test dataset collected by solving $\calI_{\Test}^{(M)}$. The numbers of agents in the instances used for data collection, the test losses and the test accuracies of picking the conflicts labeled as 1 are reported in Table \ref{exppara}. We varied the numbers of agents for data collection and found that they led to similar performances.
In general, the losses of the ranking functions for \MLO are larger and their accuracies of picking ``good'' conflicts are lower that those for \MLS.  %Even though the ranking functions trained on the other maps make fewer good decisions, we will show that they still generalize well and result in conflict selections during the search that are just as competitive as the ranking functions that are trained and tested on the same map.

We run CBSH2, \MLS and \MLO on randomly generated instances on each of the six maps and vary the number of agents. The runtime limits are set to 60 minutes for the two largest maps (the city and game maps) and 10 minutes for the others.
%We consider three metrics for evaluating the performance on the test instances: the success rate (the number of solved instances), the size of the resulting CT and the resulting runtime. 
In Table \ref{thehugetable}, we report the success rates, the average runtimes and CT sizes of instances solved by all solvers and the PAR10 scores (a commonly used metric to score solvers where we count the runs that exceed the given runtime limit as 10 times the limit when computing the average runtimes) \cite{bischl2016aslib} for some numbers of agents on each map and defer the full table to Appendix A.  We plot the success rates on the warehouse and city maps in Figure \ref{succrates} and the rest in Appendix A.
\MLS and \MLO dominate CBSH2 in all metrics on all maps for almost all cases. %, except that CBSH2 ties \MLS and is 1\% better than \MLO on the random map with 18 agents.
Overall, CBSH2, \MLS and \MLO solve 3,326 (55.43\%), 3,779 (62.98\%) and 3,758 (62.63\%) instances out of 6,000 we tested, respectively.
The improvement of \MLS and \MLO over CBSH2 on instances commonly solved by all solvers is 10.3\% to 64.4\% for the runtime and 13.0\% to 68.2\% for the CT sizes across different maps. 
For \MLS, even though we learn the ranking function from data collected on instances with a fixed number of agents, the learned function generalizes to instances with larger numbers of agents on the same map and outperforms CBSH2.
\MLO, without seeing the actual map being tested on during training, is competitive with \MLS and even outperforms \MLS sometimes on the warehouse, city, maze and game maps. The results suggest that our method, when focusing on solving instances on a particular grid map, can outperform CBSH2 significantly and, when faced with a new grid map, will still gain an advantage. To demonstrate the efficiency of our solver further, we show the success rates for different runtime limits varying the number of agents on each map in Appendix A. Typically, the three solvers tie on the easy instances but \MLS and \MLO gain an advantage on the hard ones, even more for larger numbers of agents.

Next, we look at the feature importance of the learned ranking functions. For \MLO, the six ranking functions have nine features in common among their eleven features with the largest absolute weights. Thus, they are similar when looking at the important features. We take the average of each weight and sort them in decreasing order of their absolute values. The plot and the full list of features with their indices are included in Appendix B. %We plot the top ten features with the largest average weights for \MLS and \MLO in Figure \ref{}. 
%The five top-weighted features for \MLS are (1) the binary indicator for non-cardinal conflicts; (2) the maximum of the differences of the cost of the path of agent $a_i$ ($a_j$) and $t$; (3) weight of the edge between agents $a_i$ and $a_j$ in the weighted dependency graph; (4) the binary indicator for cardinal conflicts; (5) the minimum of the numbers of conflicts involving agent $a_i$  ($a_j$) that have been selected and resolved.
The top eight %features' weights are $[0.858, -0.833, 0.776, 0.540, -0.539]$ and those 
features are (1) the weight of the edge between agents $a_i$ and $a_j$ in the weighted dependency graph (WDG) (feature 67); (2) the binary indicator for non-cardinal conflicts (feature 5); (3) the maximum of the differences of the cost of the path of agent $a_i$ ($a_j$) and $t$ (feature 23); (4)  the binary indicator for cardinal conflict (feature 3); (5) the minimum of the numbers of conflicts that agent $a_i$ ($a_j$) is involved in (feature 12); (6-8) the minimum (feature 6), the maximum (feature 7) and the sum (feature 8) of the numbers of conflicts involving agent $a_i$ ($a_j$) that have been selected and resolved. 
Those features mainly belong to three categories: features related to the cardinal conflicts (features 3 and 5), the WDG (feature 67) and the frequency of a conflict being resolved for an agent (features 6, 7 and 8), where the first is commonly used in previous work on CBS and the third is an analogue of the branching variable pseudocosts \cite{achterberg2005branching} in MILP solving.
For \MLS, we plot the weights of the features in decreasing order of their absolute values for each of the six ranking functions in Appendix B. 
For the random map and the room map, features 5, 67 and 3 come as the top three features. Among the top five features for the two largest maps (the city and game maps), three of them are features 6, 7 and 8. %while the rest of two are features 3 and 5 related to the cardinal conflicts. 
For the maze map, the top feature is the maximum of the widths of level $t-2$ of the MDD for agent $a_i$ ($a_j$), followed by features 67 and 23. For the warehouse map, the top three features are features 6, 23 and 8. As can be seen, most of the important features for each individual map also belong to the three categories. In Appendix B, we present the results for feature selection. We show that we are able to achieve slightly better results with certain combinations of the important features.

\section{7\,\,Conclusions and Future Directions}
In this paper, we proposed the first ML framework for conflict selection in CBS. The extensive experimental results showed that our learned ranking function can generalize across different numbers of agents on a fixed graph (map) or unseen graphs. Our objective was to imitate the decisions made by the oracle that picks the conflict that produces the tightest lower bound on the optimal cost in its child nodes. We are also interested in discovering a better oracle for conflict selection from which we can learn. %It is future work to conduct feature selection to see if we are still able to achieve good performance when reducing the features to those important ones for learning. 
We expect our method to work well with other newly-developed techniques, such as symmetry breaking techniques \cite{li2020new}, and it remains future work to incorporate those techniques into the framework of CBSH2 to work with our ML-guided conflict selection.

\section*{Acknowledgments}
We thank Jiaoyang Li, Peter J. Stuckey and Danial Harabor for helpful discussions.
The research at the University of Southern California was supported by the National Science Foundation (NSF) under grant numbers 1409987, 1724392, 1817189, 1837779 and 1935712 as well as a gift from Amazon.

\bibliographystyle{aaai21.bst}
\bibliography{MLguidedCBS}

\newpage
\appendix
\section*{ \LARGE Appendix:\\ Learning to Resolve Conflicts for Multi-Agent Path Finding with Conflict-Based Search}
\bigskip

\section{A\,\,Additional Experimental Results}

\begin{figure*}[htbp]
	\centering
	%\begin{subfigure}[htbp]{0.24\textwidth}
		%\centering
		%\includegraphics[height=3.6cm]{figures/SuccesRateSmall.png}
		%\caption{Success rates on the small map.\label{smallmapres}}
	%\end{subfigure}
		\begin{subfigure}[htbp]{0.99\textwidth}
		\centering
		\includegraphics[width=3.6cm]{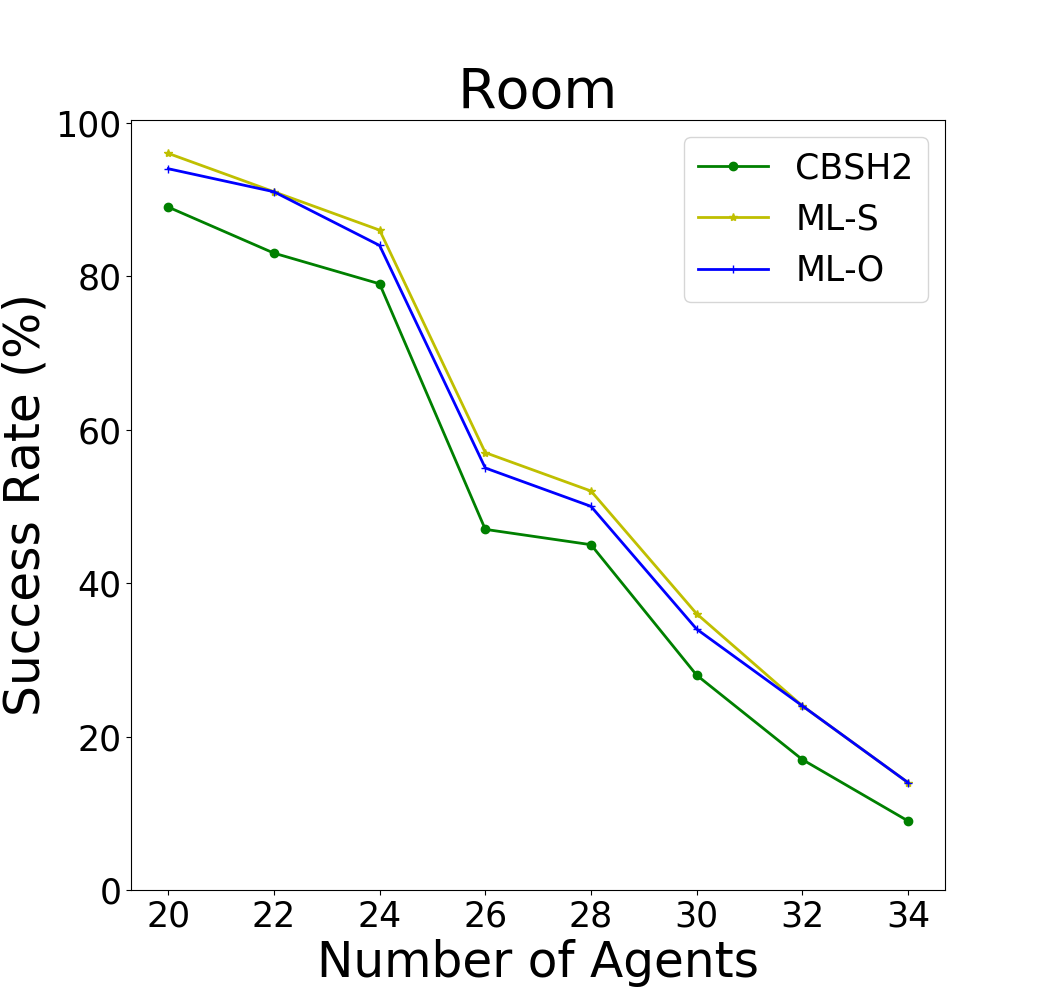}
		\includegraphics[width=3.7cm]{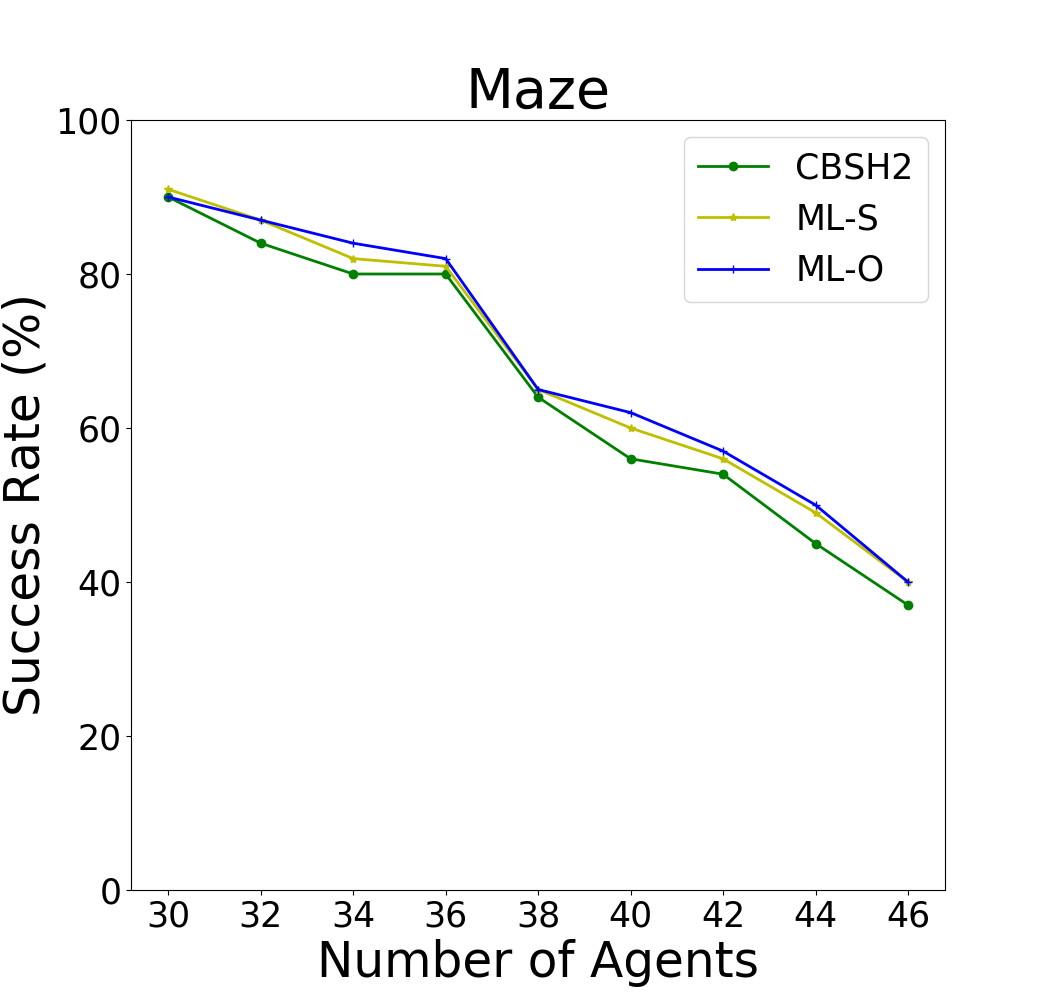}
		\includegraphics[width=3.7cm]{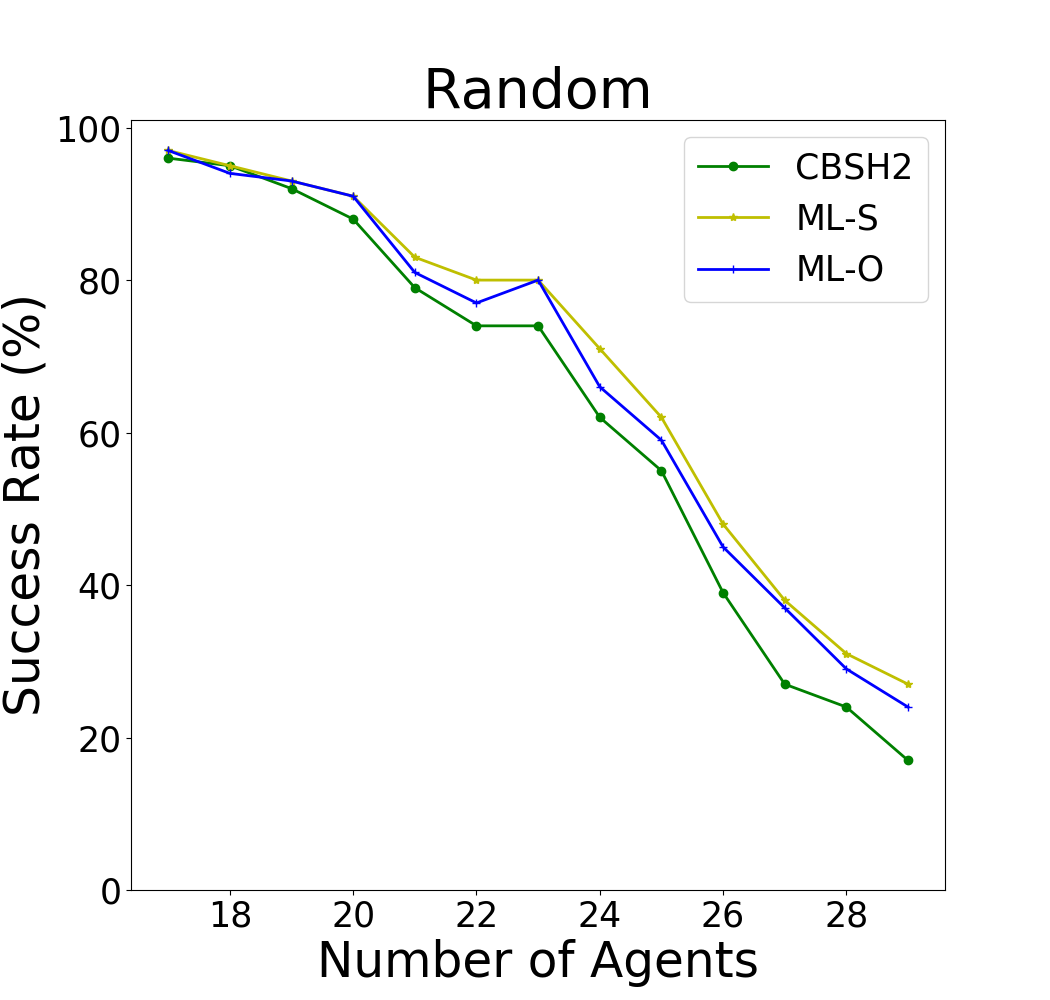}
		\includegraphics[width=3.7cm]{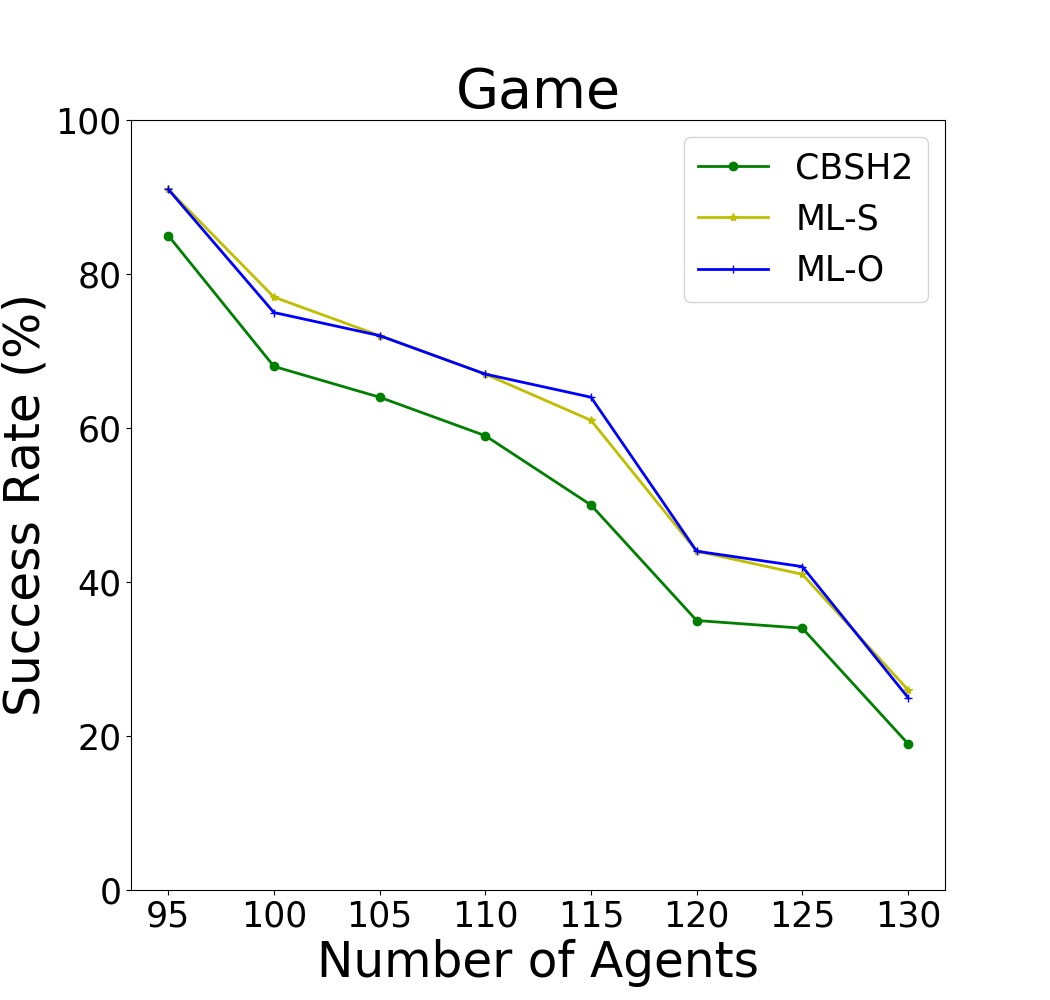}
	\end{subfigure}

	\caption{Success rates within the runtime limit.\label{succratesApp}}
\end{figure*}

\begin{figure*}[htbp]
	\centering
	%\begin{subfigure}[htbp]{0.24\textwidth}
		%\centering
		%\includegraphics[height=3.6cm]{figures/SuccesRateSmall.png}
		%\caption{Success rates on the small map.\label{smallmapres}}
	%\end{subfigure}
		\begin{subfigure}[htbp]{0.99\textwidth}
		\centering
		\includegraphics[width=5cm]{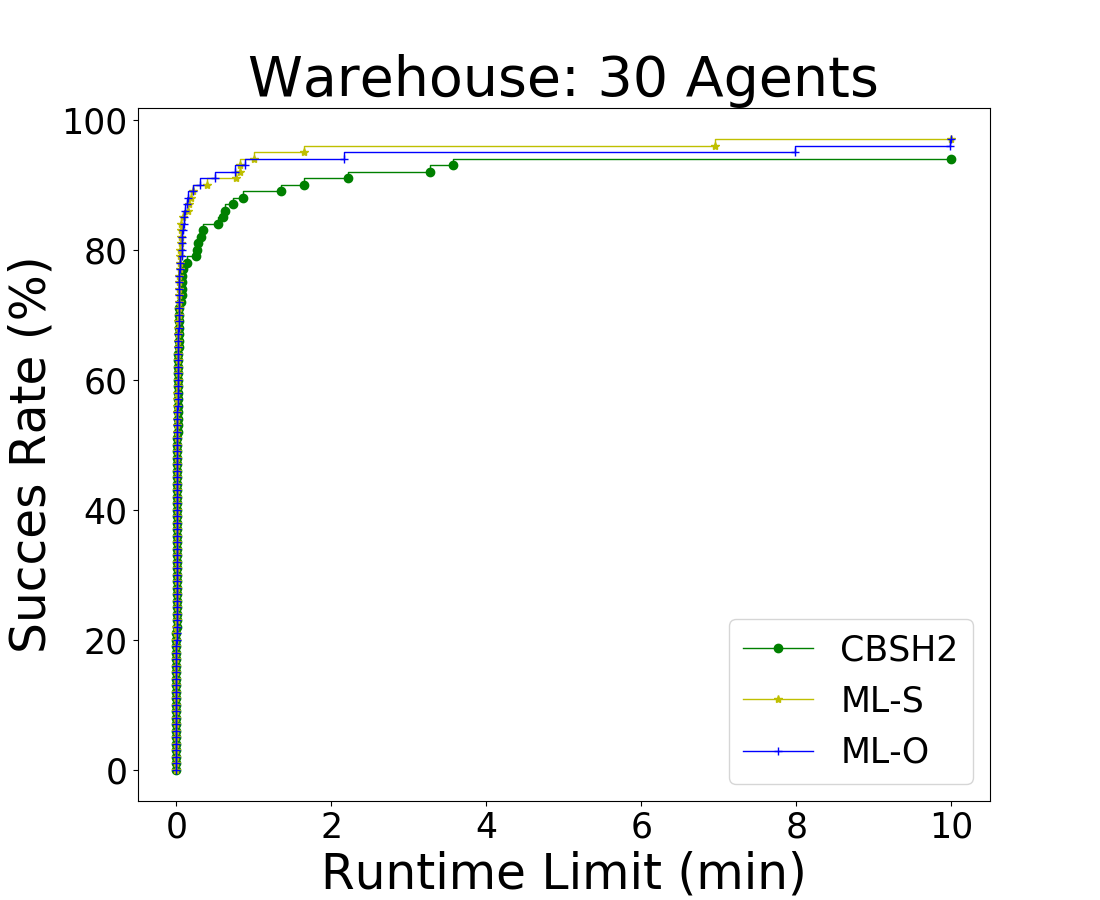}
		\includegraphics[width=5cm]{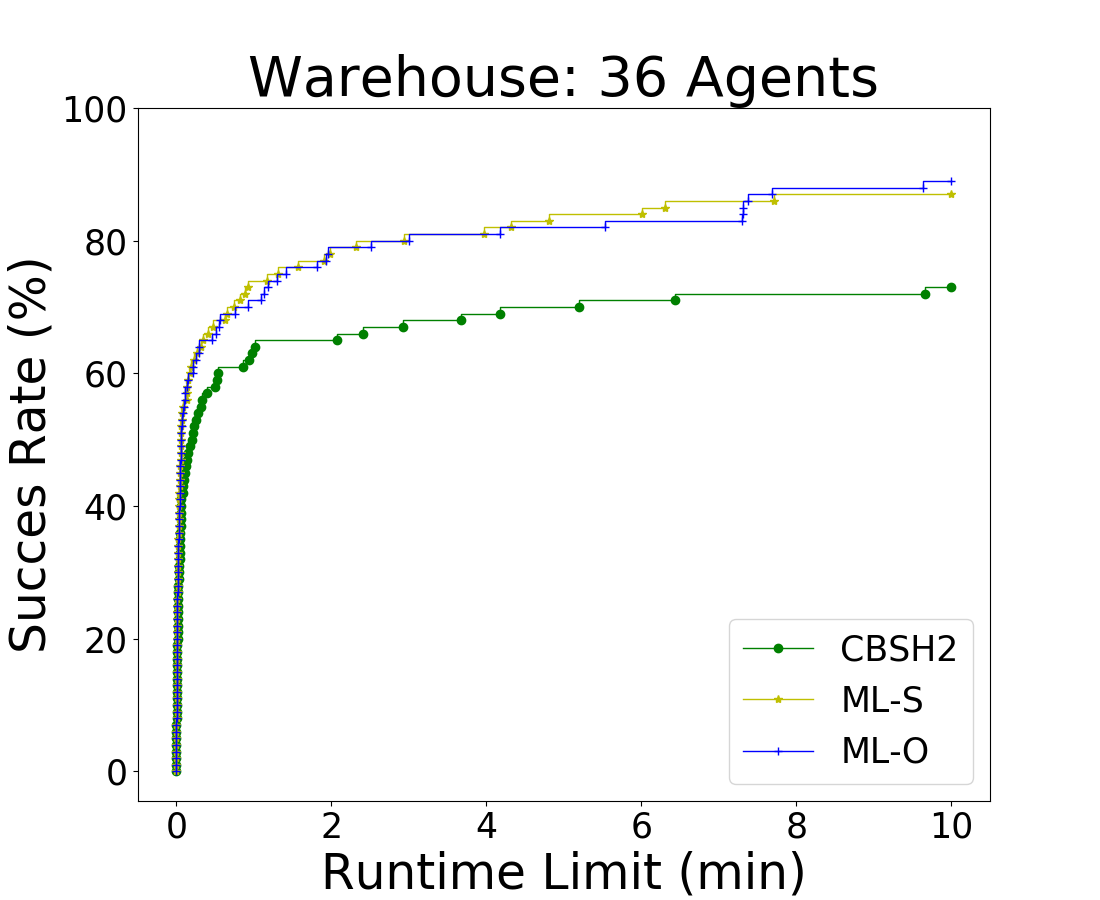}
		\includegraphics[width=5cm]{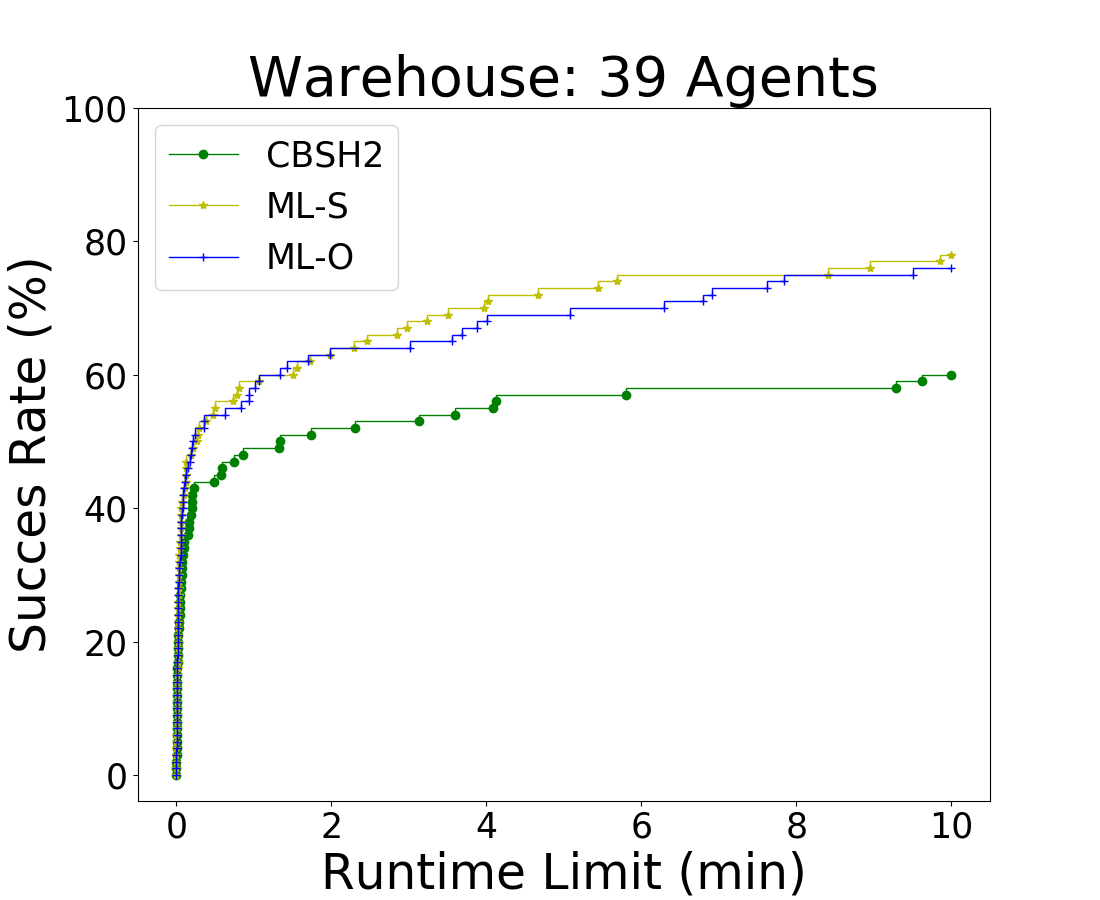}
	\end{subfigure}
	
		\begin{subfigure}[htbp]{0.99\textwidth}
		\centering
		\includegraphics[width=5cm]{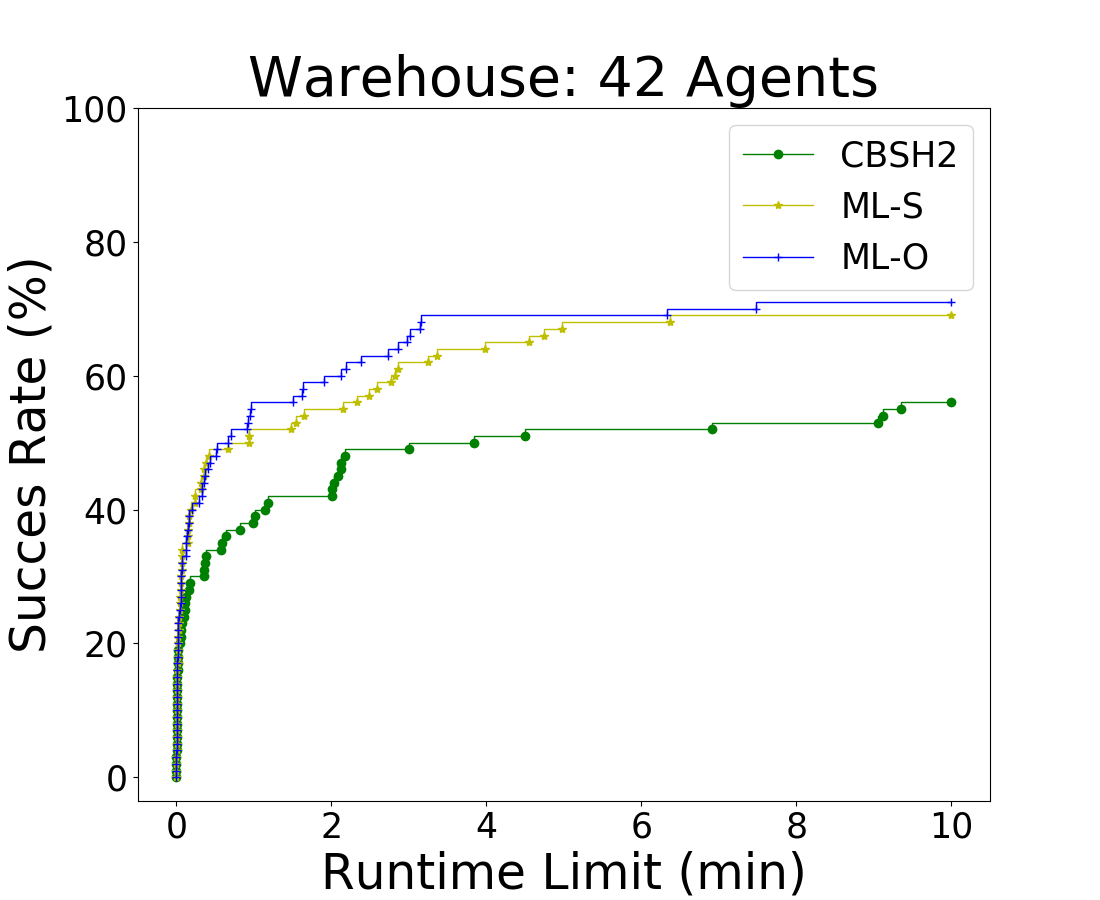}
		\includegraphics[width=5cm]{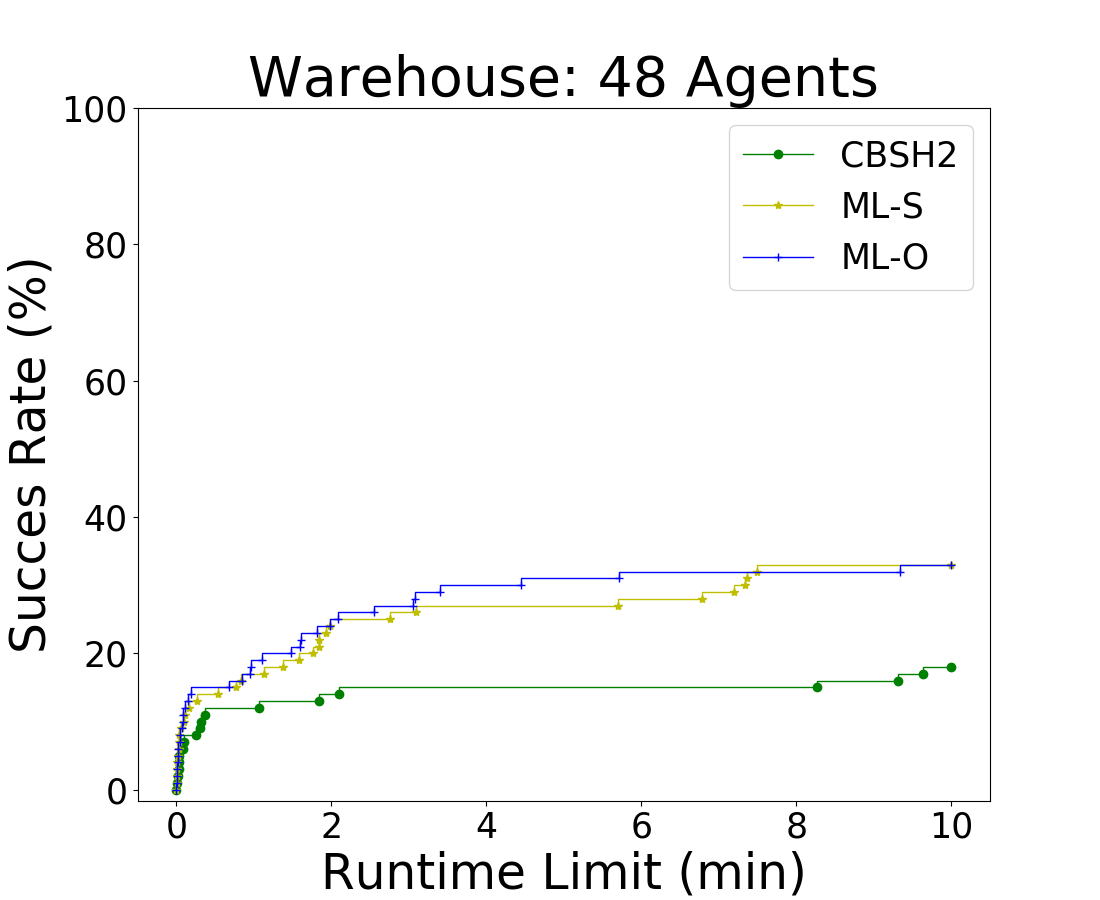}
		\includegraphics[width=5cm]{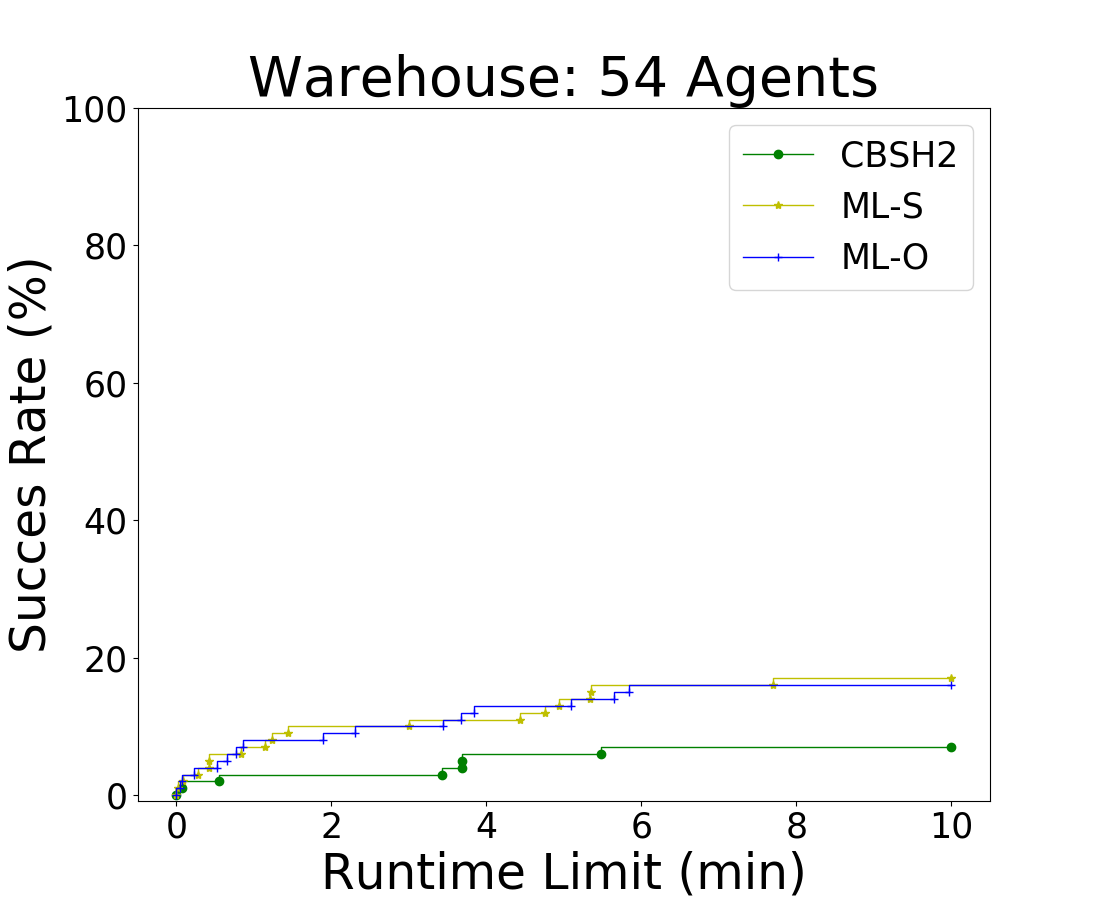}
	\end{subfigure}
	
	\caption{The warehouse map: Percentage of solved instances under problem parameters.\label{cutoffWarehouse}}
\end{figure*}

\begin{figure*}[htbp]
	\centering
		\begin{subfigure}[htbp]{0.99\textwidth}
		\centering
		\includegraphics[width=5cm]{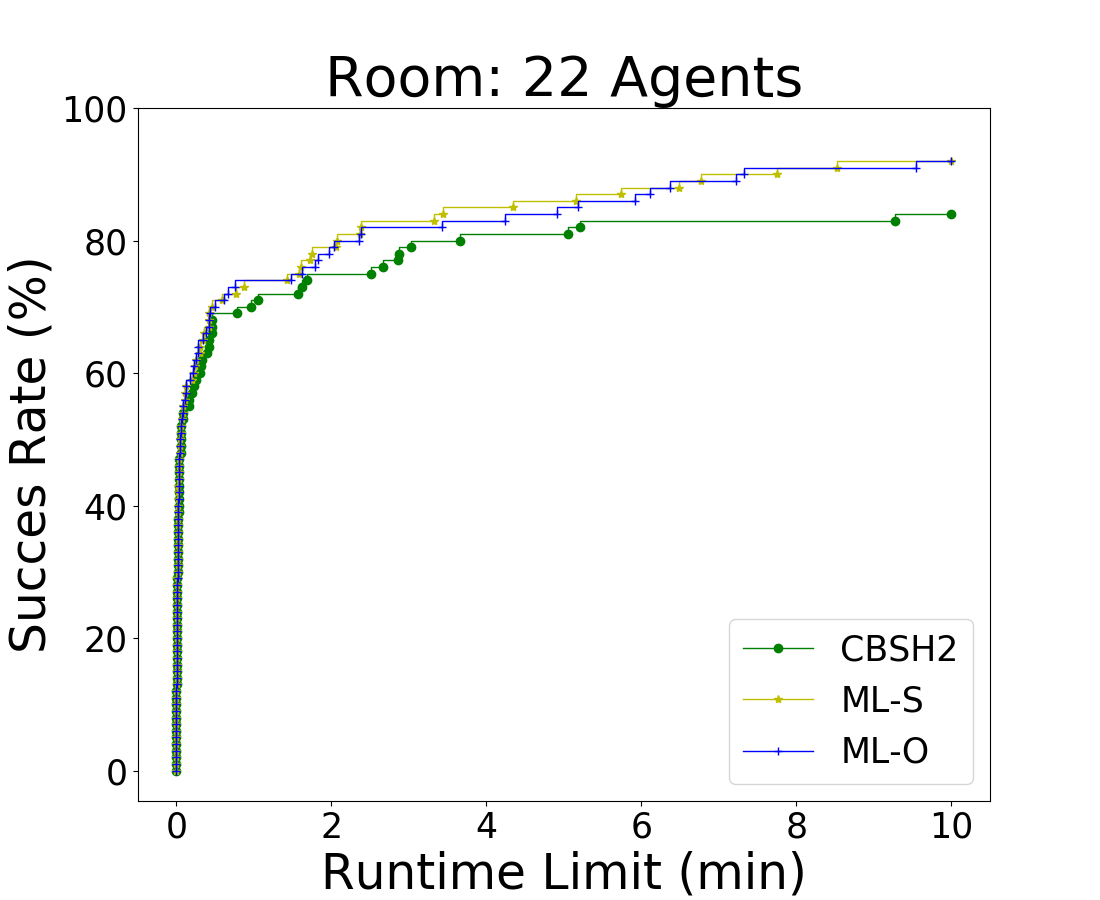}
		\includegraphics[width=5cm]{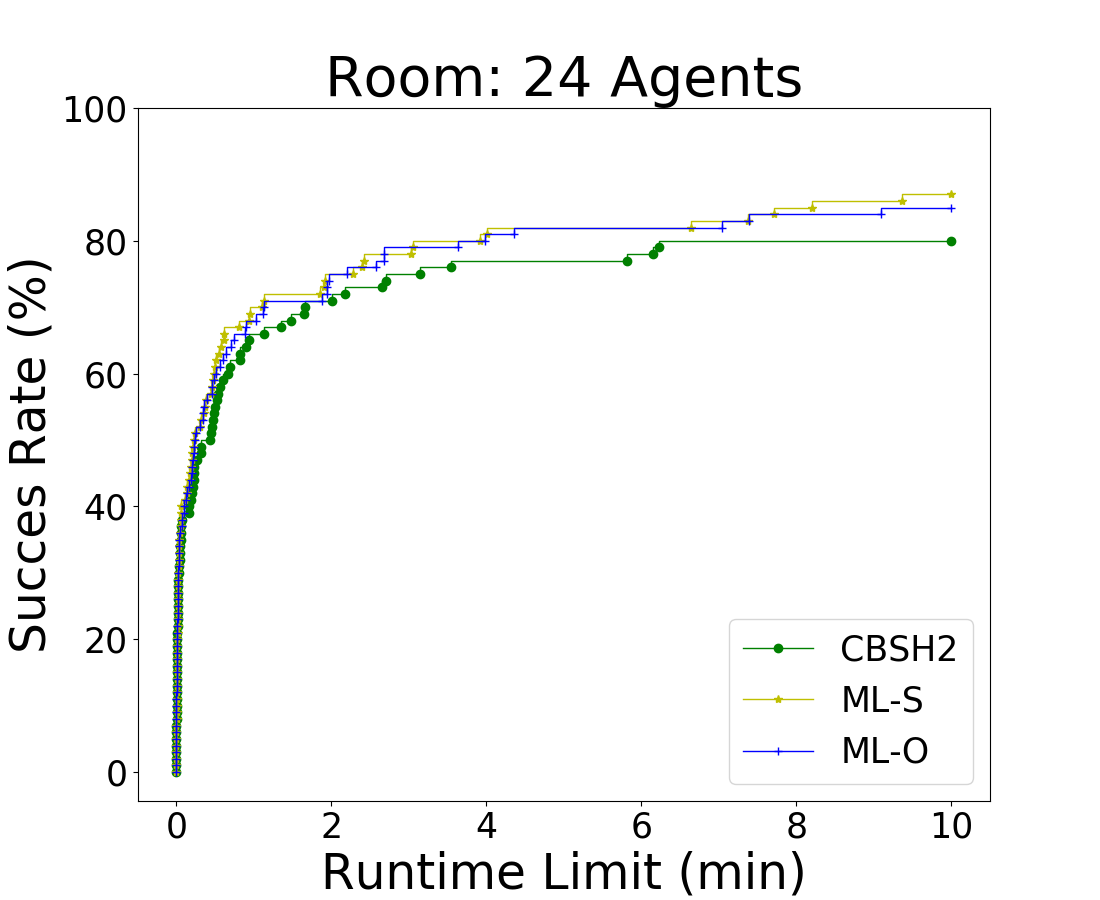}
		\includegraphics[width=5cm]{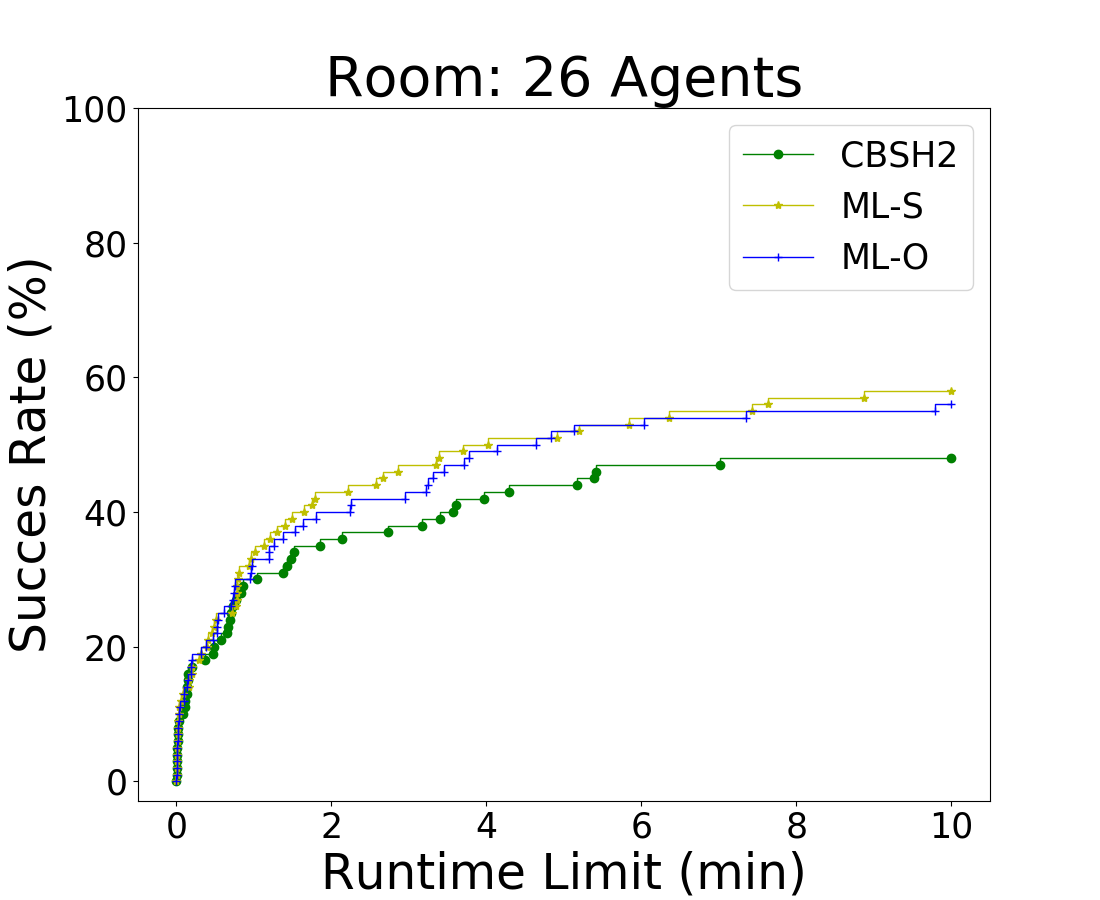}
	\end{subfigure}
	
		\begin{subfigure}[htbp]{0.99\textwidth}
		\centering
		\includegraphics[width=5cm]{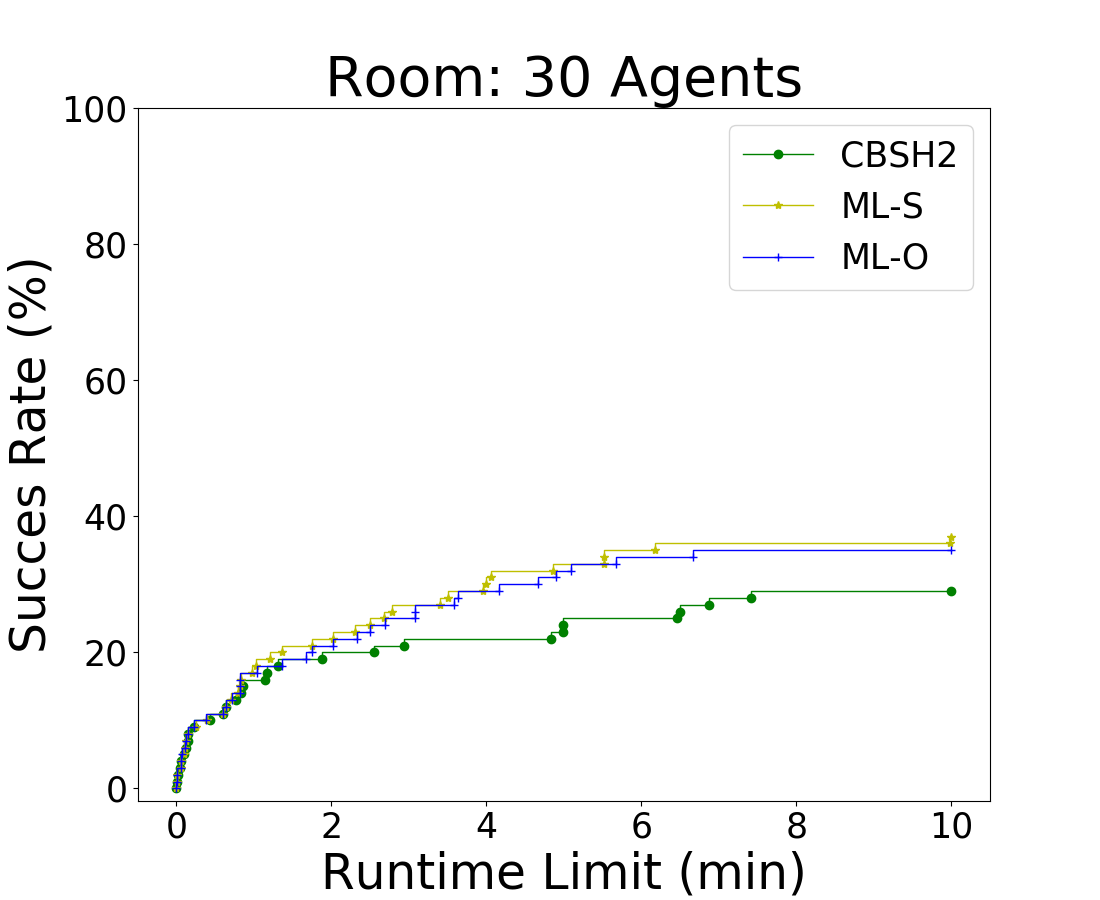}
		\includegraphics[width=5cm]{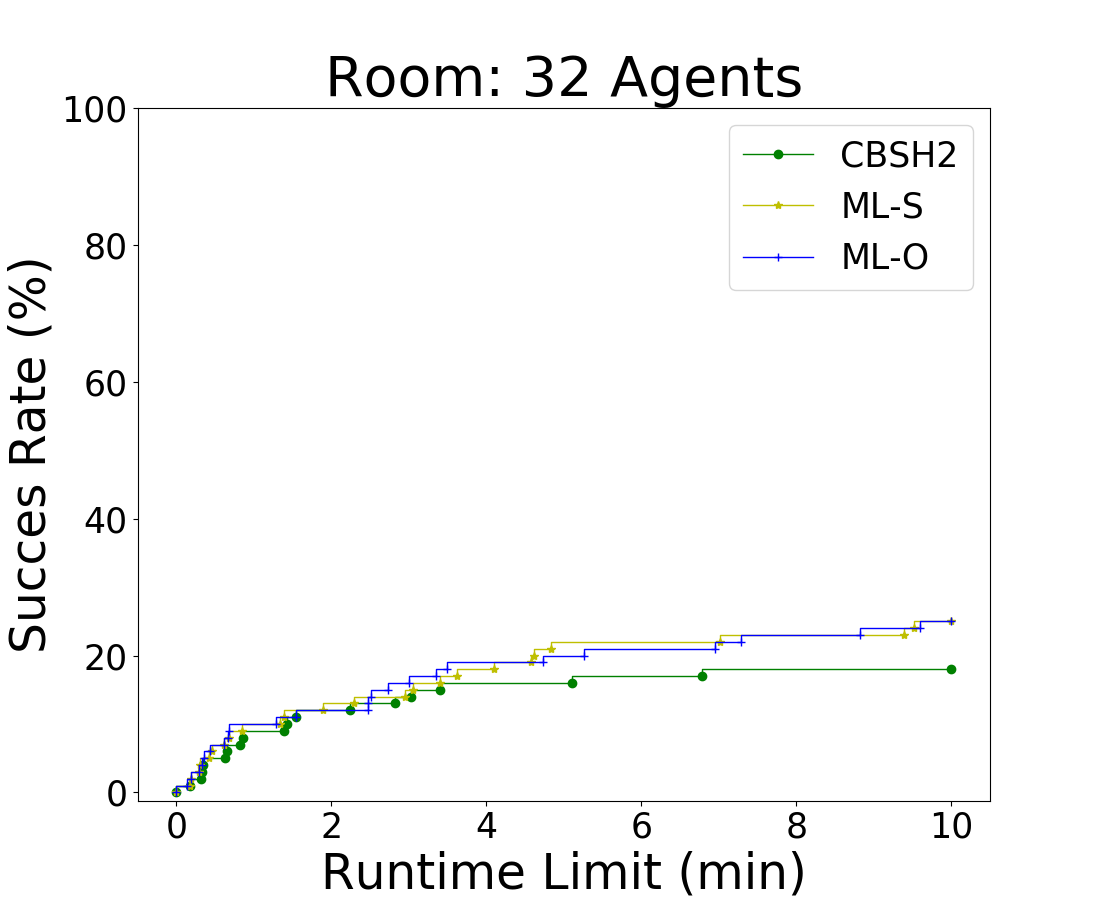}
		\includegraphics[width=5cm]{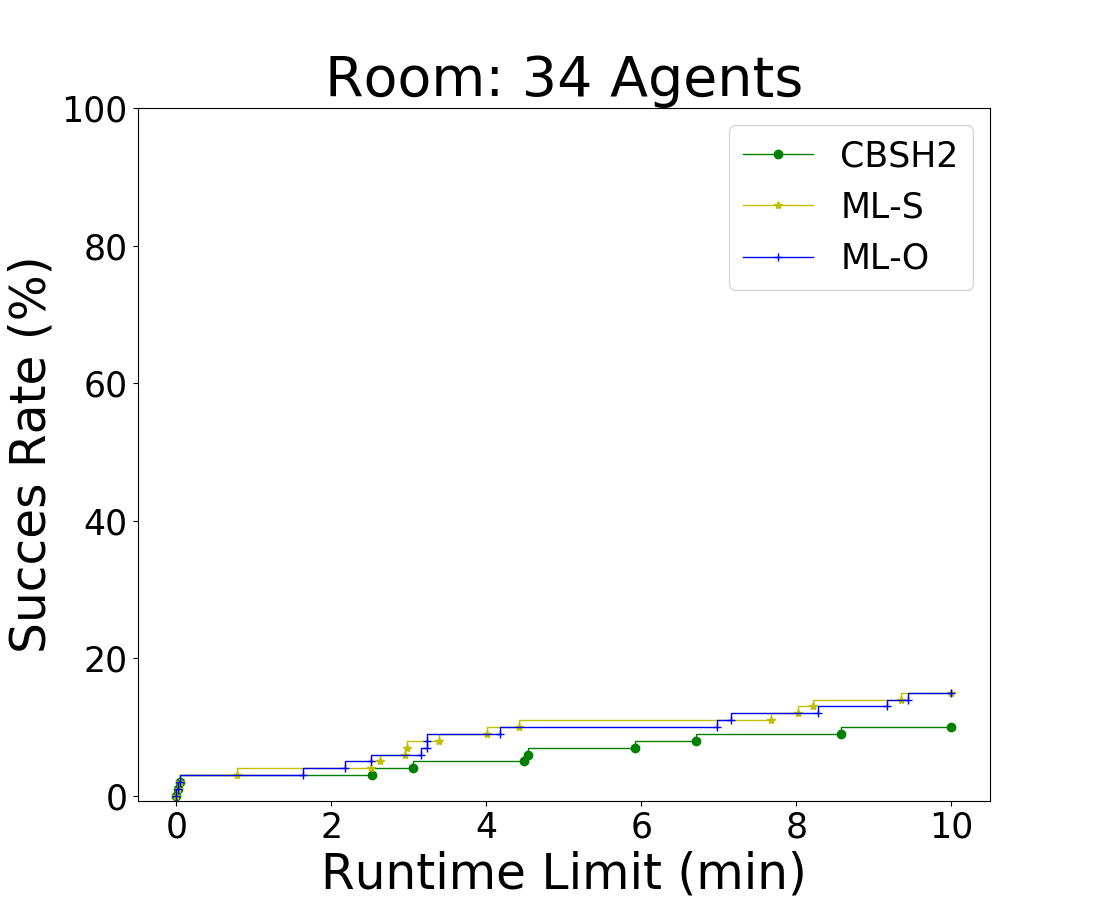}
	\end{subfigure}
	
	\caption{The room map: Percentage of solved instances under problem parameters.\label{cutoffRoom}}
\end{figure*}

\begin{figure*}[htbp]
	\centering
		\begin{subfigure}[htbp]{0.99\textwidth}
		\centering
		\includegraphics[width=5cm]{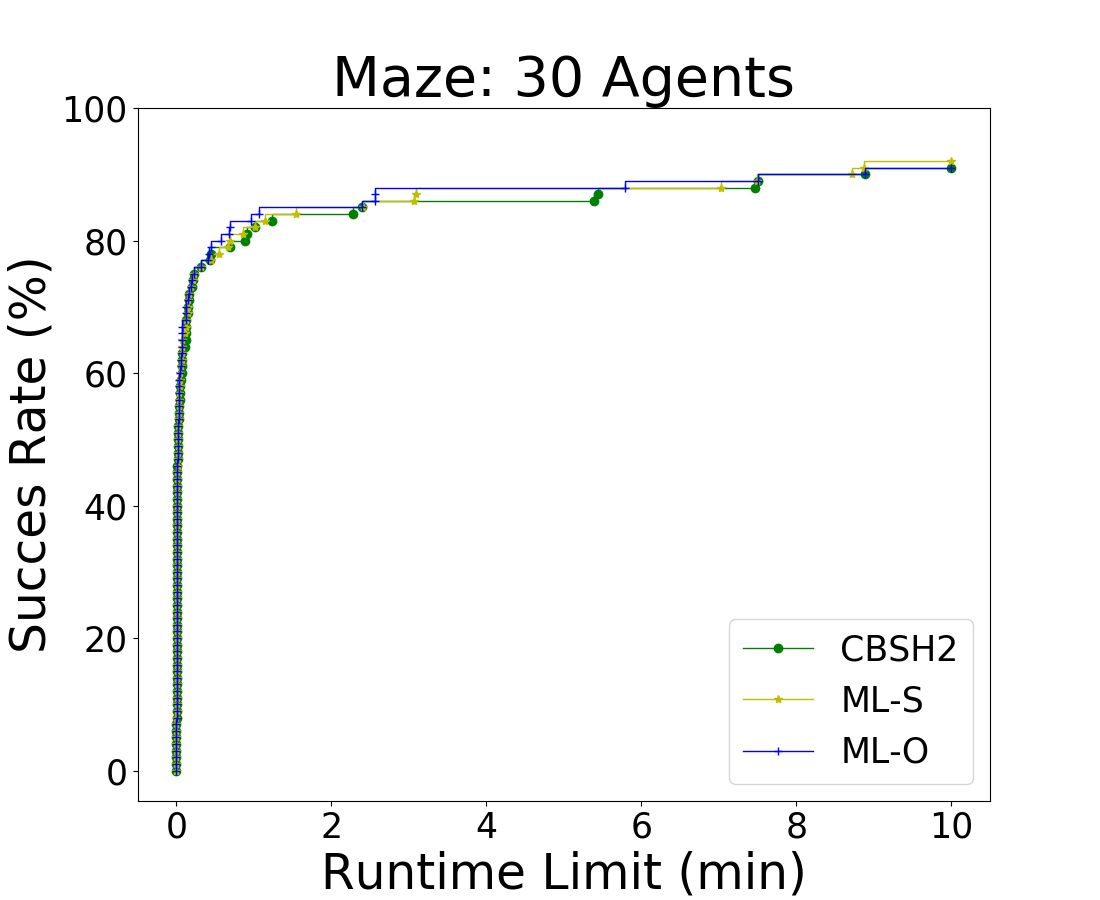}
		\includegraphics[width=5cm]{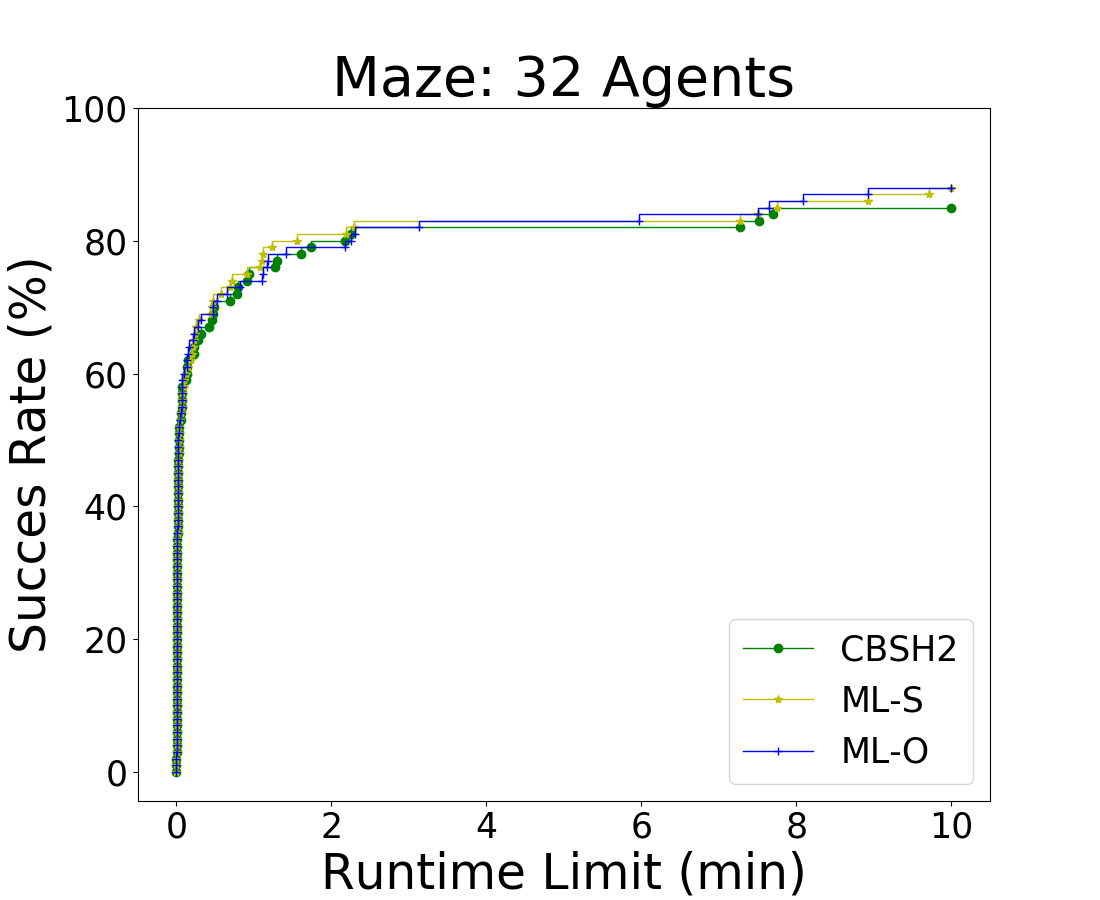}
		\includegraphics[width=5cm]{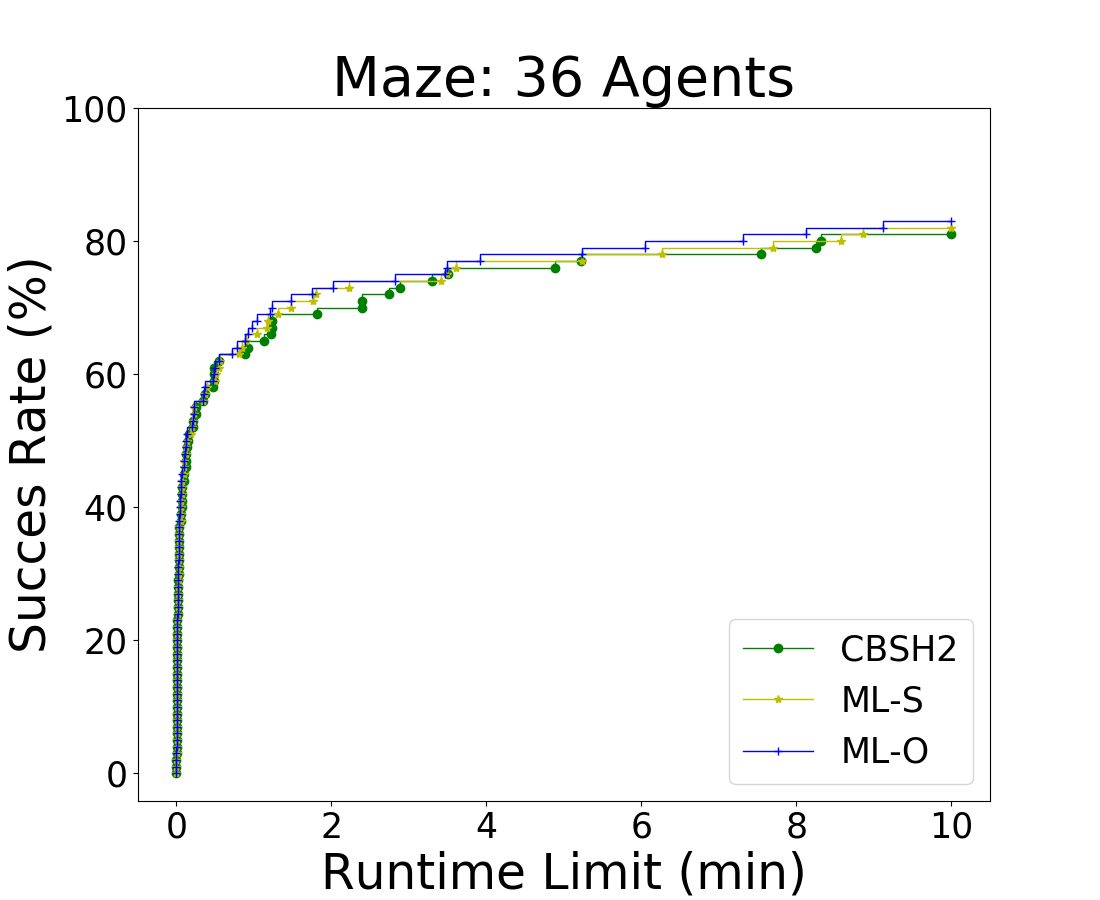}
	\end{subfigure}
	
		\begin{subfigure}[htbp]{0.99\textwidth}
		\centering
		\includegraphics[width=5cm]{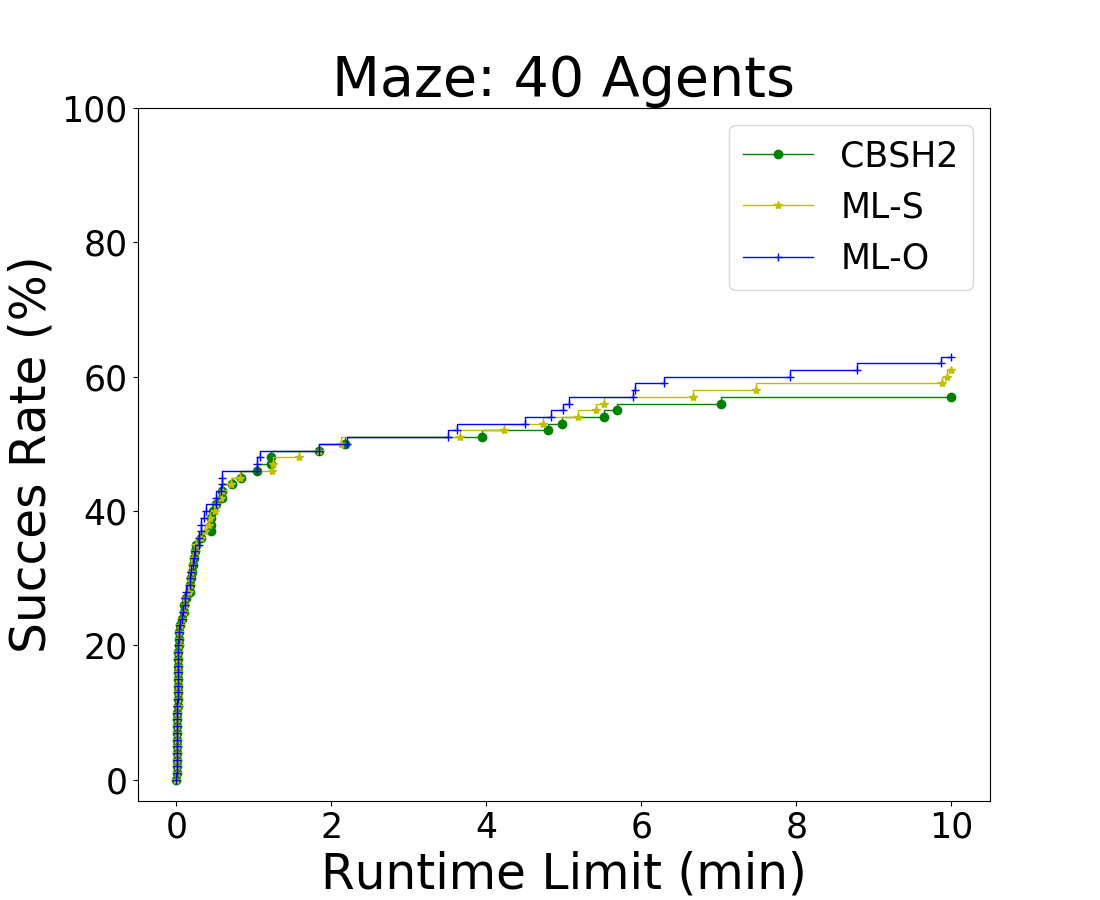}
		\includegraphics[width=5cm]{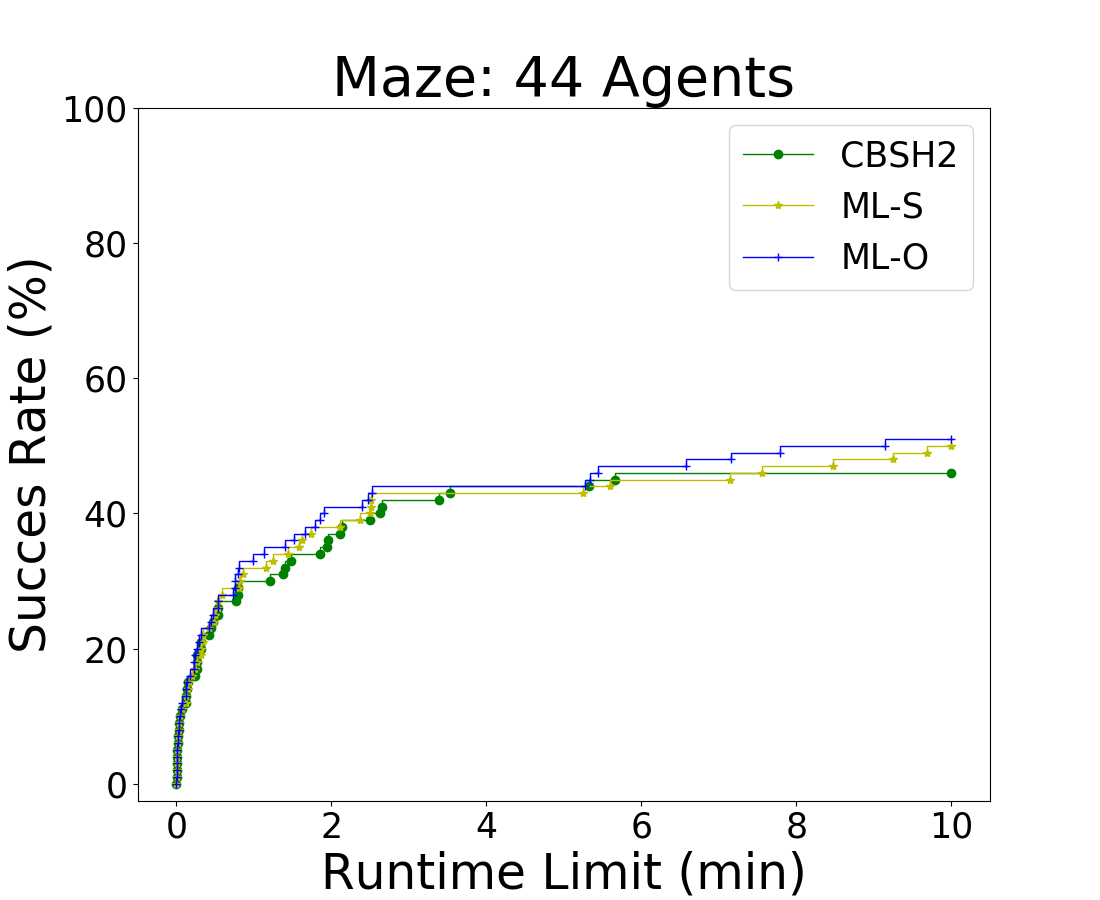}
		\includegraphics[width=5cm]{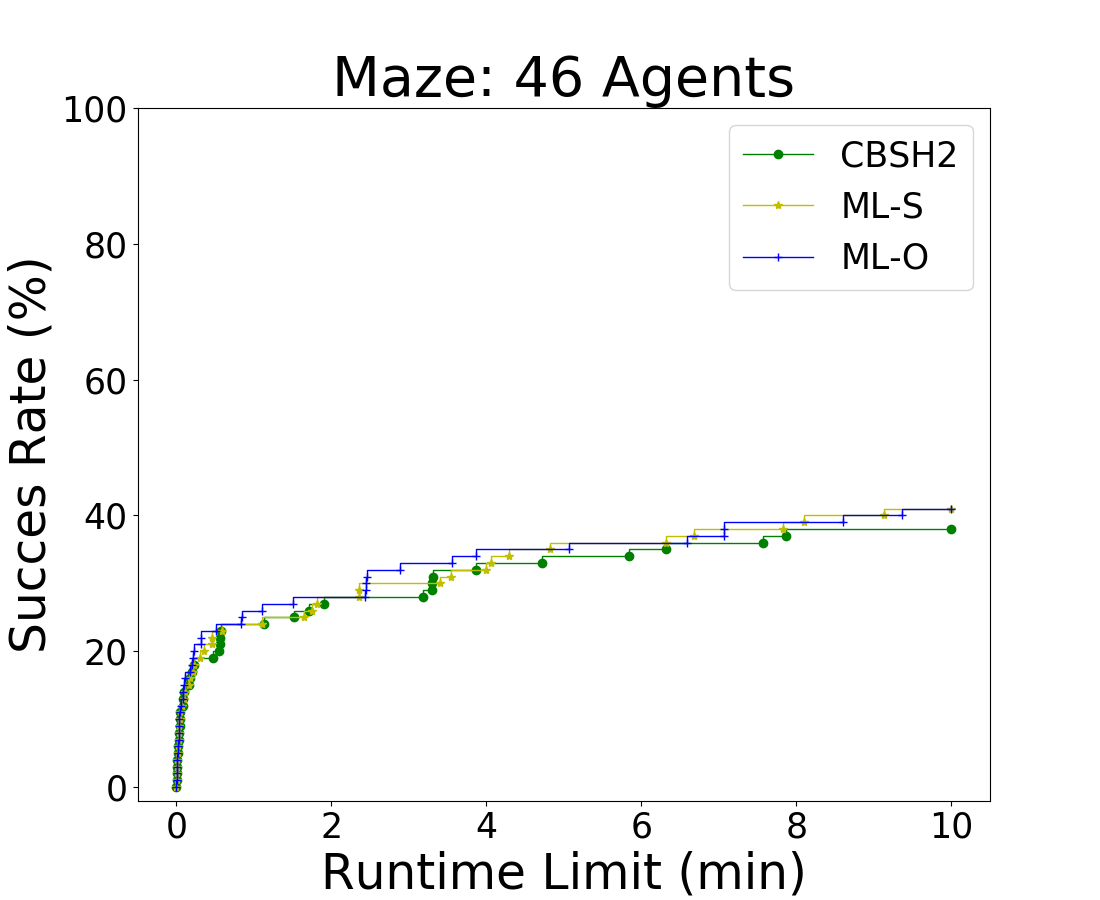}
	\end{subfigure}
	
	\caption{The maze map: Percentage of solved instances under problem parameters.\label{cutoffMaze}}
\end{figure*}

\begin{figure*}[htbp]
	\centering
		\begin{subfigure}[htbp]{0.99\textwidth}
		\centering
		\includegraphics[width=5cm]{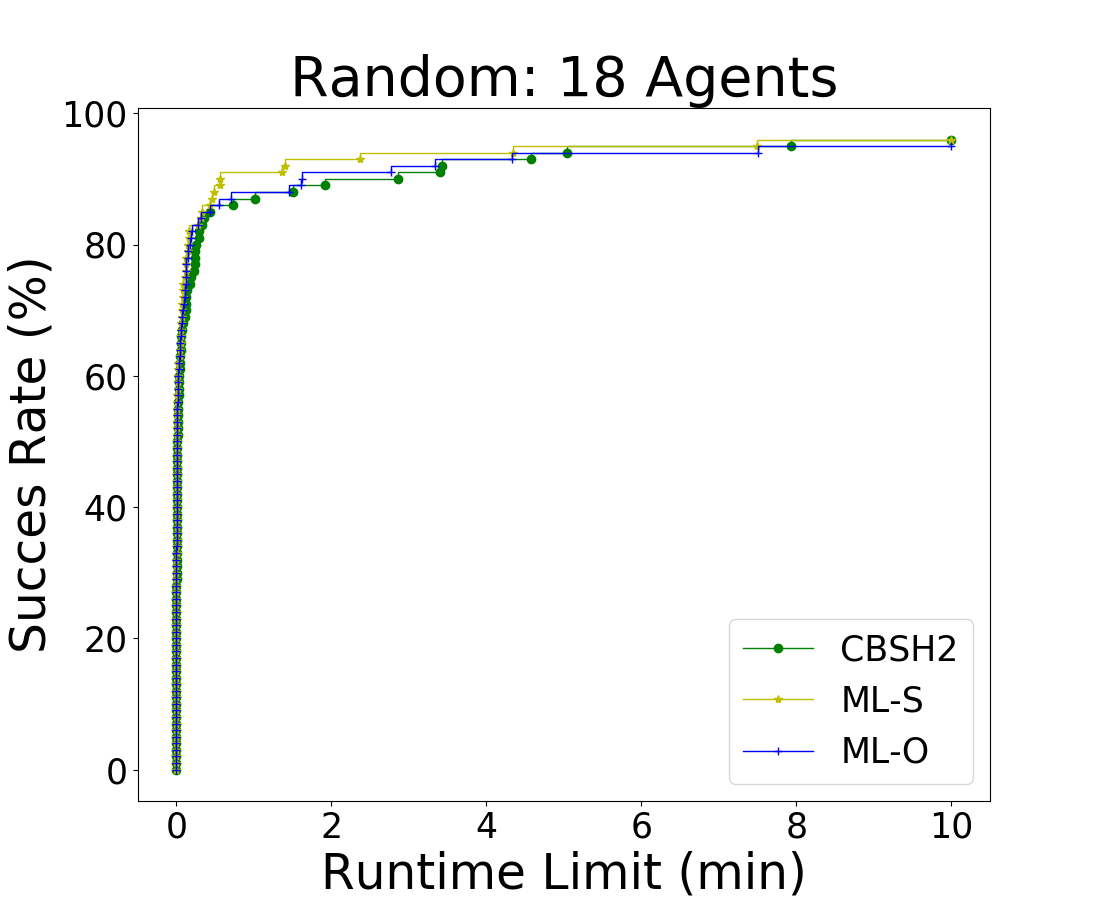}
		\includegraphics[width=5cm]{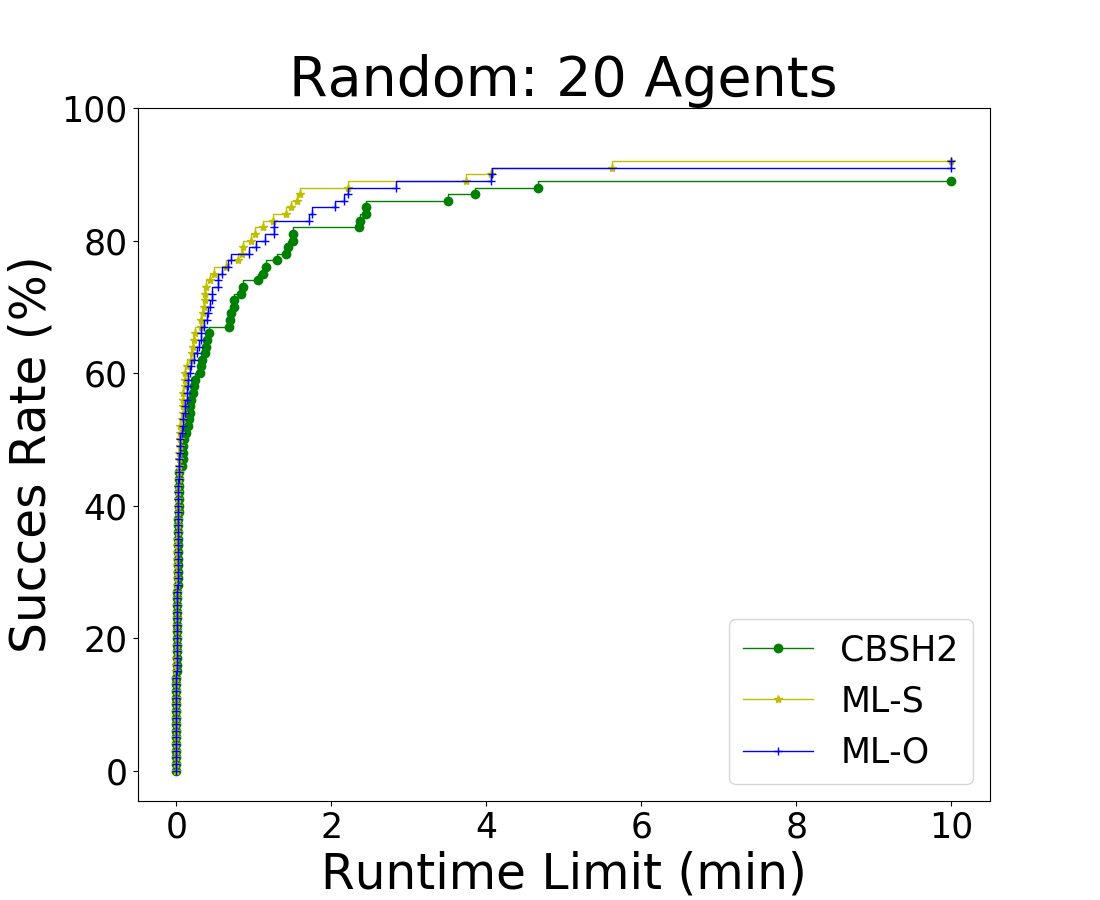}
		\includegraphics[width=5cm]{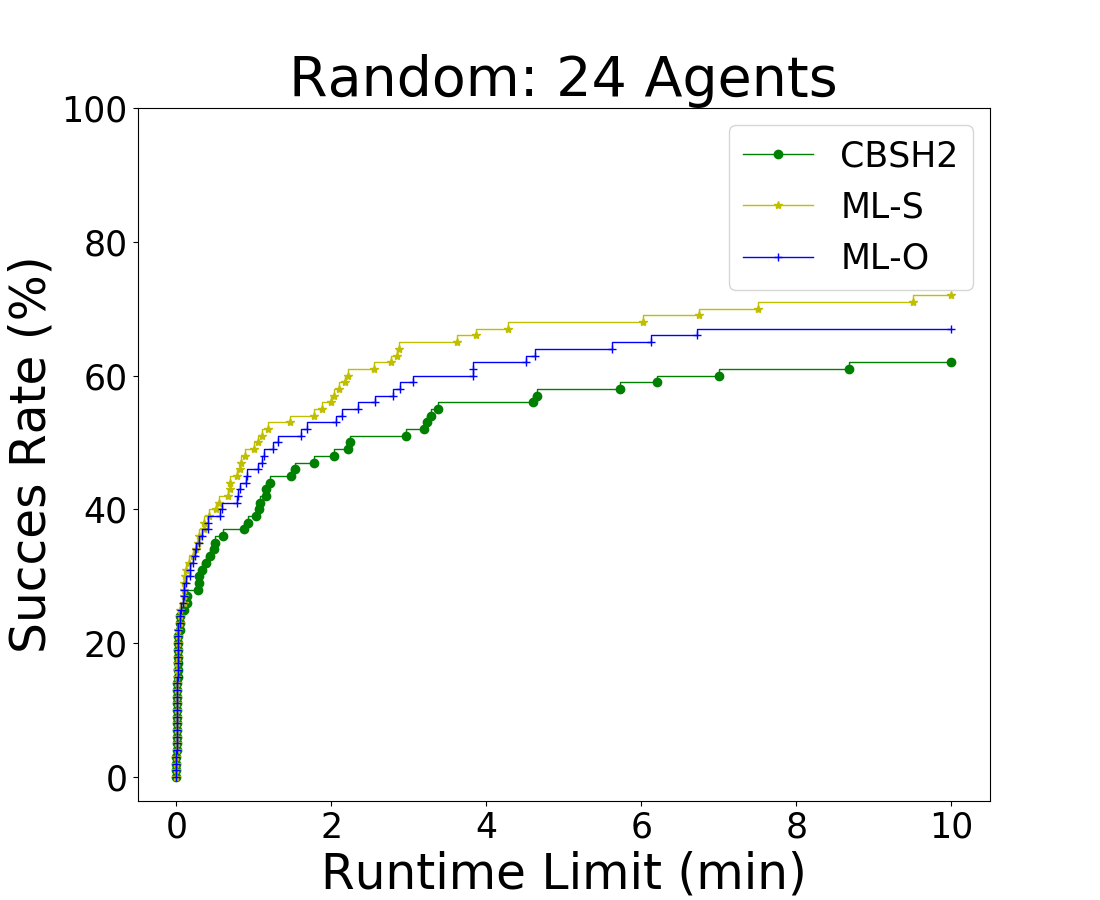}
	\end{subfigure}
	
		\begin{subfigure}[htbp]{0.99\textwidth}
		\centering
		\includegraphics[width=5cm]{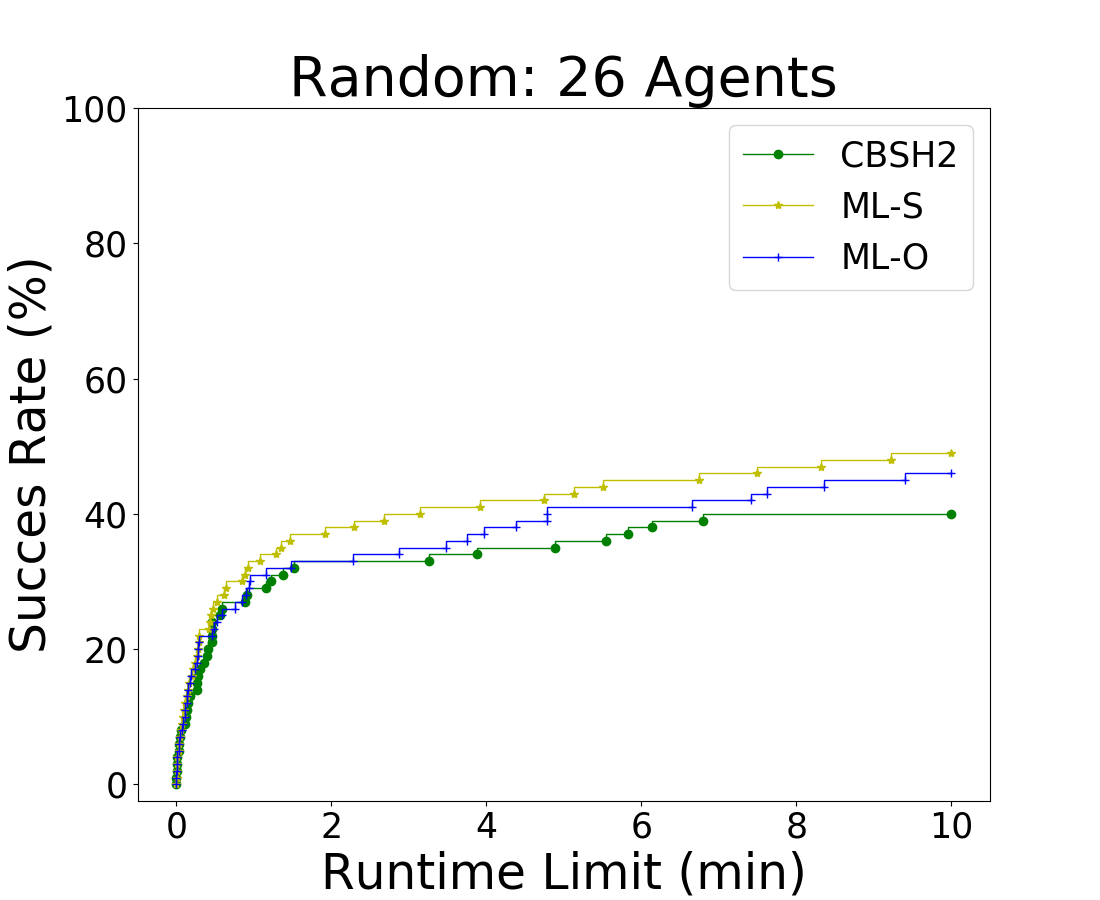}
		\includegraphics[width=5cm]{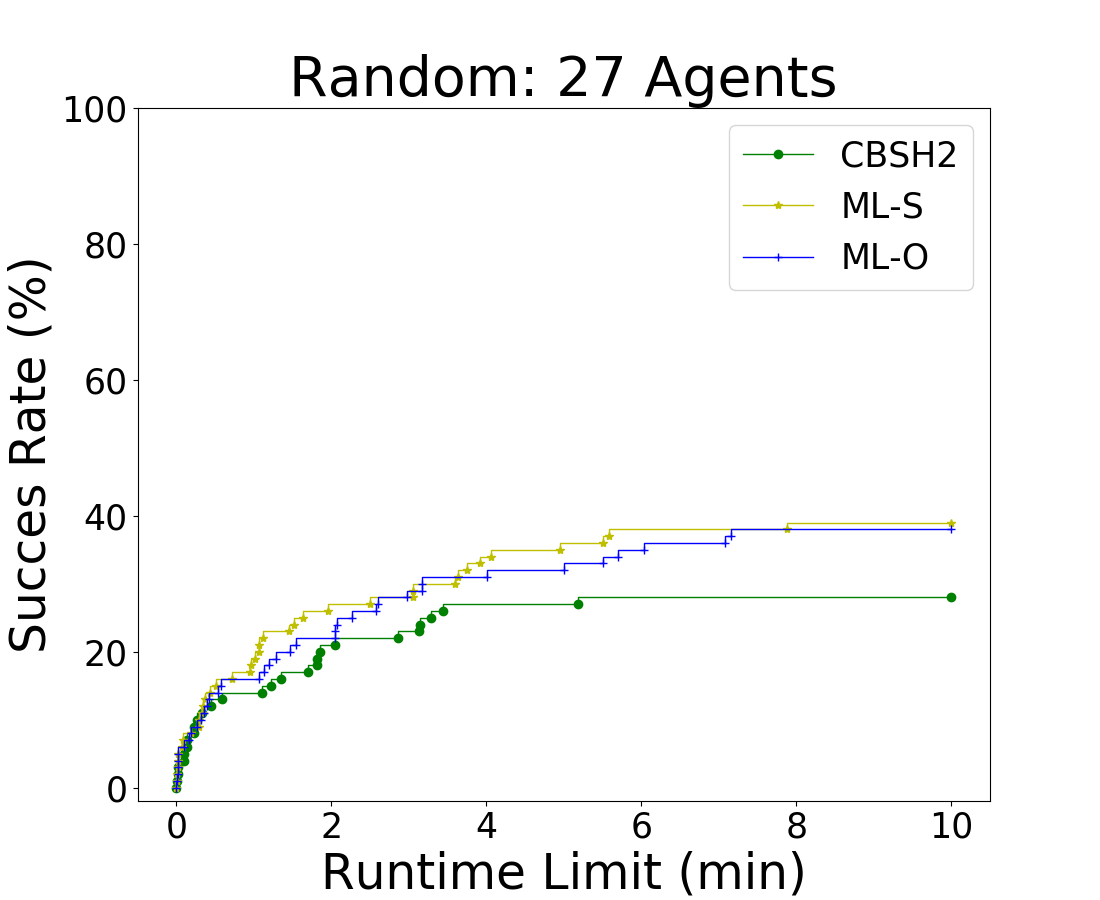}
		\includegraphics[width=5cm]{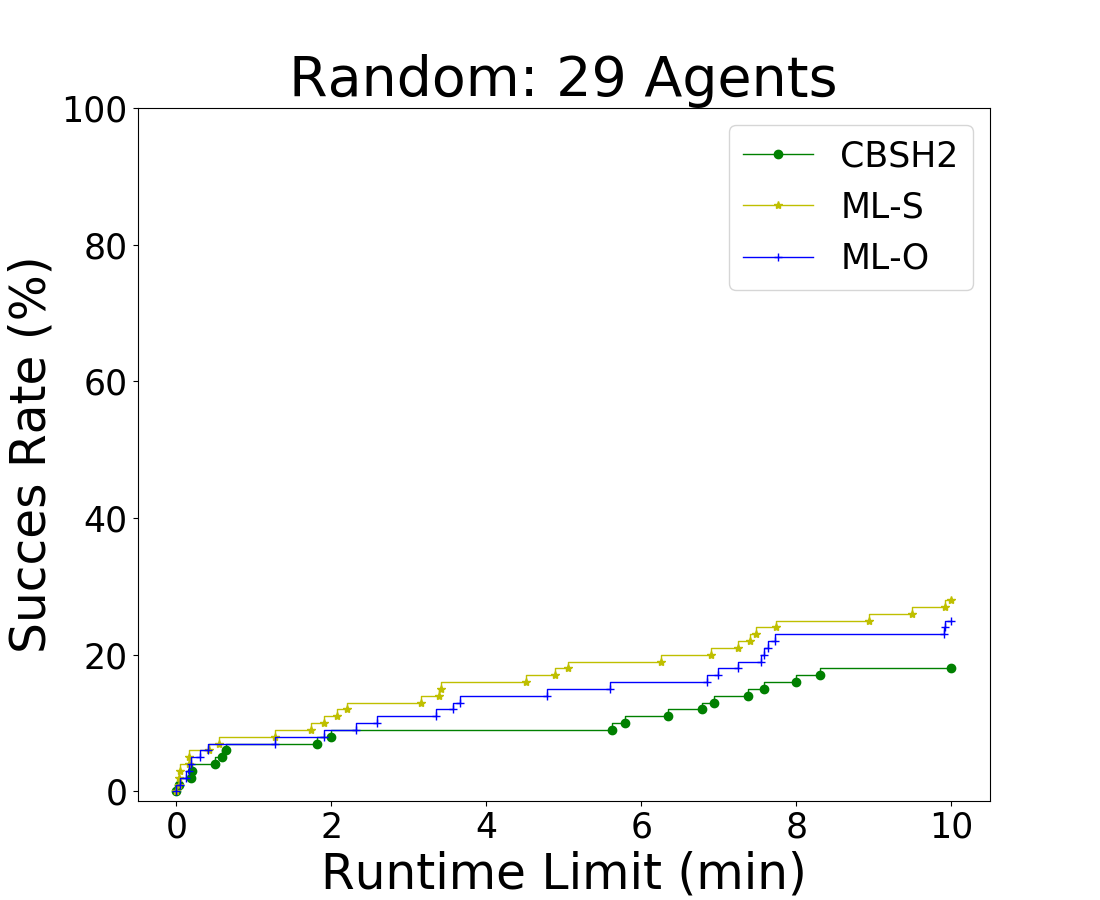}
	\end{subfigure}
	
	\caption{The random map: Percentage of solved instances under problem parameters.\label{cutoffRandom}}
\end{figure*}

\begin{figure*}[htbp]
	\centering
		\begin{subfigure}[htbp]{0.99\textwidth}
		\centering
		\includegraphics[width=5cm]{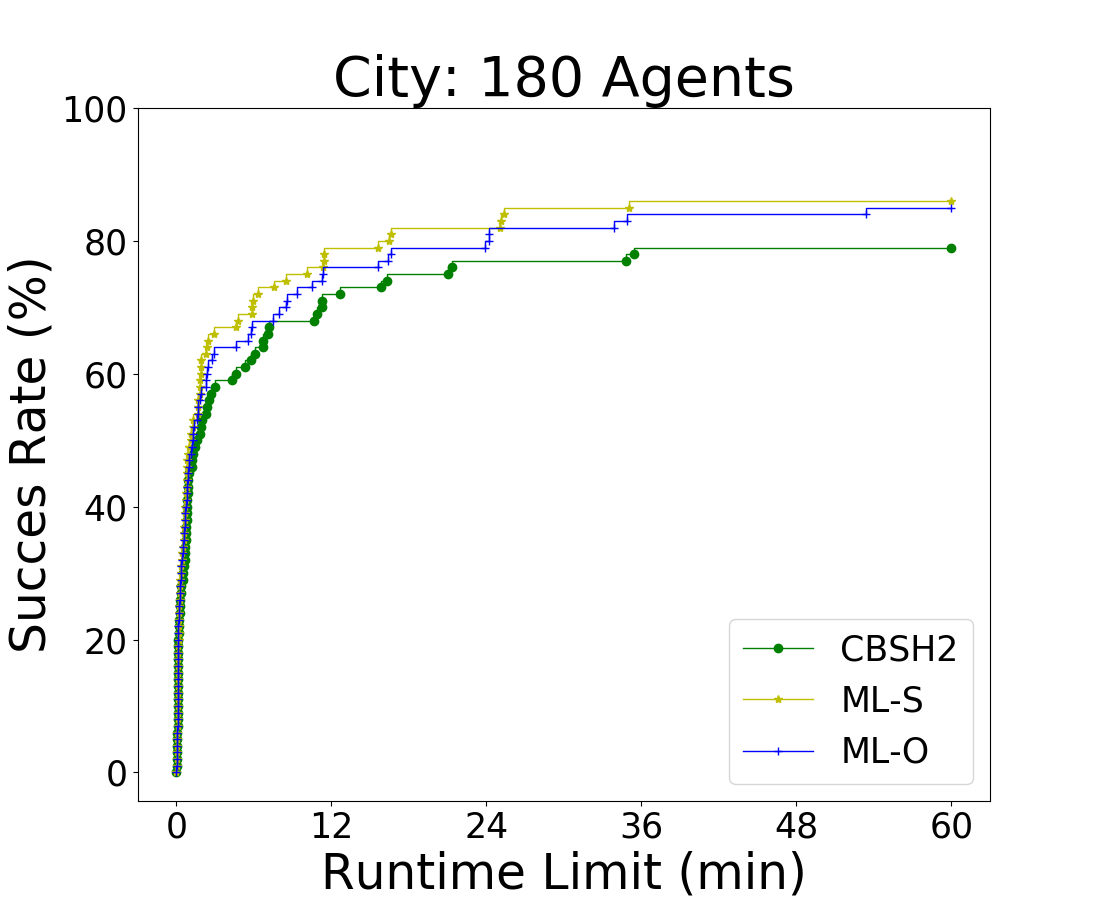}
		\includegraphics[width=5cm]{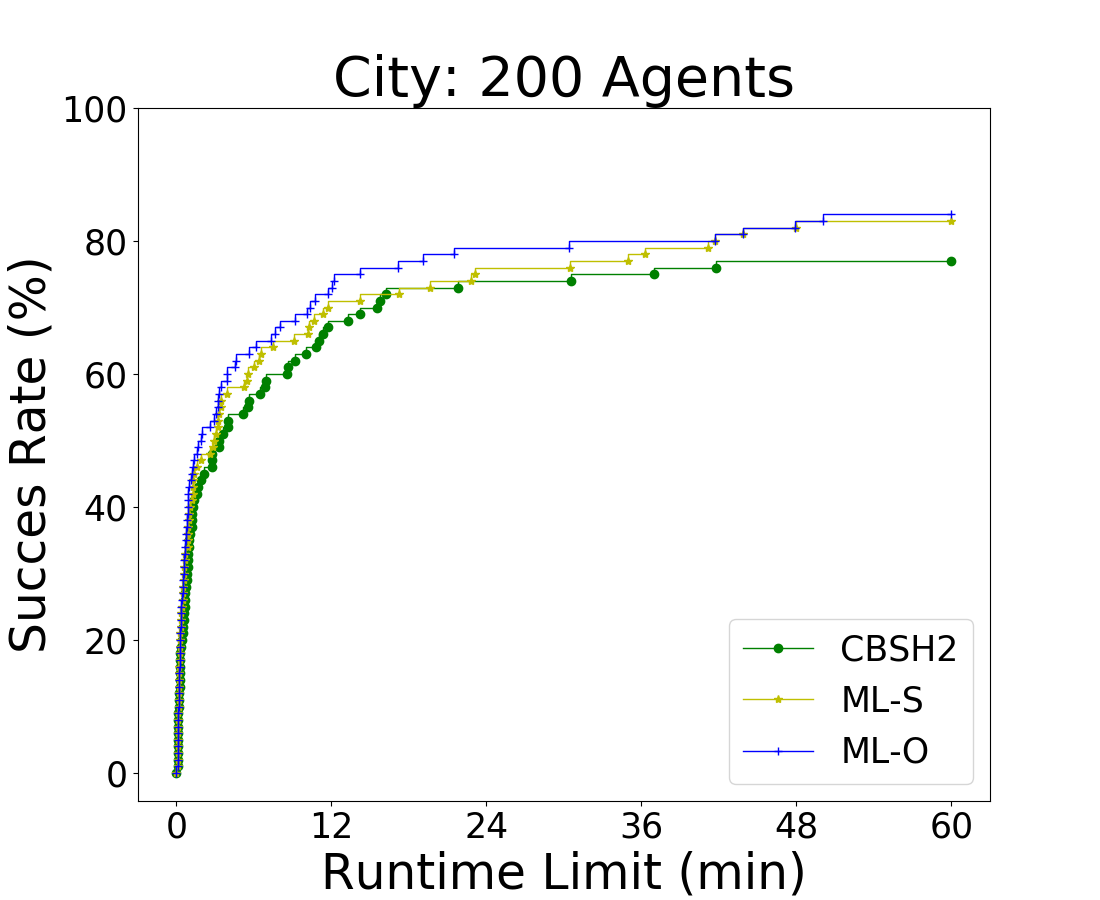}
		\includegraphics[width=5cm]{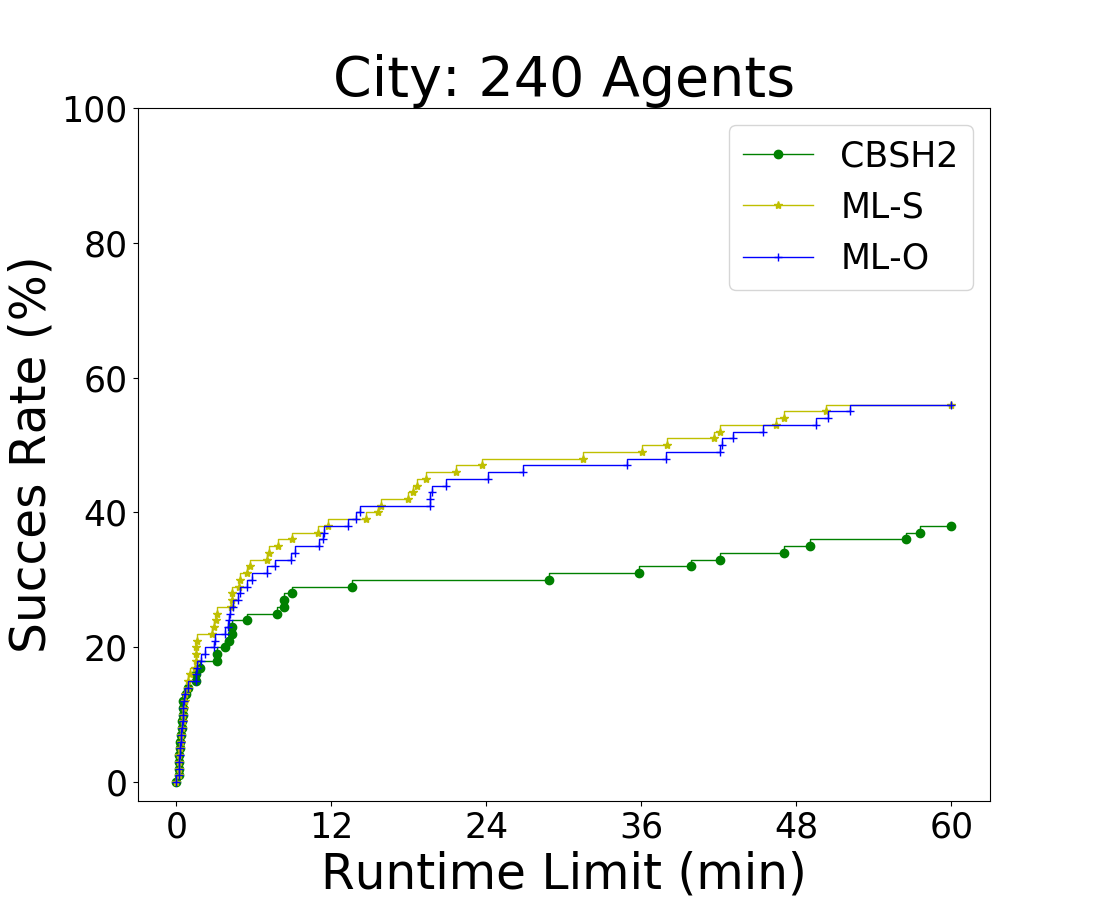}
	\end{subfigure}
	
		\begin{subfigure}[htbp]{0.99\textwidth}
		\centering
		\includegraphics[width=5cm]{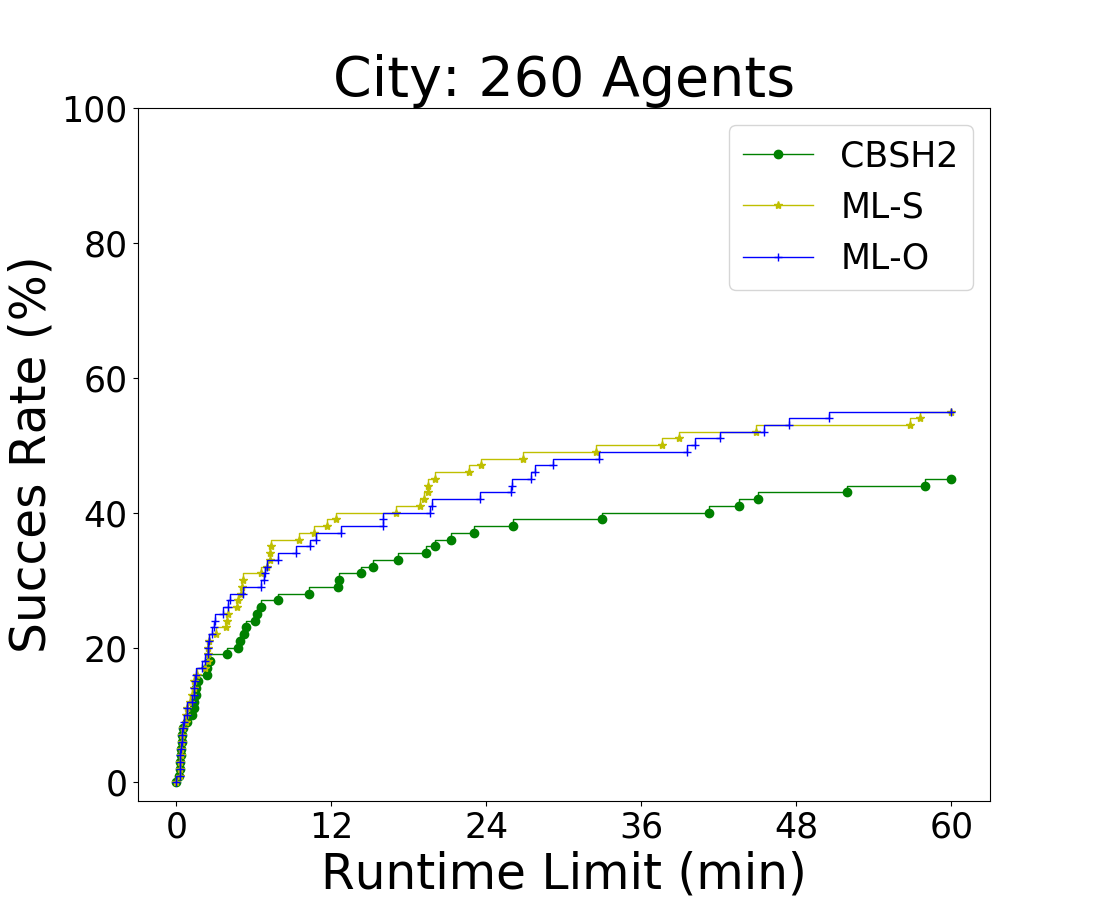}
		\includegraphics[width=5cm]{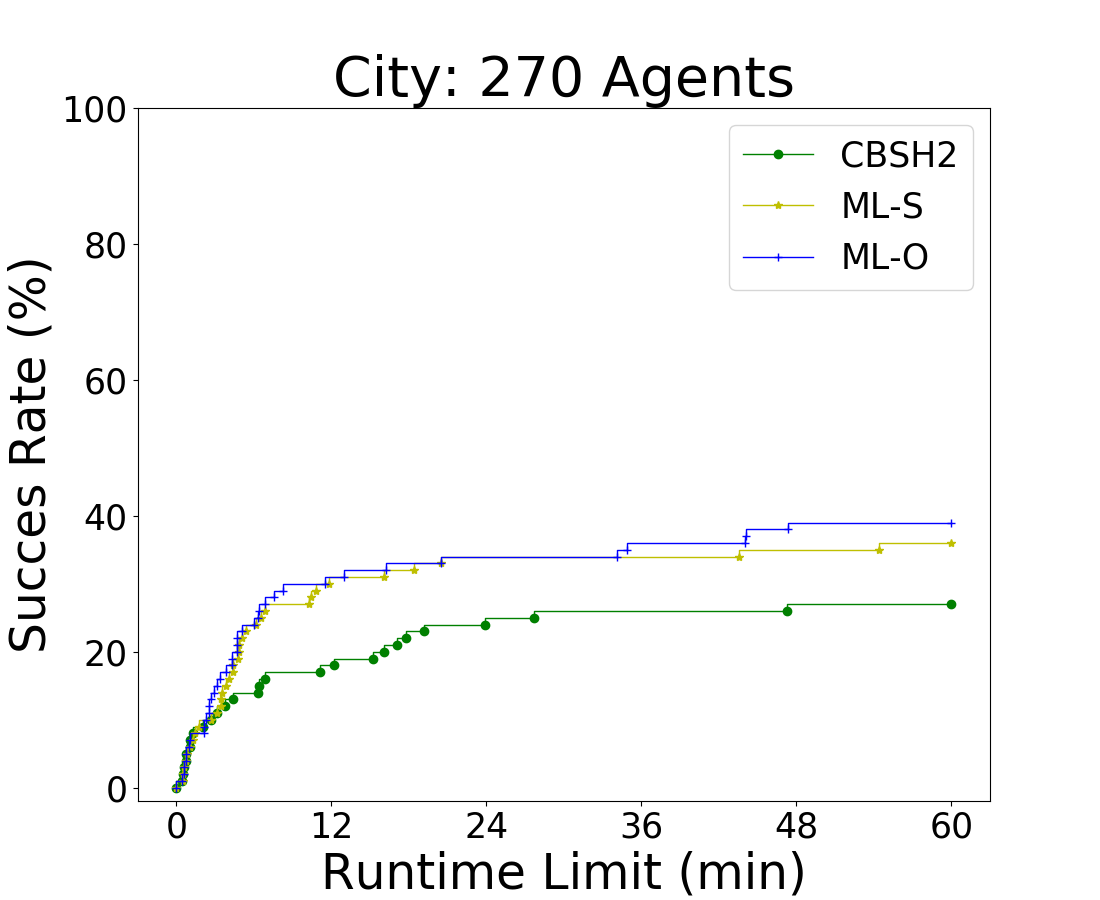}
		\includegraphics[width=5cm]{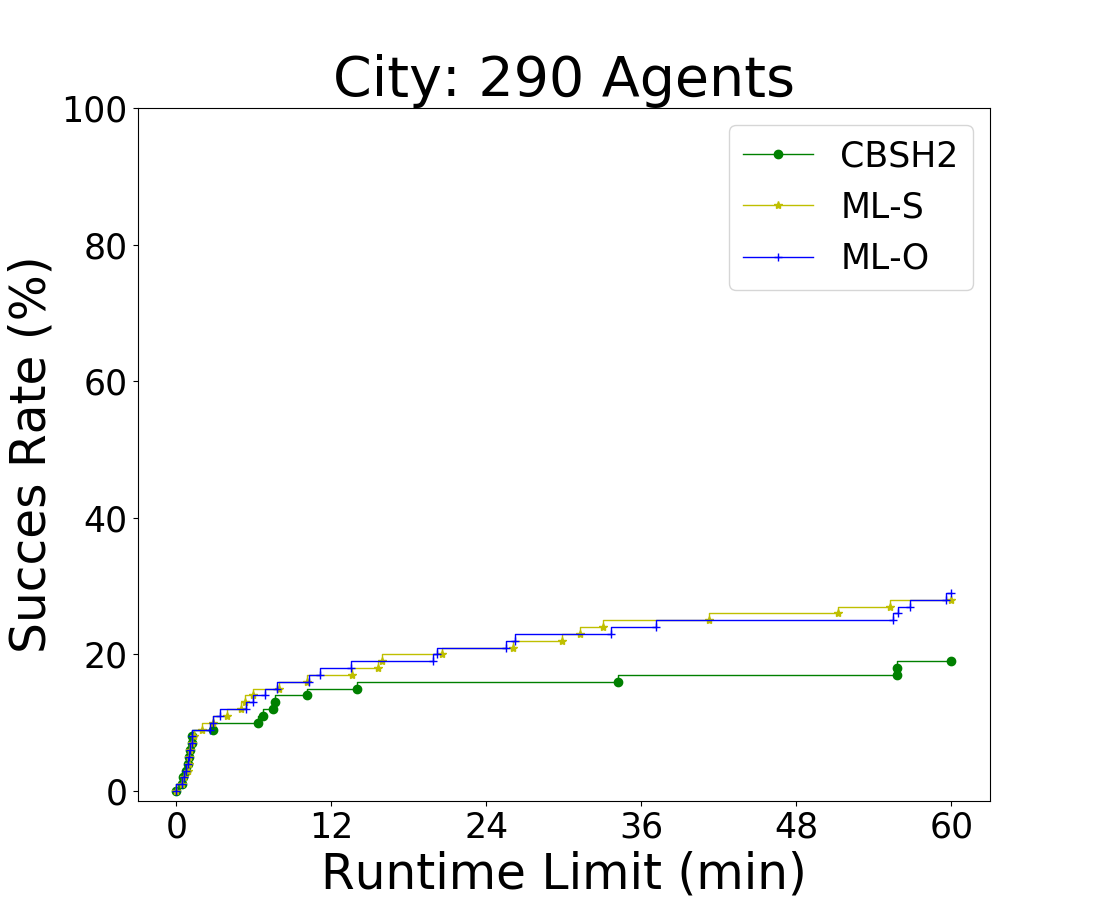}
	\end{subfigure}
	
	\caption{The city map: Percentage of solved instances under problem parameters.\label{cutoffCity}}
\end{figure*}

\begin{figure*}[htbp]
	\centering
		\begin{subfigure}[htbp]{0.99\textwidth}
		\centering
		\includegraphics[width=5cm]{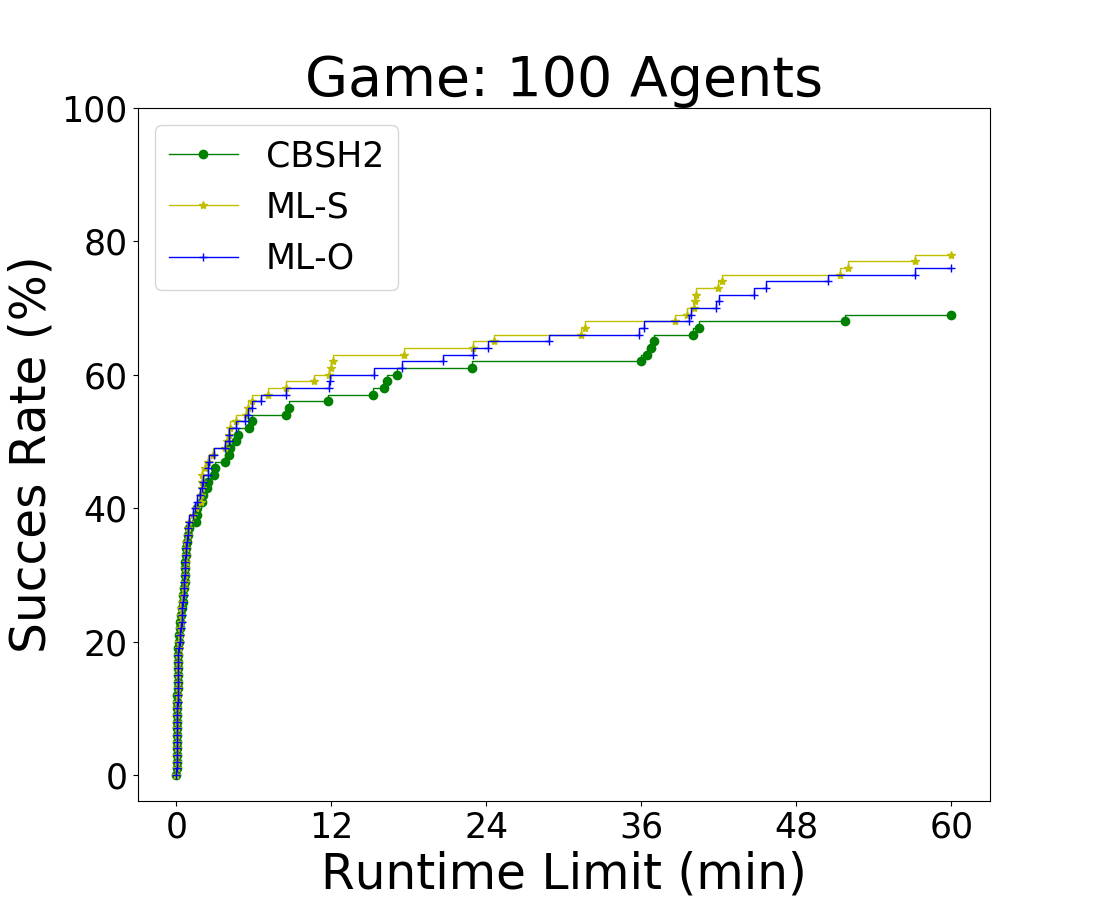}
		\includegraphics[width=5cm]{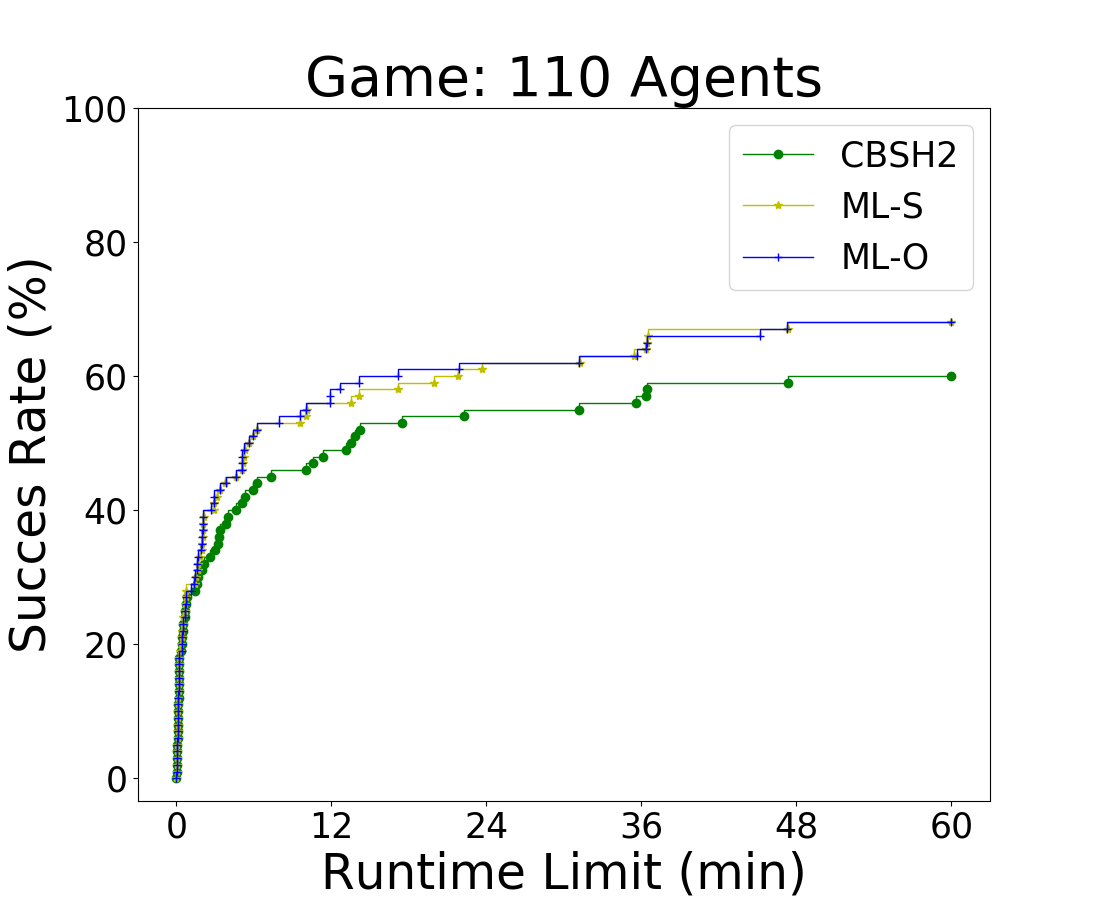}
		\includegraphics[width=5cm]{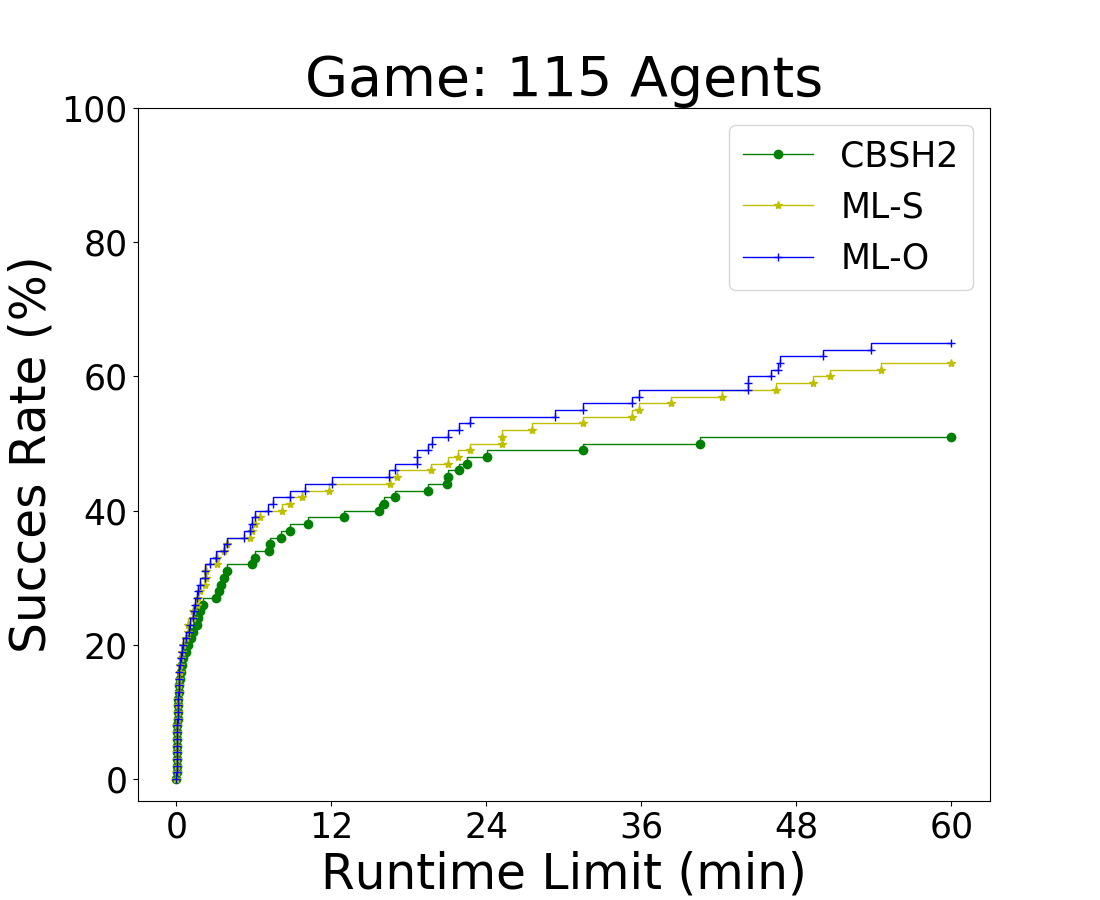}
	\end{subfigure}
	
		\begin{subfigure}[htbp]{0.99\textwidth}
		\centering
		\includegraphics[width=5cm]{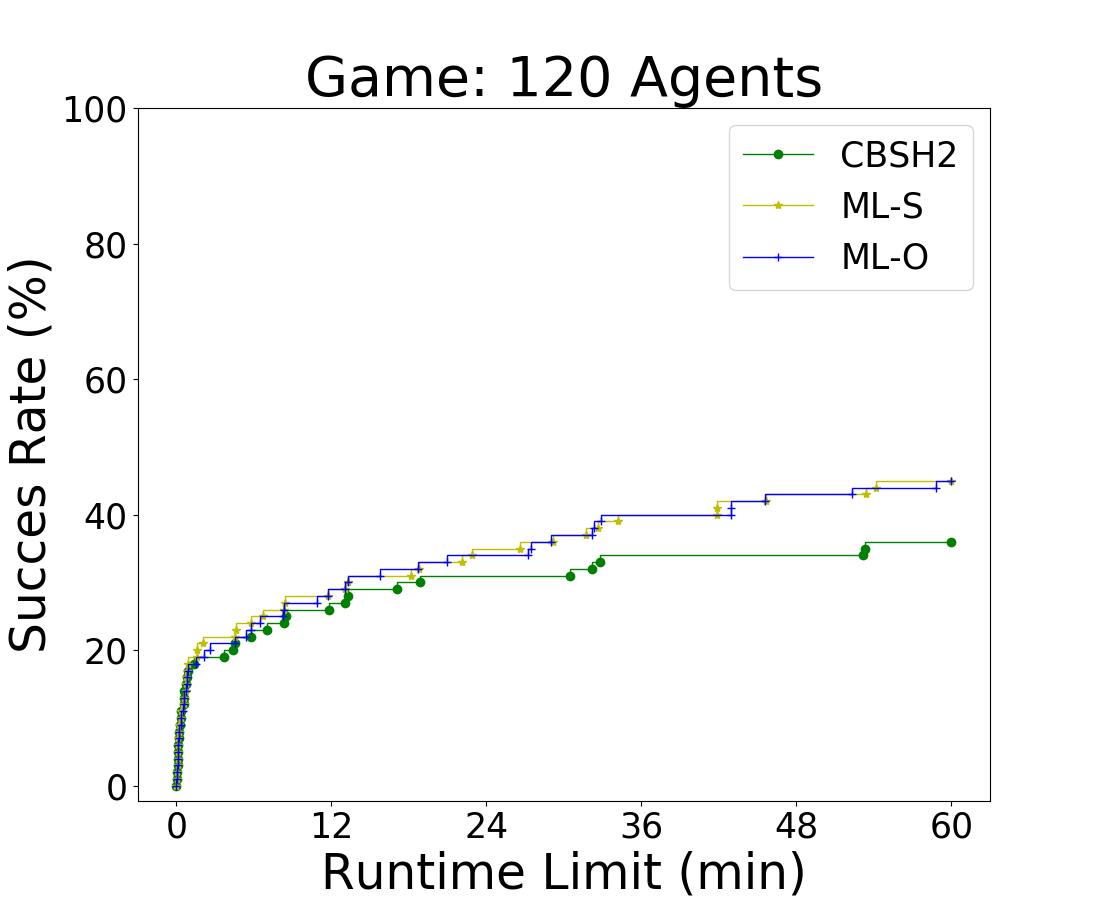}
		\includegraphics[width=5cm]{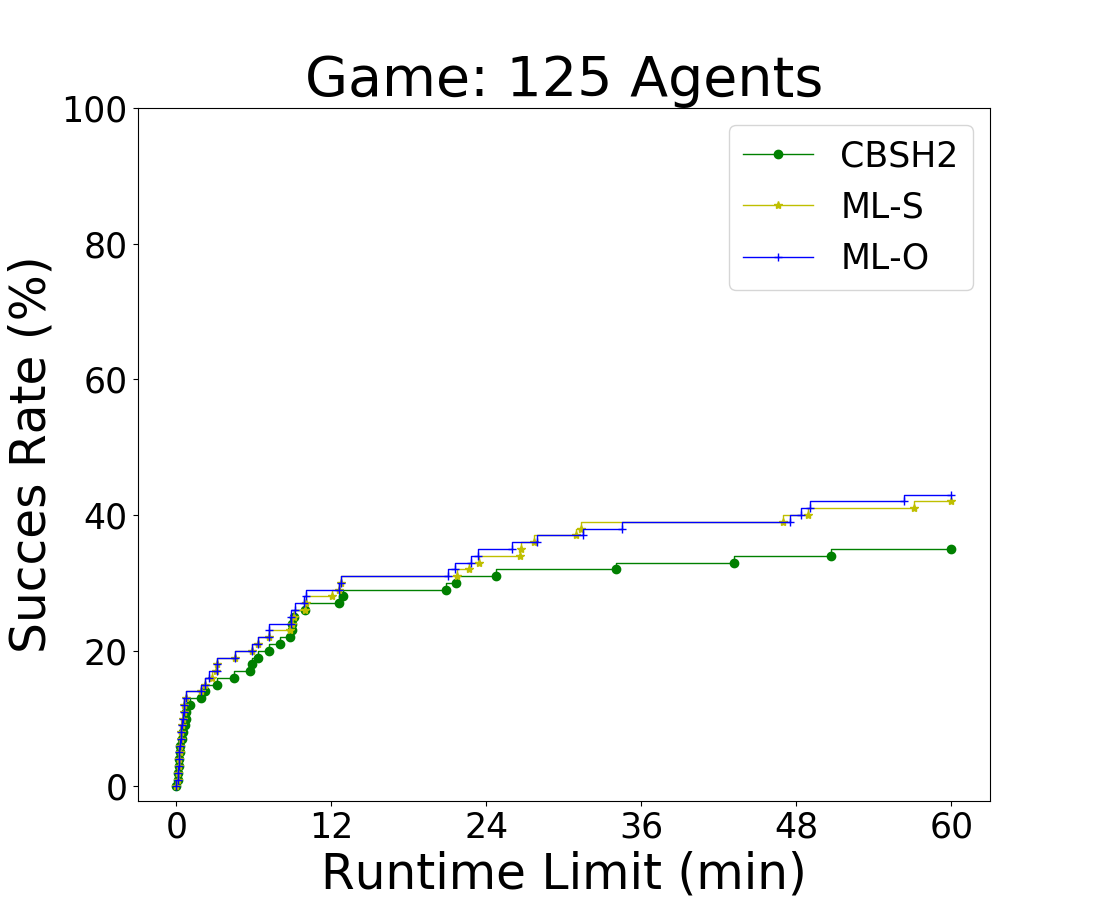}
		\includegraphics[width=5cm]{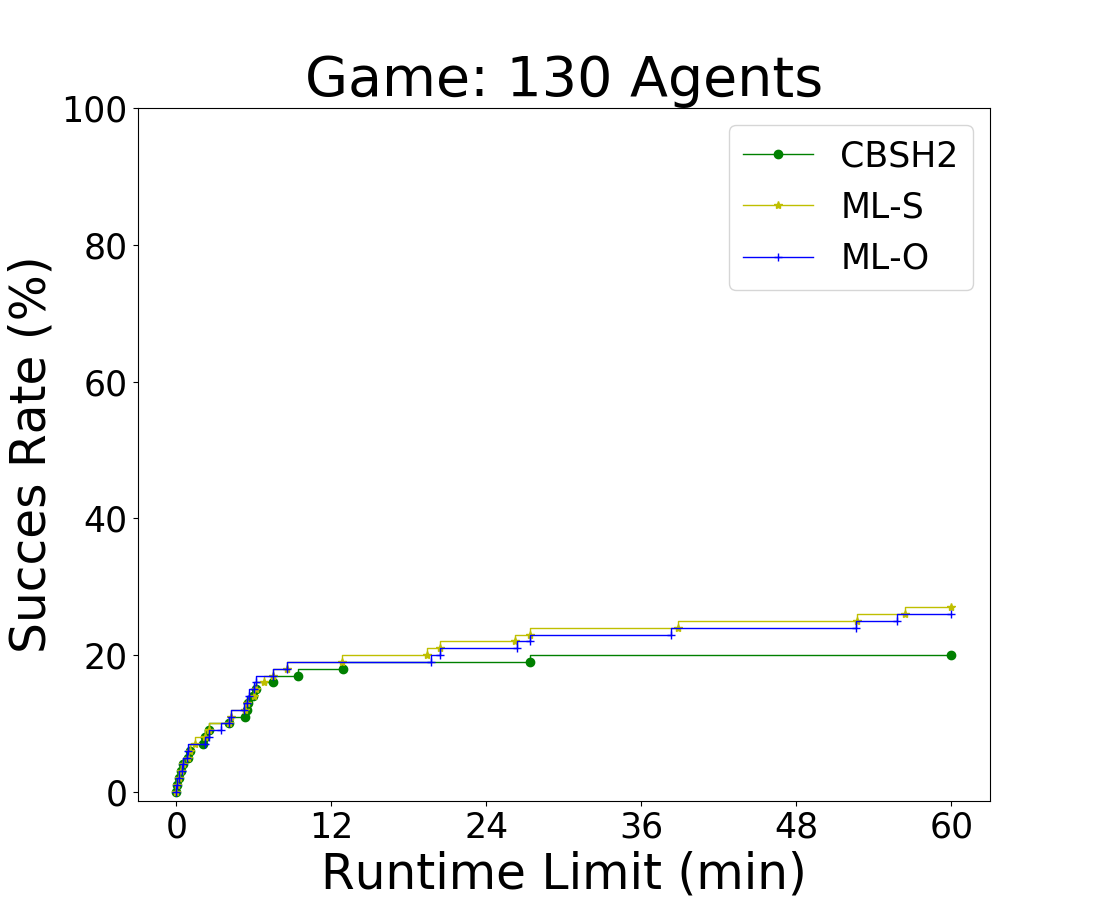}
	\end{subfigure}
	
	\caption{The game map: Percentage of solved instances under problem parameters.\label{cutoffGame}}
\end{figure*}

The success rates on the room, maze, random and game maps are shown in Figure \ref{succratesApp}, where we can see that the success rates of \MLS and \MLO are both marginally higher than CBSH2.  Table \ref{thefulltable} includes the results on all data points in Figures \ref{succrates} and \ref{succratesApp}.
The success rates for different runtime limits varying the number of agents on the warehouse, room, maze, random, city and game maps are shown in Figures \ref{cutoffWarehouse}, \ref{cutoffRoom}, \ref{cutoffMaze}, \ref{cutoffRandom}, \ref{cutoffCity} and \ref{cutoffGame}, respectively.

\section{B\,\, Feature Importance}
For \MLS, we plot the weights of the features in decreasing order of their absolute values for each of the six ranking functions in Figures \ref{FI4} and \ref{FI2}. For \MLO, we take the average weight of each feature over the six ranking functions and sort them in decreasing order of their absolute values. The plot is shown in  Figure \ref{FIO}.

The top five features for the warehouse map are:
(1) the minimum of the numbers of conflicts involving agent $a_i$ ($a_j$) that have been selected and resolved; 
(2) the maximum of the differences of the cost of the path of agent $a_i$ ($a_j$) and time step $t$;
(3) the sum of the numbers of conflicts involving agent $a_i$ ($a_j$) that have been selected and resolved; 
(4) the minimum of the differences of the cost of the path of agent $a_i$ ($a_j$) and time step $t$;
(5) the binary indicator for non-cardinal conflicts.

The top five features for the room map are:
(1) the binary indicator for non-cardinal conflicts; 
(2) the weight of the edge between agents $a_i$ and $a_j$ in the weighted dependency graph; 
(3) the binary indicator for cardinal conflicts; 
(4) number of empty cells that are two steps away from where the conflicts occur
(5) the minimum of the numbers of conflicts involving agent $a_i$ ($a_j$) that have been selected and resolved.

The top five features for the maze map are: 
(1) the maximum of the widths of level $t-2$ of the MDD for agent $a_i$ ($a_j$);
(2) the weight of the edge between agents $a_i$ and $a_j$ in the weighted dependency graph; 
(3) the maximum of the differences of the cost of the path of agent $a_i$ ($a_j$) and time step $t$;
(4) the binary indicator for semi-cardinal conflicts; 
(5) the maximum of the widths of level $t$ of the MDD for agent $a_i$ ($a_j$).

The top five features for the random map are: 
(1) the binary indicator for non-cardinal conflicts; 
(2) the  weight of the edge between agents $a_i$ and $a_j$ in the weighted dependency graph; 
(3) the binary indicator for cardinal conflicts; 
(4) the maximum of the widths of level $t$ of the MDD for agent $a_i$ ($a_j$);
(5) the maximum of the differences of the cost of the path of agent $a_i$ ($a_j$) and time step $t$.

The top five features for the city map are:
(1) the binary indicator for non-cardinal conflicts; 
(2) the maximum of the numbers of conflicts involving agent $a_i$ ($a_j$) that have been selected and resolved; 
(3) the minimum of the numbers of conflicts involving agent $a_i$ ($a_j$) that have been selected and resolved; 
(4) the binary indicator for cardinal conflicts; 
(5) the sum of the numbers of conflicts involving agent $a_i$ ($a_j$) that have been selected and resolved.

The top five features for the game map are:
(1) the minimum of the differences of the cost of the path of agent $a_i$ ($a_j$) and time step $t$;
(2) the maximum of the differences of the cost of the path of agent $a_i$ ($a_j$) and time step $t$;
(3) the maximum of the numbers of conflicts involving agent $a_i$ ($a_j$) that have been selected and resolved; 
(4) the minimum of the numbers of conflicts involving agent $a_i$ ($a_j$) that have been selected and resolved; 
(5) the sum of the numbers of conflicts involving agent $a_i$ ($a_j$) that have been selected and resolved.

\begin{figure}[tbp]
	\centering
	
		\centering
		\includegraphics[width=8cm]{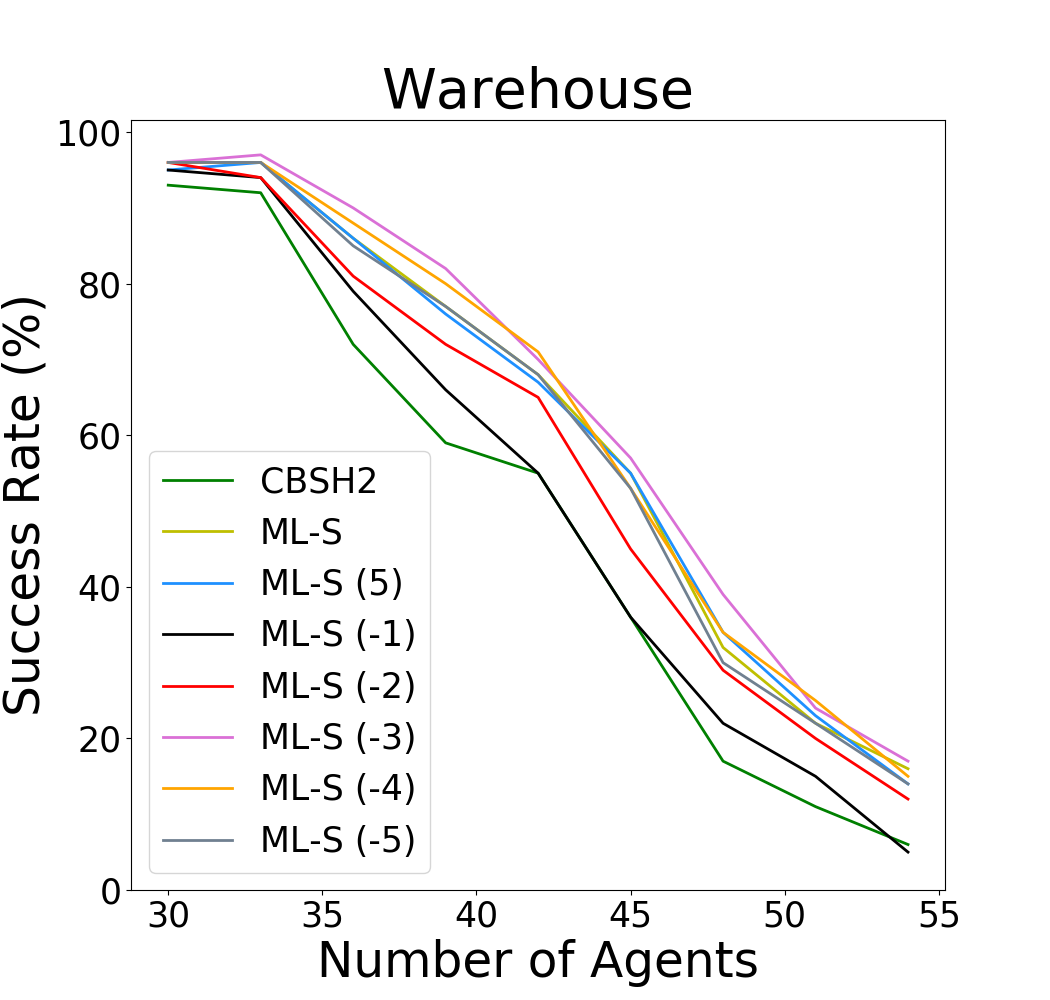}

	\caption{Success rates with feature selection on the warehouse map. The curver for \MLS is partially hidden by the curves for \MLS(5) and \MLS(-5). \label{FeatureSelection}}
\end{figure}

\subsection{Results on Feature Selection}
\begin{figure*}[htbp]
	\centering
	%\begin{subfigure}[htbp]{0.24\textwidth}
		%\centering
		%\includegraphics[height=3.6cm]{figures/SuccesRateSmall.png}
		%\caption{Success rates on the small map.\label{smallmapres}}
	%\end{subfigure}
		\begin{subfigure}[htbp]{0.99\textwidth}
		\centering
		\includegraphics[width=20cm]{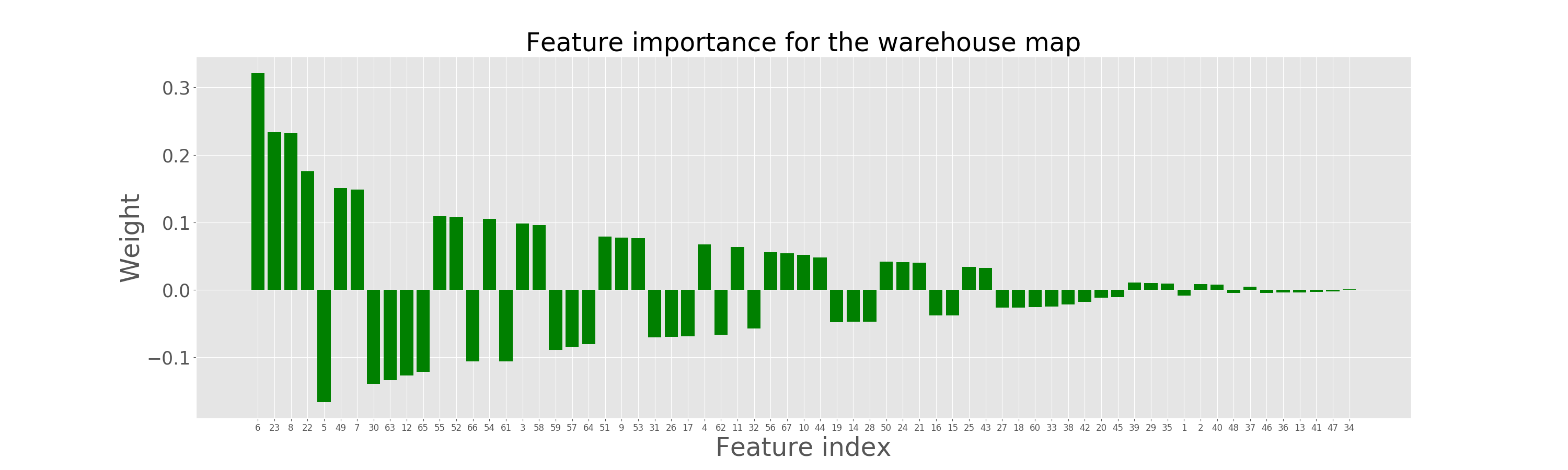}
		\includegraphics[width=20cm]{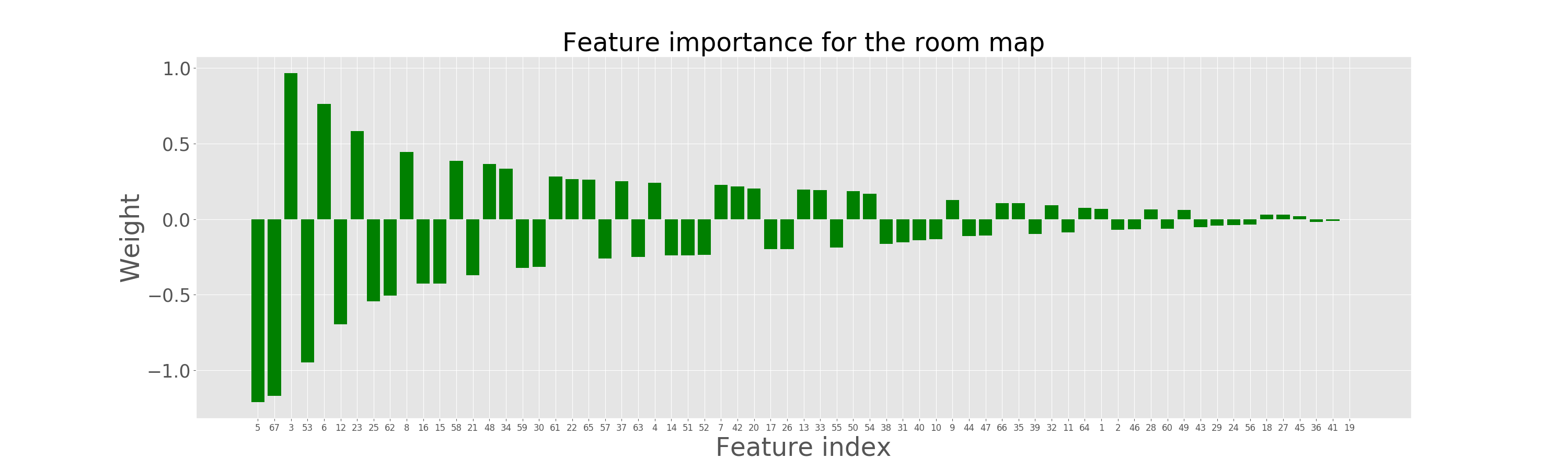}
		\includegraphics[width=20cm]{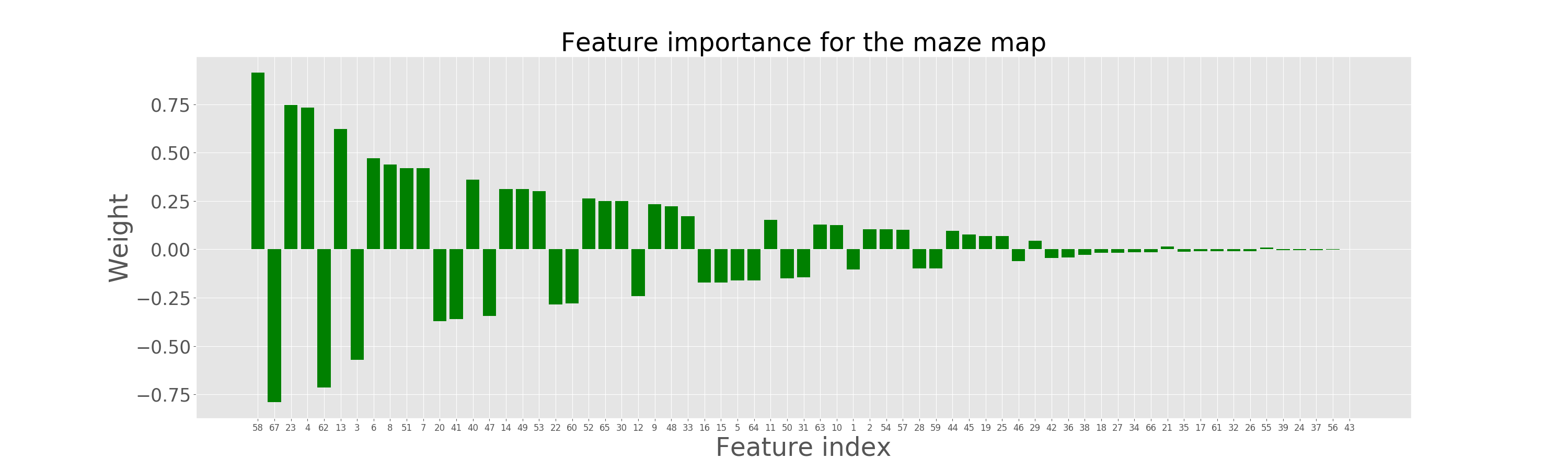}
		\includegraphics[width=20cm]{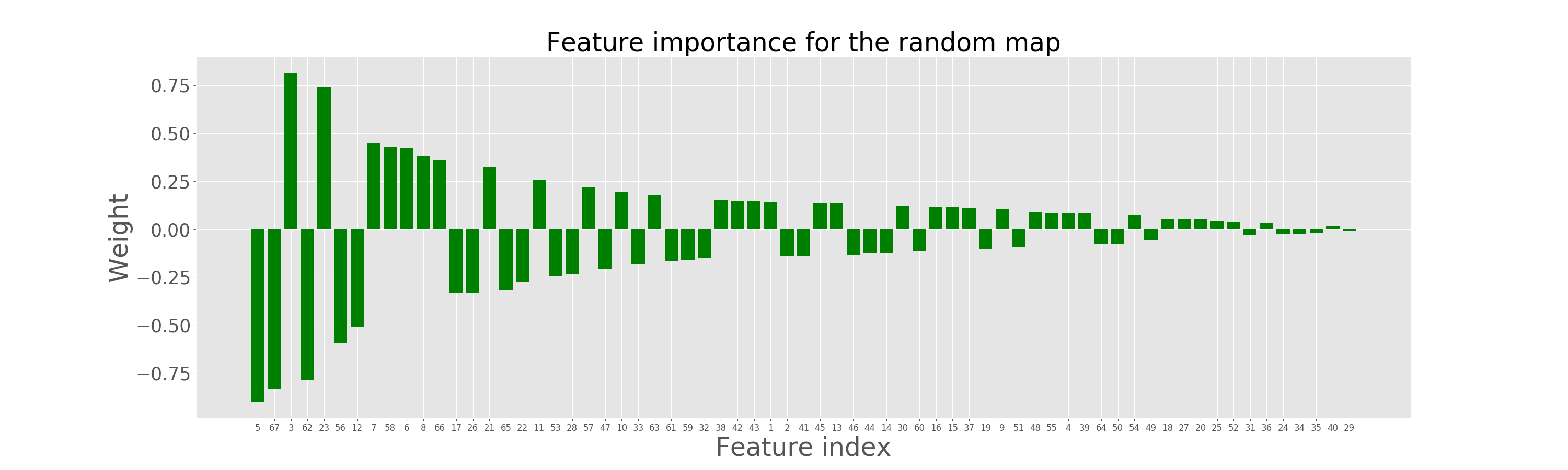}

	\end{subfigure}

	\caption{Feature importance plots for the warehouse, room, maze and random maps.\label{FI4}}
\end{figure*}

\begin{figure*}[htbp]
	\centering
	%\begin{subfigure}[htbp]{0.24\textwidth}
		%\centering
		%\includegraphics[height=3.6cm]{figures/SuccesRateSmall.png}
		%\caption{Success rates on the small map.\label{smallmapres}}
	%\end{subfigure}
		\begin{subfigure}[htbp]{0.99\textwidth}
		\centering
		\includegraphics[width=20cm]{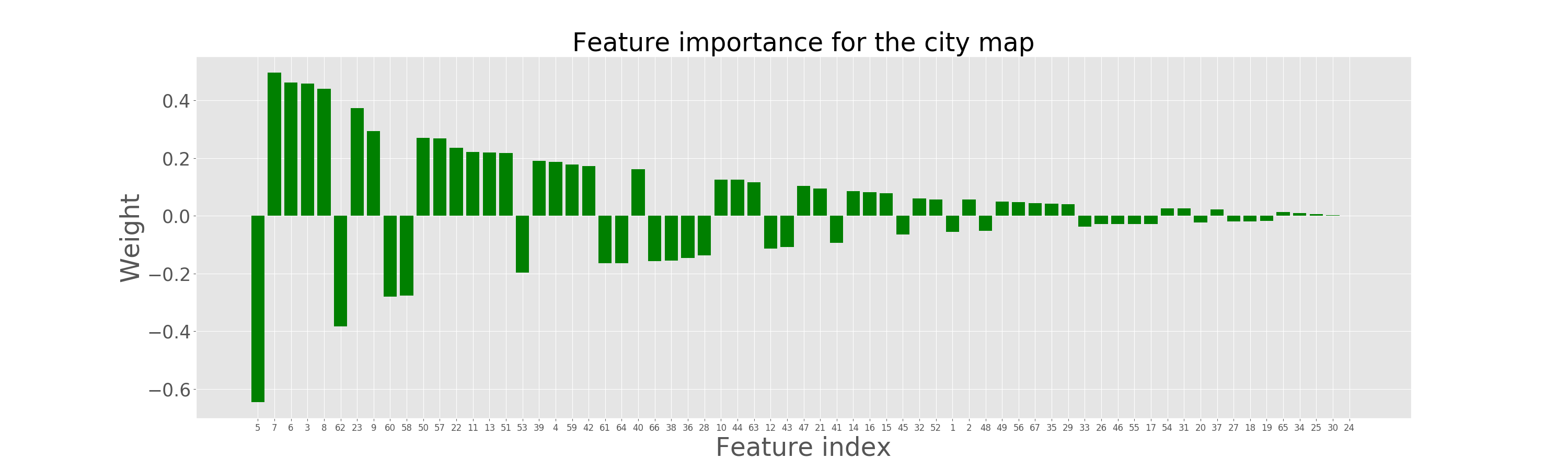}
		\includegraphics[width=20cm]{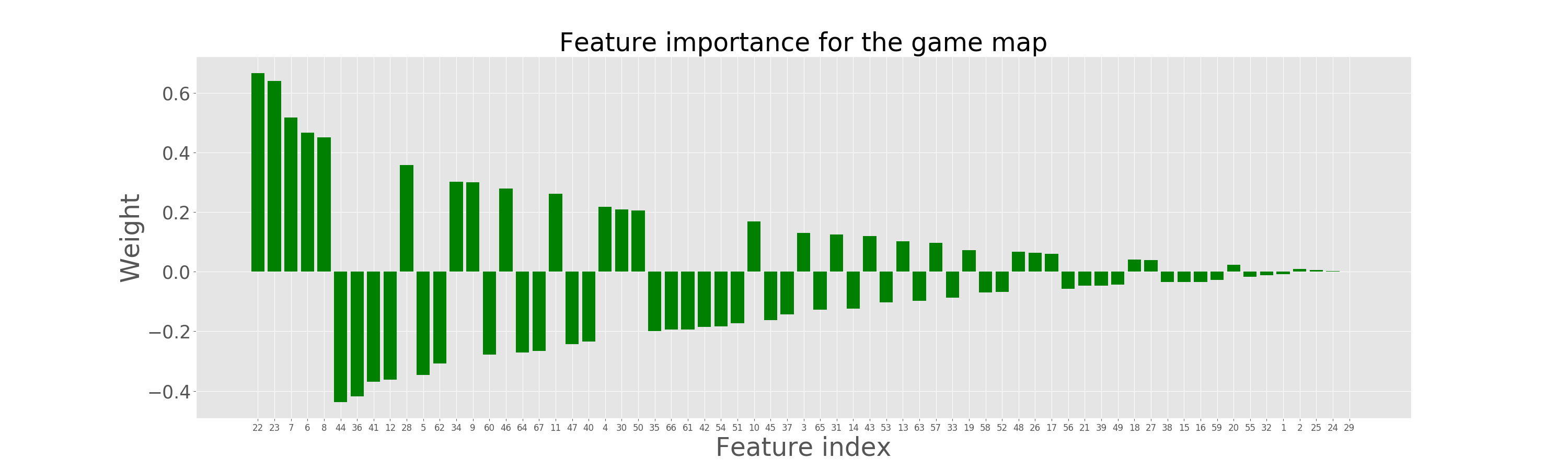}

	\end{subfigure}

	\caption{Feature importance plots for the city and game maps.\label{FI2}}
\end{figure*}

\begin{figure*}[htbp]
	\centering
	%\begin{subfigure}[htbp]{0.24\textwidth}
		%\centering
		%\includegraphics[height=3.6cm]{figures/SuccesRateSmall.png}
		%\caption{Success rates on the small map.\label{smallmapres}}
	%\end{subfigure}
		\centering
		\includegraphics[width=20cm]{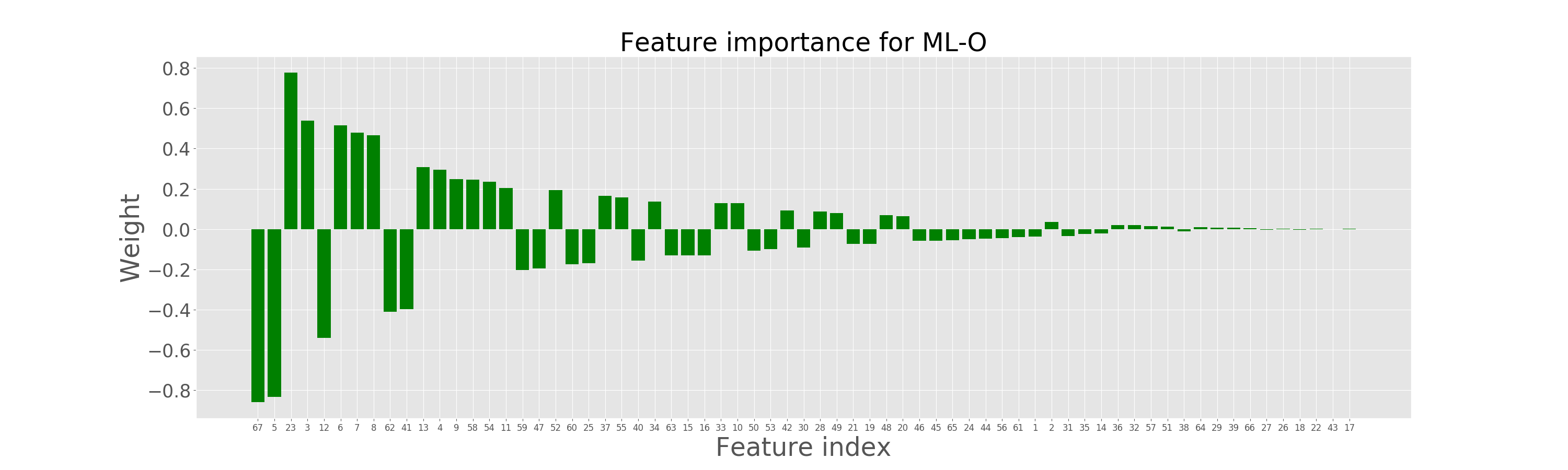}

	\caption{Feature importance plots for \MLO.\label{FIO}}
\end{figure*}

In this subsection, we present preliminary results on feature selection. We select five categories of features: (1) features related to the cardinal conflicts (features 3,4 and 5); (2) features related to the frequency of a conflict being resolved for an agent (features 6,7 and 8); (3) features related to the numbers of conflicts that the agent is involved in (features 12 and 13); (4) features related to the difference of the cost of the path of the agent and its individually cost-minimal path; (5) features related to the MDDs and the WDG (features 62 and 67).
The features selected cover the top ten features for \MLO (as shown in Figure \ref{FIO}) that can be computed in constant time and four features among the top five features for each individual map (as shown in Figures \ref{FI2} and \ref{FI4}).
We train a ranking function with all five categories of features and denote the corresponding solver as \MLS(5). We then hold out each of the five categories and train a ranking functions with the rest of four categories. We denote the solver that is trained without the $i$-th category as \MLS($-i$). Since now we use only 9 to 12 features for training, we can afford to train a SVM$^{rank}$ using a polynomial kernel of degree 2 for each solver while keeping the other parameters the same. We show the success rates on the warehouse map in Figure \ref{FeatureSelection}. \MLS(5) performs similarly to \MLS, implying that when using only the selected features, we are still able to achieve performance as good as \MLS which uses all features. \MLS(-1) trained without using the 1-st category (features related to the cardinal conflicts) performs the worse among our solvers, only slightly better than CBSH2. \MLS (-3) is the best among our solvers, which dominates \MLS.

\section{C\,\,Code and Data for Reproducibility}
We provide the core of our code in the supplementary material, which is based on the open-source code from \cite{li2019improved}. We also include the random and warehouse maps used in experiments, while other maps could be found at \cite{stern2019multi} and \cite{sturtevant2012benchmarks}. We do not include the training and test data due to the file size limit.


\begin{table*}[t]
\scriptsize
\centering
\begin{tabular}{|c|c|rrr|rrr|rrr|rrr|}
\hline
\multirow{2}{*}{Map} & \multirow{2}{*}{$k$} & \multicolumn{3}{c|}{Success Rate (\%)} & \multicolumn{3}{c|}{Runtime (min)} & \multicolumn{3}{c|}{CT Size (nodes)} & \multicolumn{3}{c|}{PAR10 Score} \\ \cline{3-14} 
 &  & CBSH2 & ML-S &ML-O & CBSH2 &  ML-S &ML-O & CBSH2 &  ML-S &ML-O & CBSH2 &  ML-S &ML-O\\ \hline
\multirow{9}{*}{
\begin{minipage}{.07\textwidth}
\begin{center}
Warehouse
\\
\includegraphics[width=1.3cm]{figures/warehouseLarge.png}\end{center}
\end{minipage}} &30 & 93 & {\bf96 (93)} & {\bf96 (93)} & 0.20 & {\bf0.06} & 0.07 & 1,154 & {\bf294} & 378 &  7.18 & {\bf4.14} & 4.25\\ \cline{2-14} 
 & 33 & 92 & {\bf96} (91) & 95 (91) & 0.37 & 0.21 & {\bf0.14} & 2,367 & 1,053 & {\bf721} &  8.37 & {\bf4.38} & 5.27\\ \cline{2-14} 
 & 36 & 72 & 86 (71) & {\bf88} (71) & 0.54 & 0.24 & {\bf0.19} & 3110 & 980 & {\bf977} &  28.46 & 14.56 & {\bf12.81} \\ \cline{2-14} 
 & 39 & 59 & {\bf77} (57) & 75 {\bf(59)} & 0.77 & {\bf0.39} & 0.49 & 4,064 & {\bf1,395} & 2,000 &  41.52 & {\bf23.83} & 25.84 \\ \cline{2-14} 
 & 42 & 55 & 68 {\bf(55)} & {\bf70 (55)} & 1.27 & 0.65 & {\bf0.38} & 6,834 & 2,874 & {\bf1,781} &  45.70 & 32.61 & {\bf30.56} \\ \cline{2-14} 
 & 45 & 36 & {\bf55 (36)} & 53 {\bf(36)} & 2.10 & 0.99 & {\bf0.68} & 9,887 & 4,980 & {\bf3,333} &  64.76 & {\bf45.94} & 47.79 \\ \cline{2-14} 
 & 48 & 17 & {\bf32 (17)} & {\bf32 (17)} & 1.99 & 1.12 & {\bf0.56} & 9,646 & 5,357 & {\bf2,221} &  83.34 & 68.64 & {\bf68.48} \\ \cline{2-14} 
  & 51 & 11 & 22 {\bf(11)} & {\bf23 (11)} & 1.71 & 0.93 & {\bf0.16} & 7,685 & 3,768 & {\bf429} &  89.19 & 78.31 & {\bf77.26} \\ \cline{2-14} 
 & 54  & 6 & {\bf16 (6)} & 15 {\bf(6)} & 2.82 & 1.70 & {\bf1.23} & 12,816 & 8,886 & {\bf6,427} &  94.17 & {\bf84.42} & 85.36\\ \hline  \hline
\multirow{13}{*}{
\begin{minipage}{.07\textwidth}
\begin{center}
City
\\
\includegraphics[width=1.3cm]{figures/Paris_1_256.png}\end{center}
\end{minipage}}
& 170 & 86 & 92 (84) & {\bf94 (86)} & 3.85 & 2.87 & {\bf2.80} & 669 & {\bf355} & 361 &  87.77 & 50.52 & {\bf38.92} \\ \cline{2-14} 
 & 180 & 78 & {\bf85} (76) & 84 (75) & 3.53 & {\bf2.43} & 2.46 & 859 & {\bf468} & 476 &  134.99 & {\bf93.04} & 99.87 \\ \cline{2-14} 
 & 190 & 70 & 76 (69) & {\bf77} (68) & 4.42 & {\bf2.96} & 3.07 & 878 & 326 & {\bf325} &  192.05 & 155.37 & {\bf150.13} \\ \cline{2-14} 
 & 200 & 76 & 82 (75) & {\bf83} (75) & 4.78 & 5.08 & {\bf4.13} & 849 & 702 & {\bf490} &  147.96 & 113.53 & {\bf106.78}\\ \cline{2-14} 
 & 210 & 67 & 75 (65) & {\bf76} (66) & 5.40 & {\bf3.21} & 3.27 & 956 & 339 & {\bf335}&  201.66 & 153.96 & {\bf147.72} \\ \cline{2-14} 
  & 220 & 46 & {\bf60} (45) & 59 (45) & 8.24 & 6.91 & {\bf5.74} & 1,546 & 648 & {\bf638} &  328.21 & {\bf246.88} & 251.54 \\ \cline{2-14} 
 & 230 & 57 & {\bf68} (56) & 64 (54) & 4.86 & 4.26 & {\bf4.24} & 835 & {\bf444} & 449  &  261.36 & {\bf196.99} & 220.50 \\ \cline{2-14} 
 & 240 & 37 & {\bf55} (35) & {\bf55} (34) & 11.11 & {\bf4.73} & 6.87 & 2,368 & {\bf541} & 968 &  382.45 & {\bf276.16} & 277.02 \\ \cline{2-14} 
 & 250& 34 & {\bf51 (34)} & 50 {\bf(34)} & 6.34 & {\bf5.44} & 6.25 & 1,288 & {\bf1,057} & 1,170 &  398.18 & {\bf299.93} & 305.62 \\ \cline{2-14}
  & 260 & 44 & {\bf54 (44)} & {\bf54} (43) & 11.69 & 9.75 & {\bf9.55} & 1,883 & {\bf1,178} & 1,219 & 341.37 & {\bf282.00} & 282.58  \\ \cline{2-14} 
 & 270 & 26 & 35 (24) & {\bf38} (24) & 8.38 & {\bf5.81} & 6.48 & 1,293 & {\bf682} & 822 &  446.51 & 392.84 & {\bf375.70}  \\ \cline{2-14} 
 & 280 & 23 & 29 (19) & {\bf33} (22) & 7.42 & 7.09 & {\bf4.97} & 1,099 & 811 & {\bf650} &  464.48 & 429.63 & {\bf406.60}  \\ \cline{2-14} 
 & 290  & 18 & 27 (16) & {\bf28} (17) & 11.65 & {\bf8.45} & 8.75 & 1,966 & {\bf1,372} & 1,429 &  494.10 & 441.87 & {\bf436.70}
  \\ \hline\hline
 
\multirow{8}{*}{
\begin{minipage}{.07\textwidth}
\begin{center}
Room
\\
\includegraphics[width=1.3cm]{figures/room-32-32-4.png}\end{center}
\end{minipage}}
& 20 & 89 & {\bf96 (89)} & 94 {\bf(89)} & 0.22 & {\bf0.17} & 0.18 & 1,823 & {\bf1,038} & 1,101 &  11.19 & {\bf4.38} & 6.33 \\ \cline{2-14} 
 & 22 & 83 & {\bf91 (83)} & {\bf91 (83)} & 0.61 & {\bf0.49} & 0.51 & 7,851 & {\bf5,648} & 5,888 &  17.51 & {\bf9.76} & 9.83 \\ \cline{2-14} 
 & 24 & 79 & {\bf86 (79)} & 84 {\bf(79)} & 0.69 & {\bf0.53} & 0.55 & 8,392 & {\bf5,007} & 5,160 &  21.55 & {\bf14.83} & 16.68 \\ \cline{2-14} 
 & 26 & 47 & {\bf57 (47)} & 55 (46) & 1.32 & {\bf1.01} & 1.14 & 15,791 & {\bf11,087} & 12,108 &  53.68 & {\bf43.97} & 45.91 \\ \cline{2-14} 
 & 28 & 45 & {\bf52 (45)} & 50 (44) & 1.45 & {\bf0.95} & 0.96 & 15,951 & {\bf10,184} & 10294 &  55.70 & {\bf48.93} & 50.81 \\ \cline{2-14} 
 & 30 & 28 & {\bf36 (28)} & 34 {\bf(28)} & 2.08 & {\bf1.21} & 1.45 & 21,279 & {\bf10,284} & 12,117&  73.22 & {\bf65.32} & 67.28  \\ \cline{2-14} 
 & 32 & 17 & {\bf24 (17)} & {\bf24 (17)} & 1.88 & {\bf1.39} & 1.70 & 22,152 & {\bf13,943} & 16,327&  83.77 & {\bf77.02} & 77.14 \\ \cline{2-14} 
  &34 & 9 & {\bf14 (9)} & {\bf14 (9)} & 3.99 & {\bf2.70} & 3.24 & 39,447 & {\bf22,611} & 28,392 &  91.36 & {\bf86.56} & 86.63 \\ \hline \hline
\multirow{9}{*}{
\begin{minipage}{.07\textwidth}
\begin{center}
Maze
\\
\includegraphics[width=1.3cm]{figures/maze-128-128-2.png}\end{center}
\end{minipage}}
 &30 & 90 & {\bf91 (90)} & 90 {\bf(90)} & 0.54 & 0.47 & {\bf0.42} & 500 & 373 & {\bf289} &  10.49 & {\bf9.51} & 10.38 \\ \cline{2-14} 
 & 32 & 84 & {\bf87 (84)} & {\bf87 (84)} & 0.49 & {\bf0.39} & 0.42 & 519 & 427 & {\bf397} &  16.42 & {\bf13.59} & 13.60 \\ \cline{2-14} 
 & 34 & 80 & 82 {\bf(80)} & {\bf84 (80)} & 0.58 & {\bf0.50} & 0.52 & 908 & {\bf763} & 780 &  20.46 & 18.59 & {\bf16.73} \\ \cline{2-14} 
 & 36 & 80 & 81 {\bf(80)} & {\bf82} (79) & 0.73 & 0.65 & {\bf0.57} & 1,200 & 1,067 & {\bf910} &  20.66 & 19.68 & {\bf18.68} \\ \cline{2-14} 
 & 38 & 64 & {\bf65 (64)} & {\bf65 (64)} & 0.68 & 0.57 & {\bf0.53} & 900 & 740 & {\bf663} &  36.44 & 35.40 & {\bf35.38} \\ \cline{2-14} 
 & 40 & 56 & 60 {\bf(56)} & {\bf62 (56)} & 0.85 & 0.80 & {\bf0.75} & 1,194 & 1,099 & {\bf1,026} &  44.47 & 40.79 & {\bf38.85} \\ \cline{2-14} 
 & 42 & 54 & 56 {\bf(54)} & {\bf57 (54)} & 1.86 & 1.69 & {\bf1.47} & 2,223 & 1,973 & {\bf1,580} &  47.00 & 45.11 & {\bf44.03} \\ \cline{2-14} 
  & 44 & 45 & 49 {\bf(45)} & {\bf50 (45)} & 1.08 & 1.06 & {\bf0.87} & 1,389 & 1,343 & {\bf1,055}&  54.49 & 50.82 & {\bf49.75}  \\ \cline{2-14}
  & 46 & 37 & {\bf40 (37)} & {\bf40 (37)} & 1.61 & 1.50 & {\bf1.31} & 2,021 & 1,743 & {\bf1,506} &  63.60 & 60.77 & {\bf60.71} \\  \hline
  \hline
  
%  \multirow{2}{*}{Map} & \multirow{2}{*}{$k$} & \multicolumn{3}{c|}{PAR10 Score} & \multicolumn{3}{c|}{Success Rate (\%)} & \multicolumn{3}{c|}{Runtime (min)} & \multicolumn{3}{c|}{CT Size (nodes)} \\ \cline{3-14} 
% &  & CBSH2 & ML-S &ML-O & CBSH2 &  ML-S &ML-O & CBSH2 &  ML-S &ML-O & CBSH2 &  ML-S &ML-O\\ \hline
  \multirow{13}{*}{
\begin{minipage}{.07\textwidth}
\begin{center}
Random
\\
\includegraphics[width=1.3cm]{figures/smallmap.png}\end{center}
\end{minipage}}
 &17 & 96 & {\bf97 (96)} & {\bf97 (96)} & 0.11 & {\bf0.08} & 0.10 & 1,195 & {\bf743} & 902 &  4.10 & {\bf3.13} & 3.19 \\ \cline{2-14} 
 & 18 & 95 & {\bf95 (95)} & 94 (94) & 0.32 & {\bf0.23} & 0.31 & 5032 & {\bf3,105} & 4,148 &  5.32 & {\bf5.27} & 6.29 \\ \cline{2-14} 
 & 19 & 92 & {\bf93 (92)} & {\bf93 (92)} & 0.44 & {\bf0.32} & 0.36 & 7,208 & {\bf4,264} & 4677 &  8.41 & {\bf7.38} & 7.43\\ \cline{2-14} 
 & 20 & 88 & {\bf91 (88)} & {\bf91 (88)} & 0.43 & {\bf0.30} & 0.36 & 7,834 & {\bf3,829} & 4,595 &  12.38 & {\bf9.37} & 9.48 \\ \cline{2-14} 
 & 21 & 79 & {\bf83 (79)} & 81 {\bf(79)} & 0.59 & {\bf0.49} & 0.56 & 8,814 & {\bf5,244} & 6,927 &  21.47 & {\bf17.53} & 19.54 \\ \cline{2-14} 
 & 22 & 74 & {\bf80 (74)} & 77 {\bf(74)} & 0.90 & {\bf0.48} & 0.59 & 15,884 & {\bf6,286} & 7,577 &  26.66 & {\bf20.69} & 23.54 \\ \cline{2-14} 
 & 23  & 74 & {\bf80 (74)} & {\bf80 (74)} & 0.96 & {\bf0.56} & 0.78 & 17,952 & {\bf8,118} & 11,555 &  26.71 & {\bf20.60} & 20.81\\ \cline{2-14} 
  & 24 & 62 & {\bf71 (62)} & 66 (60) & 1.14 & {\bf0.77} & 0.92 & 20,433 & {\bf10,413} & 12,422 &  38.87 & {\bf29.82} & 34.71 \\ \cline{2-14}
  & 25 & 55 & {\bf62 (55)} & 59 {\bf(55)} & 1.19 & {\bf0.85} & 1.02 & 21,725 & {\bf12,137} & 14,874 &  45.66 & {\bf38.84} & 41.74 \\ \cline{2-14} 
  & 26 & 39 & {\bf48 (39)} & 45 {\bf(39)} & 1.27 & {\bf0.87} & 1.24 & 19,236 & {\bf8,053} & 13,301 &  61.50 & {\bf52.75} & 55.82 \\\cline{2-14} 
  & 27 & 27 & {\bf38 (27)} & 37 {\bf(27)} & 1.36 & {\bf0.99} & 1.05 & 26,642 & {\bf16,597} & 17,130 &  73.41 & {\bf62.72} & 63.79 \\\cline{2-14} 
  & 28 & 24 & {\bf31 (24)} & 29 (23) & 1.66 & {\bf0.83} & 1.10 & 26,597 & {\bf9,239} & 13,423 &  76.45 & {\bf69.41} & 71.46 \\\cline{2-14} 
  & 29 & 17 & {\bf27 (17)} & 24 {\bf(17)} & 4.04 & {\bf2.74} & 3.39 & 63,661 & {\bf35,485} & 44,179 &  83.69 & {\bf74.07} & 77.02 \\ \hline \hline
   \multirow{8}{*}{
\begin{minipage}{.07\textwidth}
\begin{center}
Game
\\
\includegraphics[width=1.3cm]{figures/lak503d.jpg}\end{center}
\end{minipage}}
& 95 & 85 & {\bf91 (85)} & {\bf91 (85)} & 4.75 & {3.22} & {\bf3.02} & 3,006 & 1,714 & {\bf1,662} &  94.04 & 57.23 & {\bf57.13}  \\ \cline{2-14} 
 & 100 & 68 & {\bf77 (68)} & 75 {\bf(68)} & 6.76 & {\bf5.49} & 5.94 & 4,100 & {\bf3,114} & 3,341 &  196.66 & {\bf145.09} & 156.74 \\ \cline{2-14} 
 & 105  & 64 & {\bf72 (64)} & {\bf72 (64)} & 6.59 & 5.63 & {\bf5.32} & 3,959 & 3,130 & {\bf2,896} &  220.22 & 173.47 & {\bf172.21}\\ \cline{2-14} 
  & 110 & 59 & {\bf67 (59)} & {\bf67 (59)} & 6.58 & 6.03 & {\bf5.92} & 3,978 & 3,652 & {\bf3,596} &  249.89 & 202.61 & {\bf202.39} \\ \cline{2-14} 
 & 115 & 50 & 61 {\bf(50)} & {\bf64 (50)} & 6.99 & 5.66 & {\bf5.64} & 3,864 & 2,713 & {\bf2,705} &  303.50 & 240.79 & {\bf223.41} \\ \cline{2-14} 
 & 120 & 35 & {\bf44 (35)} & {\bf44} (34) & 9.59 & {\bf8.76} & 8.80 & 5,351 & {\bf4,643} & 4,691 &  393.27 & {\bf341.63} & 341.82 \\ \cline{2-14} 
 & 125 & 34 & 41 {\bf(34)} & {\bf42 (34)} & 9.32 & 7.77 & {\bf7.58} & 5,145 & 4,153 & {\bf4,054} &  399.18 & 358.91 & {\bf353.32} \\ \cline{2-14} 
 & 130 & 19 & {\bf26 (19)} & 25 (18) & {\bf4.83} & 5.00 & 4.85 & 2,486 & 2,498 & {\bf2,338} &  487.01 & {\bf447.22} & 453.05  \\ \hline 
\end{tabular}
\caption{Success rates, average runtimes and CT sizes of instances solved by all solvers, and PAR10 scores (calculated using the runtimes in minutes) for different number of agents $k$. For the success rates of \MLS and \MLO, the fractions of instances solved by both the solver and the baseline are given in parentheses (bolded if the solver solves all instances that CBSH2 solves). \label{thefulltable}
}
\end{table*}

\begin{table*}[htbp]
\tiny
\begin{center}

\begin{tabular}{|l|l|}
\hline
Index & Feature \\ \hline
1 & Binary indicators for edge conflicts. \\ \hline
2 & Binary indicators for vertex conflicts. \\ \hline
3 & Binary indicators for cardinal conflicts. \\ \hline
4 & Binary indicators for semi-cardinal conflicts. \\ \hline
5 & Binary indicators for non-cardinal conflicts. \\ \hline
6 & Minimum of the numbers of conflicts involving agent $a_i$ ($a_j$) that have been selected and resolved. \\ \hline
7 & Maximum of the numbers of conflicts involving agent $a_i$ ($a_j$) that have been selected and resolved. \\ \hline
8 & Sum of the numbers of conflicts involving agent $a_i$ ($a_j$) that have been selected and resolved. \\ \hline
9 & Minimum of the numbers of conflicts at vertex $u$ ($v$) that have been selected and resolved. \\ \hline
10 & Maximum of the numbers of conflicts at vertex $u$ ($v$) that have been selected and resolved. \\ \hline
11 & Sum of the numbers of conflicts at vertex $u$ ($v$) that have been selected and resolved. \\ \hline
12 & Minimum of the numbers of conflicts that agent $a_i$ ($a_j$) is involved in \\ \hline
13 & Maximum of the numbers of conflicts that agent $a_i$ ($a_j$) is involved in \\ \hline
14 & Sum of the numbers of conflicts that agent $a_i$ ($a_j$) is involved in \\ \hline
15 & Time step $t$ of the conflict. \\ \hline
16 & Ratio of $t$ and the makespan of the solution. \\ \hline
17 & Minimum of the costs of the path of agent $a_i$ ($a_j$). \\ \hline
18 & Maximum of the costs of the path of agent $a_i$ ($a_j$). \\ \hline
19 & Sum of the costs of the path of agent $a_i$ ($a_j$). \\ \hline
20 & Absolute difference of the costs of the path of agent $a_i$ ($a_j$) \\ \hline
21 & Ratio of the costs of the path of agent $a_i$ ($a_j$). \\ \hline
22 & Minimum of the differences of the cost of the path of agent $a_i$ ($a_j$) and its individually cost-minimal path. \\ \hline
23 & Maximum of the differences of the cost of the path of agent $a_i$ ($a_j$) and its individually cost-minimal path. \\ \hline
24 & Minimum of the ratios of the cost of the path of agent $a_i$ ($a_j$) and the cost of its individually cost-minimal path. \\ \hline
25 & Maximum of the ratios of the cost of the path of agent $a_i$ ($a_j$) and the cost of its individually cost-minimal path. \\ \hline
26 & Minimum of the ratios of the cost of the path of agent $a_i$ ($a_j$) and $N_{\Cost}$. \\ \hline
27 & Maximum of the ratios of the cost of the path of agent $a_i$ ($a_j$) and $N_{\Cost}$. \\ \hline
28 &Binary indicator whether none of agents $a_i$ and $a_j$ has reached its goal by time step $t$. \\ \hline
29 & Binary indicator whether at least one of agents $a_i$ and $a_j$ has reached its goal by time step $t$.\\ \hline
30 & Minimum of the differences of the cost of the path of agent $a_i$ ($a_j$) and $t$. \\ \hline
31 & Maximum of the differences of the cost of the path of agent $a_i$ ($a_j$) and $t$. \\ \hline
32 & Minimum of the ratios of the cost of the path of agent $a_i$ ($a_j$) and $t$. \\ \hline
33 & Maximum of the ratios of the cost of the path of agent $a_i$ ($a_j$) and $t$. \\ \hline
34 &  Number of conflicts $c'\in N_{\Conflicts}$ such that $\min\{d_{q,q'}:q\in V^T_c, q'\in V^T_{c'}\}=0$.  \\ \hline
35 &  Number of conflicts $c'\in N_{\Conflicts}$ such that $\min\{d_{q,q'}:q\in V^T_c, q'\in V^T_{c'}\}=1$.\\ \hline
36 & Number of conflicts $c'\in N_{\Conflicts}$ such that $\min\{d_{q,q'}:q\in V^T_c, q'\in V^T_{c'}\}=2$. \\ \hline
37 & Number of conflicts $c'\in N_{\Conflicts}$ such that $\min\{d_{q,q'}:q\in V^T_c, q'\in V^T_{c'}\}=3$. \\ \hline
38 & Number of conflicts $c'\in N_{\Conflicts}$ such that $\min\{d_{q,q'}:q\in V^T_c, q'\in V^T_{c'}\}=4$. \\ \hline
39 & Number of conflicts $c'\in N_{\Conflicts}$ such that $\min\{d_{q,q'}:q\in V^T_c, q'\in V^T_{c'}\}=5$. \\ \hline
40 & Number of agents $a$ such that there exists $q'\in V_a$ and $q\in V_c^T$ such that $d_{q,q'}=0$. \\ \hline
41 & Number of agents $a$ such that there exists $q'\in V_a$ and $q\in V_c^T$ such that $d_{q,q'}=1$.  \\ \hline
42 & Number of agents $a$ such that there exists $q'\in V_a$ and $q\in V_c^T$ such that $d_{q,q'}=2$.  \\ \hline
43 & Number of agents $a$ such that there exists $q'\in V_a$ and $q\in V_c^T$ such that $d_{q,q'}=3$.  \\ \hline
44 & Number of agents $a$ such that there exists $q'\in V_a$ and $q\in V_c^T$ such that $d_{q,q'}=4$.  \\ \hline
45 & Number of agents $a$ such that there exists $q'\in V_a$ and $q\in V_c^T$ such that $d_{q,q'}=5$.  \\ \hline
46 &  Number of conflicts $c'\in N_{\Conflicts}$ such that $\min\{d_{q,q'}:q\in V_c,q'\in V_{c'}\}=0$.\\ \hline
47 &  Number of conflicts $c'\in N_{\Conflicts}$ such that $\min\{d_{q,q'}:q\in V_c,q'\in V_{c'}\}=1$.\\ \hline
48 &  Number of conflicts $c'\in N_{\Conflicts}$ such that $\min\{d_{q,q'}:q\in V_c,q'\in V_{c'}\}=2$.\\ \hline
49 &  Number of conflicts $c'\in N_{\Conflicts}$ such that $\min\{d_{q,q'}:q\in V_c,q'\in V_{c'}\}=3$.\\ \hline
50 & Number of conflicts $c'\in N_{\Conflicts}$ such that $\min\{d_{q,q'}:q\in V_c,q'\in V_{c'}\}=4$. \\ \hline
51 & Number of conflicts $c'\in N_{\Conflicts}$ such that $\min\{d_{q,q'}:q\in V_c,q'\in V_{c'}\}=5$. \\ \hline
52 & Minimum of the widths of level $t-2$ of the MDDs for agent $a_i (a_j)$. \\ \hline
53 & Maximum of the widths of level $t-2$ of the MDDs for agent $a_i (a_j)$. \\ \hline
54 & Minimum of the widths of level $t-1$ of the MDDs for agent $a_i (a_j)$. \\ \hline
55 & Maximum of the widths of level $t-1$ of the MDDs for agent $a_i (a_j)$. \\ \hline
56 & Minimum of the widths of level $t$ of the MDDs for agent $a_i (a_j)$. \\ \hline
57 & Maximum of the widths of level $t$ of the MDDs for agent $a_i (a_j)$. \\ \hline
58 & Minimum of the widths of level $t+1$ of the MDDs for agent $a_i (a_j)$. \\ \hline
59 & Maximum of the widths of level $t+1$ of the MDDs for agent $a_i (a_j)$. \\ \hline
60 & Minimum of the widths of level $t+2$ of the MDDs for agent $a_i (a_j)$. \\ \hline
61 & Maximum of the widths of level $t+2$ of the MDDs for agent $a_i (a_j)$. \\ \hline
62 & Number of vertices $q'$ in graph $G$ such that $\min\{d_{q',q}:q\in V_c\}=1$. \\ \hline
63 & Number of vertices $q'$ in graph $G$ such that $\min\{d_{q',q}:q\in V_c\}=2$. \\ \hline
64 & Number of vertices $q'$ in graph $G$ such that $\min\{d_{q',q}:q\in V_c\}=3$. \\ \hline
65 & Number of vertices $q'$ in graph $G$ such that $\min\{d_{q',q}:q\in V_c\}=4$. \\ \hline
66 & Number of vertices $q'$ in graph $G$ such that $\min\{d_{q',q}:q\in V_c\}=5$.\\ \hline
67 & Weight of the edge between agents $a_i$ and $a_j$ in the weighted dependency graph. \\ \hline
\end{tabular}
\end{center}

\caption{Features with their indices.\label{featureIndex}}
\end{table*}

\iffalse
\begin{table*}[]
\begin{tabular}{|l|r|r|r|r|r|r|r|}
\hline
 &  & Game & Random & Maze & Room & Warehouse & City \\ \hline
 &Training \#Agents & 100 & 18 & 30 & 26 & 30 & 180\\\hline
\multirow{2}{*}{\begin{tabular}[l]{@{}l@{}}Training on \\ the same map\end{tabular}} & Swapped (\%) & 4.40 & 10.89 & 4.5 & 12.58 & 5.78 & 2.89 \\ \cline{2-8} 
% & 0-1 Error (\%) & 48.74 & 37.28 & 15.94 & 34.33 & 15.08 & 25.23 \\ 
& Top 10\% Accuracy (\%) & 60.16 & 69.03 & 87.69 & 67.56 & 84.93 & 83.05\\
\hline
\multirow{2}{*}{\begin{tabular}[l]{@{}l@{}}Training on \\ other maps\end{tabular}} & Swapped (\%) & 7.45 & 19.64 & 21.98 & 15.24 & 6.08 & 7.66 \\ \cline{2-8} 
% & 0-1 Error (\%) & 55.49 & 56.55 & 52.05 & 35.67 & 13.23 & 25.23 \\ 
& Top 10\% Accuracy (\%) & 53.13 & 50.44 & 49.90 & 66.80 & 86.85 & 78.57\\

\hline
\end{tabular}
\end{table*}
\fi
\end{document}